\documentclass{article} 
\usepackage[numbers]{natbib}

\usepackage[final]{neurips_2024}

\usepackage[utf8]{inputenc} 
\usepackage[T1]{fontenc}    
\usepackage{hyperref}       
\usepackage{url}            
\usepackage{booktabs}       
\usepackage{amsfonts}       
\usepackage{nicefrac}       
\usepackage{microtype}      
\usepackage{xcolor}   
\usepackage[table]{xcolor} 
\usepackage{tabularx}       
\usepackage{pifont}
\usepackage{graphicx}
\usepackage{xcolor}
\usepackage{lipsum}
\usepackage{multirow}
\usepackage{marvosym}
\usepackage{xspace}
\usepackage{blindtext}
\usepackage{amsmath}
\usepackage{algorithm}

\usepackage{tcolorbox}
\usepackage{fontawesome5}

\usepackage{algpseudocode}
\usepackage{wrapfig}
\usepackage{enumitem}

\usepackage{subcaption}

\usepackage{amssymb}
\usepackage{amsmath}

\usepackage[table,dvipsnames,svgnames]{xcolor}
\usepackage{tablefootnote}
\usepackage{subcaption}

\usepackage{wrapfig}
\usepackage{lipsum} 
\usepackage{booktabs} 
\usepackage{adjustbox}

\definecolor{catgray}{gray}{0.92}
\hypersetup{colorlinks,urlcolor=purple,citecolor=cyan,linkcolor=purple}

\makeatletter
\DeclareRobustCommand\onedot{\futurelet\@let@token\@onedot}
\def\@onedot{\ifx\@let@token.\else.\null\fi\xspace}

\def\eg{\emph{e.g}\onedot} 
\def\ie{\emph{i.e}\onedot}

\makeatother

\definecolor{citecolor}{rgb}{0.21,0.49,0.74}
\definecolor{linkcolor}{HTML}{ED1C24}
\definecolor{graycolor}{rgb}{0.95,0.95,0.95}
\definecolor{cvprblue}{rgb}{0.21,0.49,0.74}
\definecolor{memory}{rgb}{0.82, 0.51, 0.50} 
\definecolor{current}{rgb}{0.49, 0.60, 0.74} 
\definecolor{softgreen}{HTML}{4CAF50}       
\definecolor{softyellow}{HTML}{E0A800}      
\definecolor{rowgreen}{HTML}{E8F5E9}        
\definecolor{rowyellow}{HTML}{FFF8E1}       

\usepackage[capitalize]{cleveref}
\crefname{section}{Sec.}{Secs.}

\crefname{table}{Tab.}{Tabs.}
\crefname{figure}{Fig.}{Figs.}

\newcommand{\xmark}{\ding{55}}

\usepackage{tikzsymbols}
\definecolor{goodgreen}{RGB}{46, 204, 113}  
\definecolor{normalyellow}{RGB}{241, 196, 15}
\definecolor{badred}{RGB}{231, 76, 60}    

\newcommand{\good}{\Smiley[1.2][goodgreen!60!white]}
\newcommand{\bad}{\Sadey[1.2][badred!60!white]}
\newcommand{\normal}{\Neutrey[1.2][normalyellow!80!orange]}

\usepackage[toc,page]{appendix}

\title{HY-World 2.0: A Multi-Modal World Model for Reconstructing, Generating, and Simulating 3D Worlds}

\author{\textbf{Tencent Hunyuan*}}

\begin{document}

\maketitle

\makeatletter
\let\original@makefnmark\@makefnmark  
\makeatother

\makeatletter
\def\@makefnmark{}
\makeatother
\footnotetext{$*$ HY-World team contributors are listed at the end of the report.}

\makeatletter
\let\@makefnmark\original@makefnmark  
\makeatother

\begin{abstract}

We introduce \textbf{HY-World 2.0}, a multi-modal world model framework that advances our prior project {HY-World 1.0}. 
HY-World 2.0 accommodates diverse input modalities, including text prompts, single-view images, multi-view images, and videos, and produces 3D world representations.
With text or single-view image inputs, the model performs \emph{world generation}, synthesizing high-fidelity, navigable 3D Gaussian Splatting (3DGS) scenes. 
This is achieved through a four-stage method: a) \textcolor[HTML]{4285F4}{\textbf{Panorama Generation}} with HY-Pano 2.0, b) \textcolor[HTML]{34A853}{\textbf{Trajectory Planning}} with WorldNav, c) \textcolor[HTML]{FBBC05}{\textbf{World Expansion}} with WorldStereo 2.0, and d) \textcolor[HTML]{EA4335}{\textbf{World Composition}} with WorldMirror 2.0.
Specifically, we introduce key innovations to enhance panorama fidelity, enable 3D scene understanding and planning, and upgrade WorldStereo, our keyframe-based view generation model with consistent memory. 
We also upgrade WorldMirror, a feed-forward model for universal 3D prediction, by refining model architecture and learning strategy, enabling \emph{world reconstruction} from  multi-view images or videos. 
Also, we introduce \textcolor[HTML]{7B1FA2}{\textbf{WorldLens}}, a high-performance 3DGS rendering platform featuring a flexible engine-agnostic architecture, automatic IBL lighting, efficient collision detection, and training-rendering co-design, enabling interactive exploration of 3D worlds with character support.
Extensive experiments demonstrate that HY-World 2.0 achieves state-of-the-art performance on several benchmarks among open-source approaches, delivering results comparable to the closed-source model Marble. We release all model weights, code, and technical details to facilitate reproducibility and support further research on 3D world models. 
\textcolor{purple}{Project Page: \url{https://3d-models.hunyuan.tencent.com/world/}}
\end{abstract}

\begin{figure}[h!]
\centering
\includegraphics[width=0.97\textwidth]{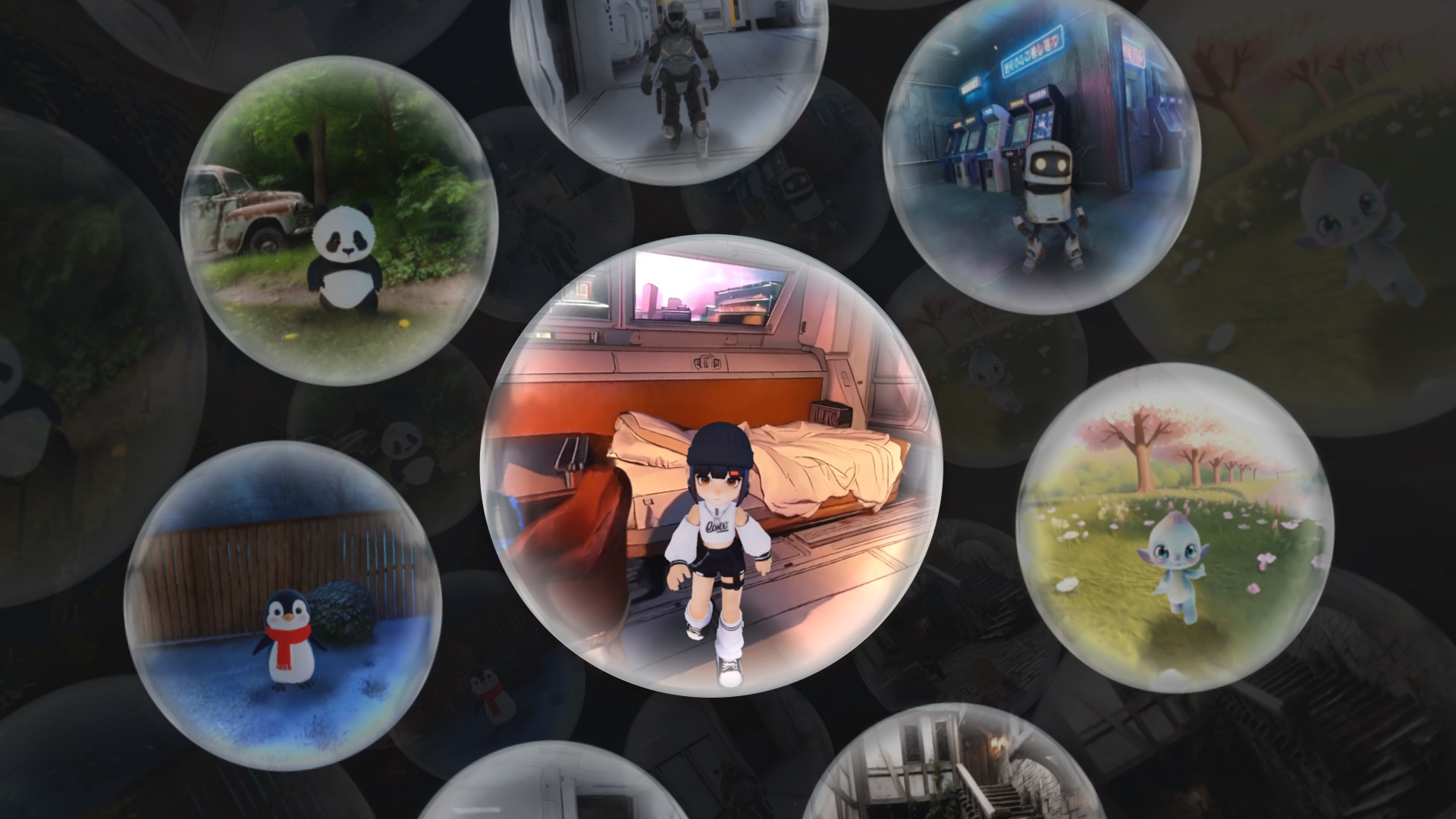}
\end{figure}


\clearpage
{
\fontsize{8.5}{8}\selectfont
\tableofcontents
}
\clearpage

\begin{figure*}[t]
    \centering
    \includegraphics[width=1.0\linewidth]{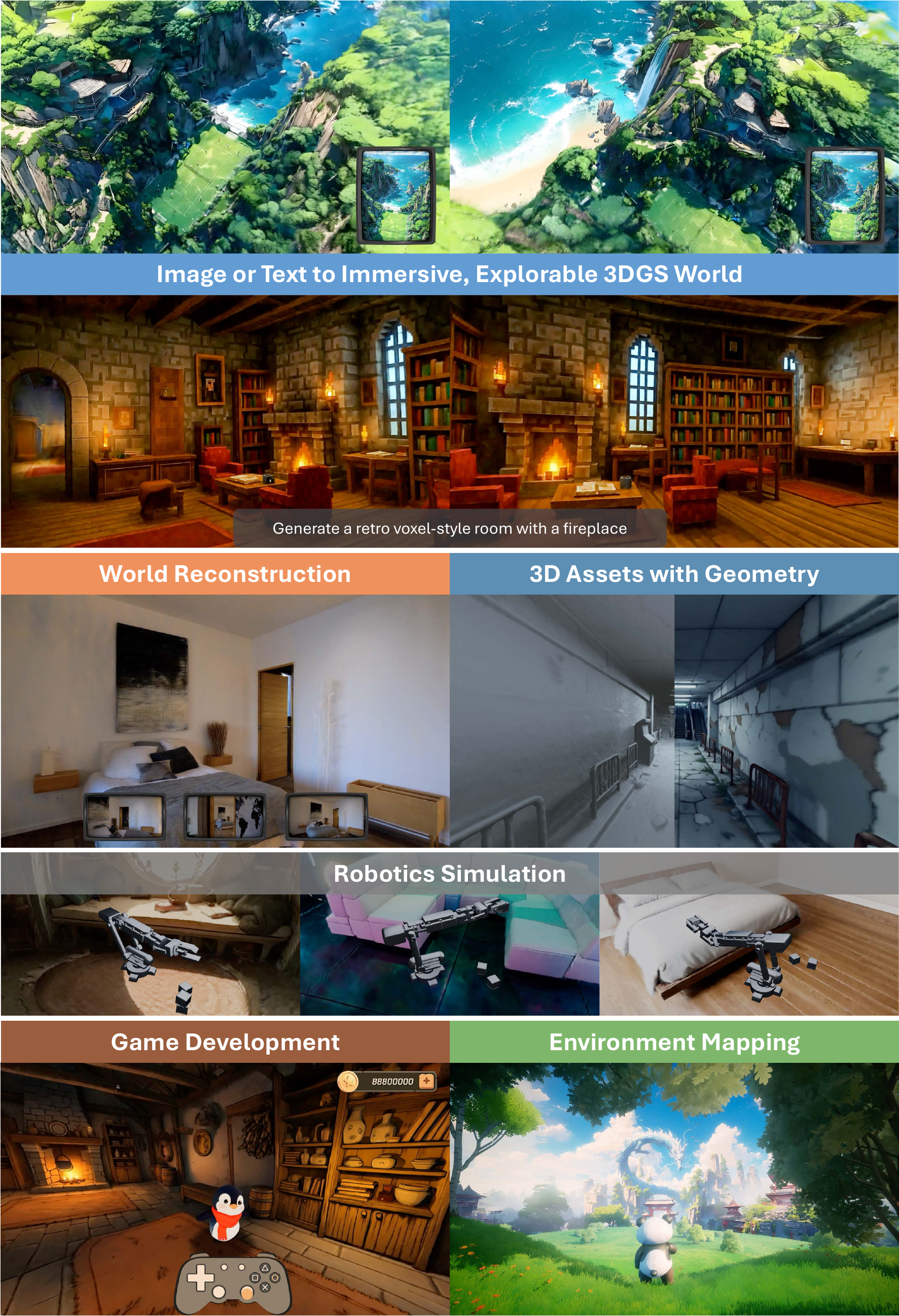}
    \caption{\textbf{Versatile applications of HY-World 2.0.} 
    Our framework unifies \textit{world generation} (synthesizing immersive, explorable 3D worlds from text or single-view images) and \textit{world reconstruction} (recovering 3D representation from multi-view inputs). These capabilities empower diverse applications, including robotics simulation, game development, and environment mapping. \label{fig:teaser}}
\end{figure*}

\clearpage

\section{Introduction}

\begin{quote}
\emph{``What Is Now Proved Was Once Only Imagined'' } \\
\raggedleft \emph{--- William Blake}
\vspace{-1mm}
\end{quote}

World models have rapidly evolved into a transformative paradigm for AI, enabling agents to simulate, understand, and interact with complex 3D environments~\cite{ha2018recurrent,dong2026learning}. 
By capturing the physical and spatial dynamics of the real world, these models are unlocking unprecedented possibilities across diverse applications, including virtual reality~\cite{hunyuanworld2025tencent}, embodied robotics~\cite{janner2022planning}, and video games~\cite {deepmind2025genie3,hyworld15}. 

Our previous explorations of generative world models involved two primary paradigms:
(1) \textbf{HY-World 1.0}~\cite{hunyuanworld2025tencent} established a robust foundation for \textit{offline 3D-based world generation}~\cite{hunyuanworld2025tencent,marble_worldlabs_2026,yang2025layerpano3d,yang2025flash,schneider2025worldexplorer,liu2025worldmirror}, explicitly modeling explorable 3D worlds with inherent 3D consistency, making them seamlessly compatible with standard computer graphics pipelines.
(2) \textbf{HY-World 1.5}~\cite{hyworld15,sun2025worldplay,wang2026worldcompass} advanced the frontier of \textit{online video-based world generation}~\cite{deepmind2025genie3,mao2025yume,hyworld15,team2026advancing}, enabling real-time, interactive world modeling driven by user actions.

Despite these remarkable advancements, the current landscape of 3D world modeling remains largely bifurcated. Existing solutions typically specialize in either \textit{generation} or \textit{reconstruction}. 
Generative approaches excel at synthesizing impressive, explorable scenes from sparse inputs like texts or single-view images, but often struggle to maintain strict reconstruction accuracy~\cite{yang2025matrix,team2026advancing}. 
Conversely, reconstruction methods focus on recovering precise 3D structures (\eg, depth, normals, and point clouds) from dense multi-view images or videos, yet they lack the generative priors necessary to hallucinate unseen regions~\cite{wang2025vggt,wang2025pi,liu2025worldmirror,lin2025depth}. 
Furthermore, while recent closed-source pioneers~\cite{marble_worldlabs_2026} have demonstrated impressive capabilities in unifying these tasks, the open-source community still lacks a comprehensive, multi-modal foundational world model that bridges the gap between imaginative generation and accurate physical reconstruction.

To address these fundamental challenges, we introduce \textbf{HY-World 2.0}, the first open-source, systematic multi-modal world model that seamlessly unifies both ``generation'' and ``reconstruction'' within an \textit{offline 3D world model} paradigm, as illustrated in \Cref{fig:teaser}. 
Designed to accommodate diverse input modalities---ranging from texts and single-view images to multi-view images and videos---HY-World 2.0 dynamically adapts its behavior based on the available conditions.

For sparse inputs (texts or single-view images), the model performs \textit{world generation} to synthesize high-fidelity, navigable 3D Gaussian Splatting (3DGS) worlds. Formally, this generation capability is driven by a novel four-stage pipeline: panorama generation, trajectory planning, world expansion, and world composition. 
Crucially, although HY-World 2.0 is fundamentally designed as an \textit{offline 3D world model}, it successfully bridges the gap between the geometric rigor of 3D representations and the rich, dynamic priors of video generation. 
By leveraging the powerful generative priors of \textit{video diffusion models} during the expansion stage, HY-World 2.0 achieves significantly expanded exploratory spaces and superior visual quality compared to the previous HY-World 1.0.

For richer visual observations (multi-view images or videos), the framework performs \textit{world reconstruction} to recover geometrically consistent and accurate 3D structures. Notably, rather than functioning as an isolated module, this \textit{world reconstruction} capability also serves as a foundational component of \textit{world generation}, powered by our upgraded feed-forward 3D reconstruction.

Beyond paradigm integration, we systematically push every component of HY-World 2.0 to its limits. 
First, we scale up  \textcolor[HTML]{4285F4}{\textit{Panorama Generation}} to \textcolor[HTML]{4285F4}{\textbf{HY-Pano 2.0}} in terms of both data and model capacity, enabling adaptive perspective-to-equirectangular (ERP) transformations from input images at arbitrary viewpoints.
Next, a scene-parsing enhanced \textcolor[HTML]{34A853}{\textit{Trajectory Planning}} algorithm, called \textcolor[HTML]{34A853}{\textbf{WorldNav}}, is introduced to produce camera trajectories for subsequent world expansion, considering both information maximization and obstacle avoidance.
For \textcolor[HTML]{FBBC05}{\textit{World Expansion}}, we upgrade our previous controllable video model~\cite{worldstereo2026} to \textcolor[HTML]{FBBC05}{\textbf{WorldStereo 2.0}}: 1) Rather than video generation, we perform generation within a keyframe space, thereby achieving superior visual fidelity. 2) We introduce a more consistent and robust memory mechanism.
In the final stage of \textcolor[HTML]{EA4335}{\textit{World Composition}}, we reconstruct the 3D environment using the upgraded \textcolor[HTML]{EA4335}{\textbf{WorldMirror 2.0}}: improved through generalized position encoding and enhanced training strategy.
Unlike standard 3DGS learning for reconstruction~\cite{kerbl20233d}, we incorporate tailored enhancements to strengthen 3DGS training on generated views, effectively bridging the gap between 3D reconstruction and generative world modeling.

By unifying all aforementioned capabilities into a cohesive system, HY-World 2.0 achieves state-of-the-art performance in 3D-based world modeling. 
Extensive experiments demonstrate our model's superiority over existing open-source competitors and competitiveness with closed-source commercial products like Marble~\cite{marble_worldlabs_2026}. 
We release all models, codes, and technical details, aiming to democratize spatial intelligence and provide a robust, open-source foundation for the research on world models.

\section{Overview}

\begin{figure}
    \centering
    \includegraphics[width=1.0\linewidth]{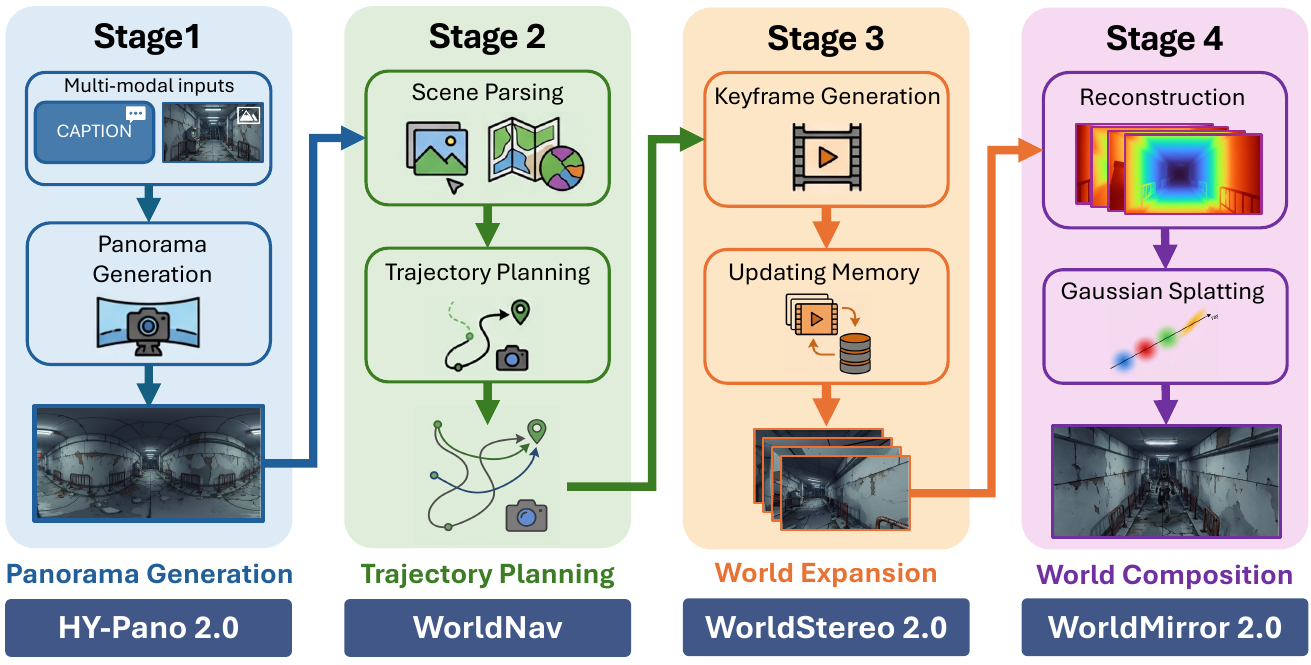}
    \caption{\textbf{Architecture of HY-World 2.0.}
    Our framework presents a four-stage process to transform multi-modal inputs into immersive 3D worlds: (1) initializing the world via \textit{Panorama Generation}, (2) deriving exploration camera paths through \textit{Trajectory Planning}, (3) expanding the world observations via memory-driven \textit{World Expansion}, and (4) constructing the final 3DGS assets using \textit{World Composition}. The core model/algorithm used in each stage is denoted at the bottom.
    \label{fig:overview}}
\end{figure}

We show the overview of HY-World 2.0 in \Cref{fig:overview}, which introduces the multi-modal world model as a four-stage pipeline, simulating the process of understanding, synthesizing, and reconstructing worlds.
Specifically, the pipeline begins with \textbf{Panorama Generation} (\Cref{sec:panorama_gen}), which translates arbitrary text or image inputs into a high-fidelity $360^{\circ}$ world initialization.
Subsequently, the elaborate \textbf{Trajectory Planning} (\Cref{sec:traj_plan}) is performed to parse and understand the initialized world, deriving optimal and information-rich observation paths.
Following these planned routes, the generative \textbf{World Expansion} (\Cref{sec:world_expansion}) utilizes a memory-updating mechanism to ensure precise camera control and multi-view consistency across generated keyframes.
Finally, \textbf{World Composition} (\Cref{sec:world_composition}) is achieved by feeding these generated sequences into \textbf{WorldMirror 2.0} (\Cref{sec:world_recon}) for robust 3D reconstruction, followed by tailored 3DGS optimization to yield immersive 3D worlds.

\section{World Generation Stage I: Panorama Generation}
\label{sec:panorama_gen}
A panorama captures a complete $360^{\circ} \times 180^{\circ}$ field-of-view (FoV) from a fixed viewpoint, offering a comprehensive and information-rich representation of entire scenes. 
Unlike standard perspective images that provide only a limited view of the physical world, $360^{\circ}$ panoramas preserve global spatial contexts and intricate semantic relationships.
Consequently, this holistic representation is increasingly recognized as a cornerstone for large-scale 3D world generation, providing the essential spatial consistency required for coherent viewpoint synthesis and immersive virtual exploration.

In this stage, we propose \textbf{HY-Pano 2.0}, which aims to synthesize high-fidelity panoramas from multi-modal conditions, including texts and single-view images. 
To achieve this, we optimize our generative pipeline across two orthogonal dimensions: 
(1) implementing an advanced data curation pipeline to overcome the inherent scarcity of panoramic data by curating high-resolution and diverse samples; and 
(2) introducing a dedicated $360^{\circ}$ generative model that implicitly learns the spatial mapping between perspective inputs and panoramic targets in a geometry-free manner, facilitating the synthesis of structurally coherent environments without requiring explicit camera metadata.

\subsection{Data}

To construct a robust foundation for high-fidelity panoramic synthesis, our data curation pipeline builds upon the established framework of HY-World 1.0~\cite{hunyuanworld2025tencent} while significantly scaling up the richness and diversity of the training data. 
Specifically, our upgraded dataset integrates two primary data sources: 
\textit{(1) Real-world captures:} We incorporate a massive collection of high-resolution, real-world panoramas to instill the model with authentic lighting, complex textures, and natural structural priors. 
\textit{(2) Synthetic assets:} To complement the real-world data, we utilize a large-scale set of synthetic environments rendered via high-end engines such as Unreal Engine (UE). These assets provide precise geometric labels and diverse, imaginative scene configurations that are otherwise difficult to obtain in the wild. 
To ensure data integrity, we implement a rigorous data filtering stage to eliminate low-quality samples, particularly those exhibiting noticeable stitching artifacts or exposed capturing equipment (\eg, panoramic camera).
This hybrid data strategy effectively broadens the semantic distribution of our dataset and mitigates the domain gap between synthetic and real-world distributions, enabling the model to generalize robustly across complex indoor and outdoor environments.

\subsection{Model}

\begin{figure}
    \centering
    \includegraphics[width=1.0\linewidth]{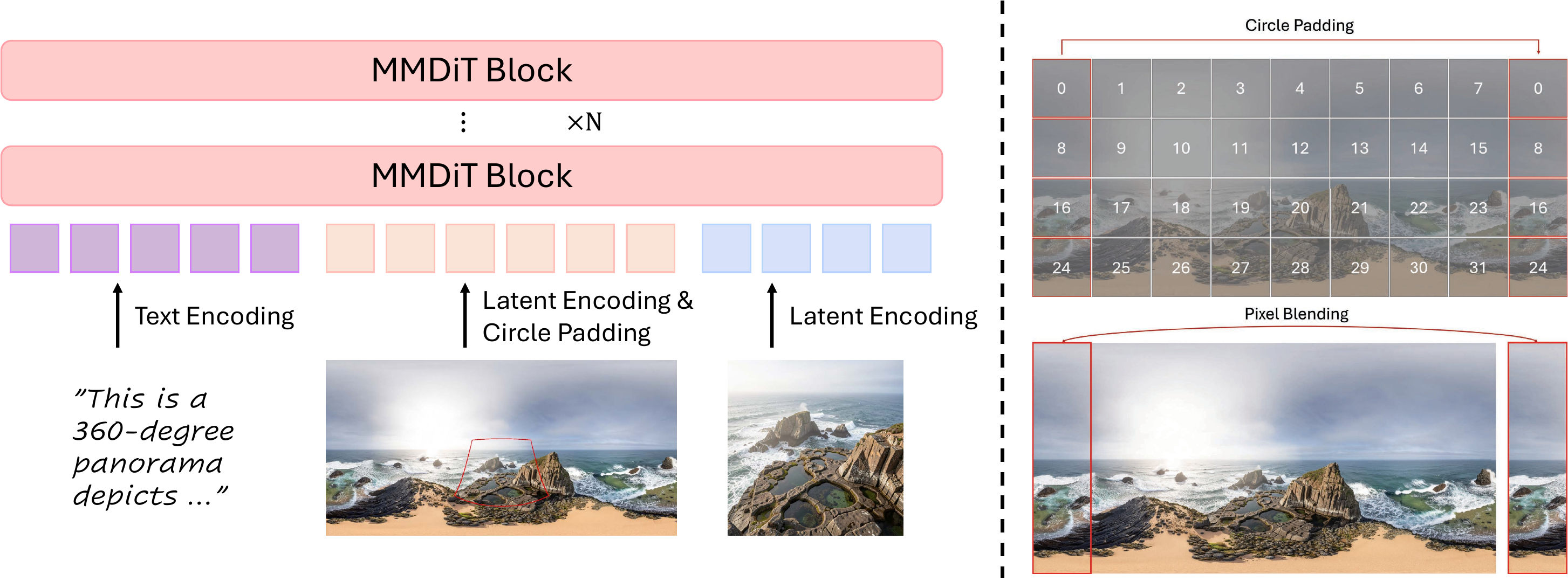}
    \caption{\textbf{Overview of the panorama generation architecture of HY-Pano 2.0.} The Left side shows the framework pipeline of panorama generation, while the right side illustrates the circle padding (latent space) and the pixel blending (pixel space) for seamless panorama generation. 
    \label{fig:pano_arch}}
\end{figure}

To achieve high-fidelity panorama synthesis from perspective inputs, we move beyond conventional methods that rely on explicit geometric warping, a paradigm previously employed in HY-World 1.0~\cite{hunyuanworld2025tencent}. This traditional pipeline typically needs precise camera intrinsic parameters (\eg, focal length and FoV) to perform spatial alignment between the perspective and equirectangular projection (ERP) domains.
However, such metadata is frequently unavailable or inaccurate in real-world scenarios.
This bottleneck inherently limits the flexibility of the HY-World 1.0 framework and often leads to noticeable projection distortion.
To address this, we adopt an \textit{implicit, adaptive mapping strategy} powered by a Multi-Modal Diffusion Transformer (MMDiT), as illustrated in \Cref{fig:pano_arch}.
Instead of relying on explicit camera priors, we process both the conditional input and the panoramic target within a unified latent space. By concatenating the conditional image latent with the panoramic noise latent as a unified sequence of tokens, the MMDiT leverages its self-attention mechanism to autonomously learn the underlying perspective-to-ERP transformation.
This purely data-driven approach allows the network to establish spatial correspondences directly within the feature space, enabling it to flexibly hallucinate missing environmental details and maintain global structural coherence, even with uncalibrated and diverse input images.

A common challenge in ERP generation is the discontinuity at the left and right edges. To eliminate these boundary artifacts, we introduce a combined refinement strategy comprising circular padding and pixel blending, as shown in the right of \Cref{fig:pano_arch}.
At the latent level, we apply circular padding to the latent features, enforcing periodic boundary conditions during the denoising process.
The padded latent is then decoded into the pixel space, where a linear pixel blending strategy is employed along the equirectangular edges. This combined harmonization effectively smooths the $360^{\circ}$ wrap-around transition, ensuring a perfectly seamless and structurally coherent panoramic output.

\section{World Generation Stage II: Trajectory Planning}
\label{sec:traj_plan}

\paragraph{Task Formulation.}
Following the synthesis of a high-fidelity panorama (\Cref{sec:panorama_gen}), the subsequent objective is to derive exploration trajectories that maximize the coverage of navigable space. To bridge this with the upcoming world expansion stage (\Cref{sec:world_expansion}), we introduce \textbf{WorldNav}, a comprehensive trajectory planning strategy. WorldNav not only generates diverse camera paths to ensure extensive viewpoint coverage but also pairs them with precise textual instructions, thereby providing explicit guidance for the downstream generative process.

\subsection{Geometric and Semantic Scene Parsing}
\label{sec:geo_init}

\begin{figure}
    \centering
    \includegraphics[width=1.0\linewidth]{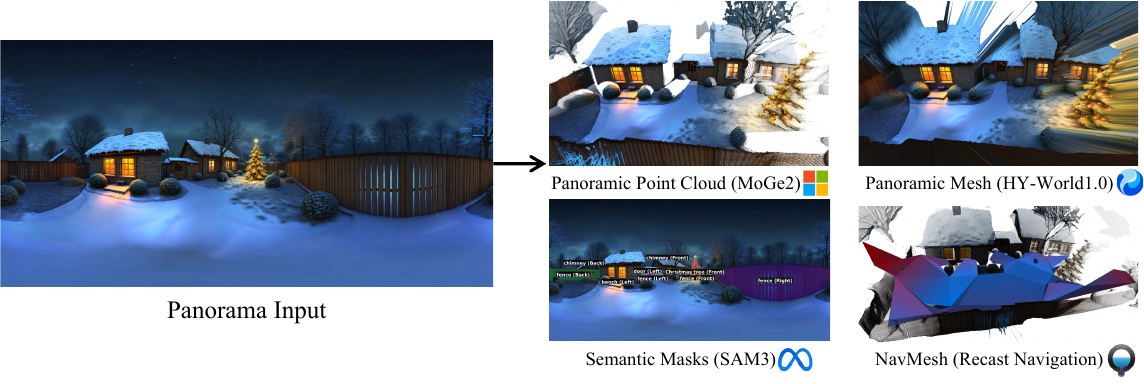}
    \caption{\textbf{The initial scene parsing for trajectory planning.} We obtain panoramic point clouds, meshes, semantic masks, and NavMesh via several pioneering works~\cite{wang2025moge,carion2025sam3segmentconcepts,hunyuanworld2025tencent,recastnavigation}.
    \label{fig:scene_parsing}}
\end{figure}

Given the panoramas, we first employ scene parsing to obtain panoramic point clouds, meshes, semantic masks, and navigation meshes for the subsequent trajectory planning, as shown in \Cref{fig:scene_parsing}.

\paragraph{Geometry-Aware Initialization.}
We initialize the scene geometry by constructing a global panoramic point cloud, $\mathbf{P}^{pan}$.
Leveraging the optimization framework from MoGe2~\cite{wang2025moge}, we align monocular depth maps via the Least-Squares Minimal Residual (LSMR) across perspective views subdivided from the ERP space. 
Crucially, to enhance the geometric quality, we increase the sampling density from the default 12 views to 42, managing the computational overhead via a GPU-accelerated LSMR solver.
Furthermore, we employ a hybrid filtering strategy, utilizing a vision-language grounding pipeline~\cite{liu2024grounding,kim2025zim} to mask unbounded sky regions, and then removing depth discontinuities (\ie, edge floaters).
This panoramic point cloud $\mathbf{P}^{pan}$ serves as the fundamental geometric representation across the subsequent trajectory planning, world expansion, and composition stages.
Following HY-World 1.0~\cite{hunyuanworld2025tencent}, we build the panoramic mesh at a lower resolution, which works for strict collision detection during trajectory planning.

\paragraph{Semantic Grounding and Navigability Analysis.}
To facilitate scene-aware camera control, we perform both semantic parsing and topological analysis of the panoramic scene.
Specifically, we apply Qwen3-VL~\cite{yang2025qwen3} to identify key spatial landmarks and obstacles within the panorama.
Subsequently, SAM3~\cite{carion2025sam3segmentconcepts} is utilized to yield 2D semantic masks for these objects. 
We then localize their centroids into the 3D space as 3D masks, applying statistical filtering to eliminate background outliers.

Simultaneously, we construct a Navigation Mesh (NavMesh) using Recast Navigation~\cite{recastnavigation} to define the traversable regions for the camera agent.
To ensure physically plausible camera movement, we apply several geometric refinements to the raw NavMesh.
First, we correct surface irregularities by snapping misaligned vertices to the physical ground via dense ray-casting.
Second, we perform boundary erosion using a KD-Tree accelerated search to prevent the camera from moving too close to obstacles.
Finally, we connect isolated navigable areas by detecting boundary nodes and synthesizing bridge polygons, thereby ensuring a continuous and fully navigable NavMesh.

\subsection{WorldNav}
\label{sec:traj_plan_details}

\begin{figure}
    \centering
    \includegraphics[width=1.0\linewidth]{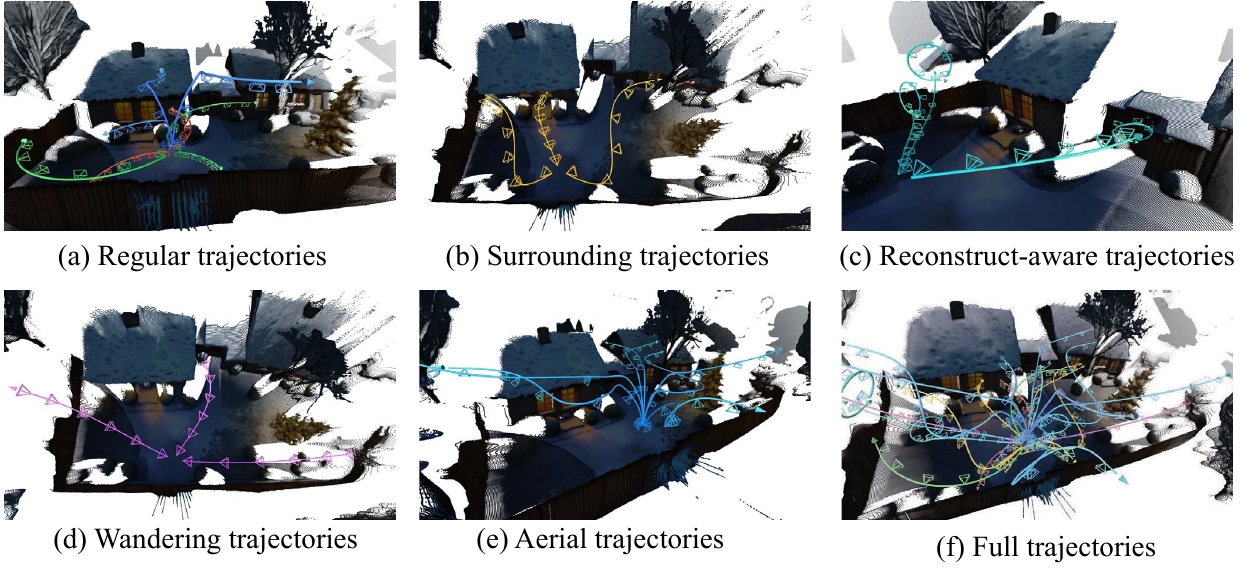}
    \caption{\textbf{Illustration of five modes of trajectories planned in WorldNav.} Some trajectories are omitted for a simplified visualization.
    \label{fig:traj_plan}}
\end{figure}

\begin{table}
\centering
\caption{\textbf{Trajectory details of WorldNav.} The aerial category comprises both surrounding and wandering trajectories. Note that the maximum number for surrounding and reconstruct-aware trajectories is determined by the count of object segments detected within the panorama.\label{tab:traj_details}}
\small
\begin{tabular}{lcccccc}
\toprule
 & Regular & Surrounding & Recon-Aware & Wandering & Aerial & Total\tabularnewline
\midrule
Max Number & 9 & 5 & 10 & 3 & 8 & 35\tabularnewline
Attached to Object & $\times$ & $\checkmark$ & $\checkmark$ & $\times$ & -- & --\tabularnewline
Iterative & $\times$ & $\times$ & $\checkmark$ & $\times$ & $\times$ & --\tabularnewline
\bottomrule
\end{tabular}
\end{table}

Given the panoramic mesh, the NavMesh, and the 3D semantic landmarks, we design five heuristic trajectory modes for WorldNav.
These trajectories start from the panorama's center and are designed to comprehensively cover diverse viewpoints while ensuring collision-free movement, as illustrated in \Cref{fig:traj_plan}.

\paragraph{Regular Trajectories.}
We employ regular trajectories to generally expand the visual coverage beyond the fixed origin of the panoramic space, as visualized in \Cref{fig:traj_plan}(a).
First, we uniformly subdivide the panorama into three perspective views with a $120^\circ$ FoV-x.
For each view, we define an orbital target at the center point, positioned at the median depth of this view.
The camera then orbits this target with a pitch angle of $+45^\circ$ and azimuth offsets of $\pm120^\circ$. 
Specifically, we prioritize generating the pitch rotation before the azimuthal ones; this sequence ensures a global overview and facilitates consistent background generation.
To further strengthen coverage with aerial perspectives, we apply an additional $+60^\circ$ azimuth rotation to the pitched orbits.
Crucially, we utilize ray-casting to prevent the camera from clipping into the panoramic mesh.
Trajectories that result in negligible movement due to collision detection are discarded. 

\paragraph{Surrounding Trajectories.} 
To facilitate the visual quality of foregrounds during the scene generation, we design surrounding trajectories that circle around the most significant objects, as shown in \Cref{fig:traj_plan}(b).
The orbit radius is adaptively adjusted based on the object's 3D size: larger landmarks are observed from a greater distance to ensure the entire target fits within the FoV. 
To ensure collision-free navigation, we uniformly sample 72 candidate nodes along the ideal circle and validate them via ray-casting against the NavMesh.
Valid nodes are then connected to form a continuous arc using a bidirectional greedy search.
To maintain a smooth path, we apply a tail pruning mechanism that removes the ends of the trajectory if they diverge significantly from the intended circular direction.
Finally, we connect the start node to the nearest endpoint of the arc using the Dijkstra algorithm~\cite{dijkstra2022note} on the NavMesh. 

\paragraph{Reconstruct-Aware Trajectories.} 
To mitigate the gaps for the subsequent 3D reconstruction, we introduce iterative reconstruction-aware trajectories that specifically target under-observed regions, as illustrated in \Cref{fig:traj_plan}(c).
In the panoramic mesh, these missing areas typically manifest as stretched and sharp faces (refer to \Cref{fig:scene_parsing}). 
We detect these regions by identifying mesh faces that exceed a heuristic aspect ratio threshold. To prioritize significant reconstruction targets, we employ Non-Maximum Suppression (NMS) to extract representative cluster centers of these degenerate faces and associate them with their nearest semantic landmarks, establishing them as key reconstruction nodes.
Similar to surrounding trajectories, we generate candidate viewpoints around these nodes, selecting the endpoint that aligns its vertical viewing angle with the missing region. When multiple candidates exist, we prioritize the one offering the maximum visible range within the NavMesh.
Moreover, to increase the ratio of novel views, we append an iterative orbiting trajectory: starting from the selected endpoint, the camera orbits the reconstruction node while maintaining a fixed gaze direction toward the target.

\paragraph{Wandering Trajectories.}
To maximize scene coverage and reach the environmental boundaries of the scene, we present wandering trajectories as shown in \Cref{fig:traj_plan}(d).
These paths simulate the exploration of an autonomous agent, specifically targeting the farthest reachable points within the panoramic scene.
This trajectory is particularly effective for extending visibility in narrow environments, such as streets and corridors.
Formally, we partition the NavMesh into eight uniform angular sectors relative to the origin. Within each reachable sector, we utilize the Dijkstra distance field to identify and direct the camera toward the node farthest from the starting point.

\paragraph{Aerial Trajectories.}
Finally, we introduce auxiliary aerial trajectories to eliminate remaining blind viewpoints, as visualized in \Cref{fig:traj_plan}(e). Specifically, we augment the existing surrounding and wandering trajectories by applying a $+45^\circ$ upward pitch. To ensure geometric validity, this pitch angle is dynamically reduced when the camera view intersects the panoramic mesh, thereby preventing collisions.

\section{World Generation Stage III: World Expansion}
\label{sec:world_expansion}

\begin{figure}
    \centering
    \includegraphics[width=1.0\linewidth]{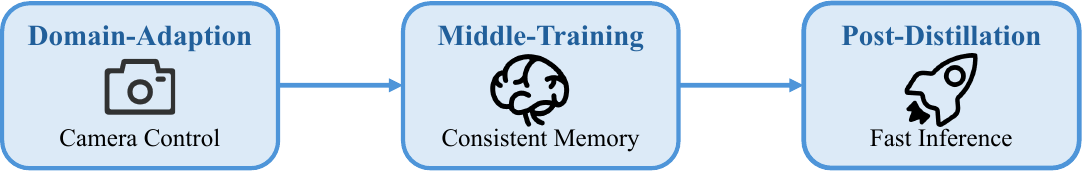}
    \caption{\textbf{Three training stages of WorldStereo 2.0}, progressively enabling camera control, memory-based consistency, and fast inference.
    \label{fig:worldexpand_training_stage}}
\end{figure}

\begin{figure}
    \centering
    \includegraphics[width=1.0\linewidth]{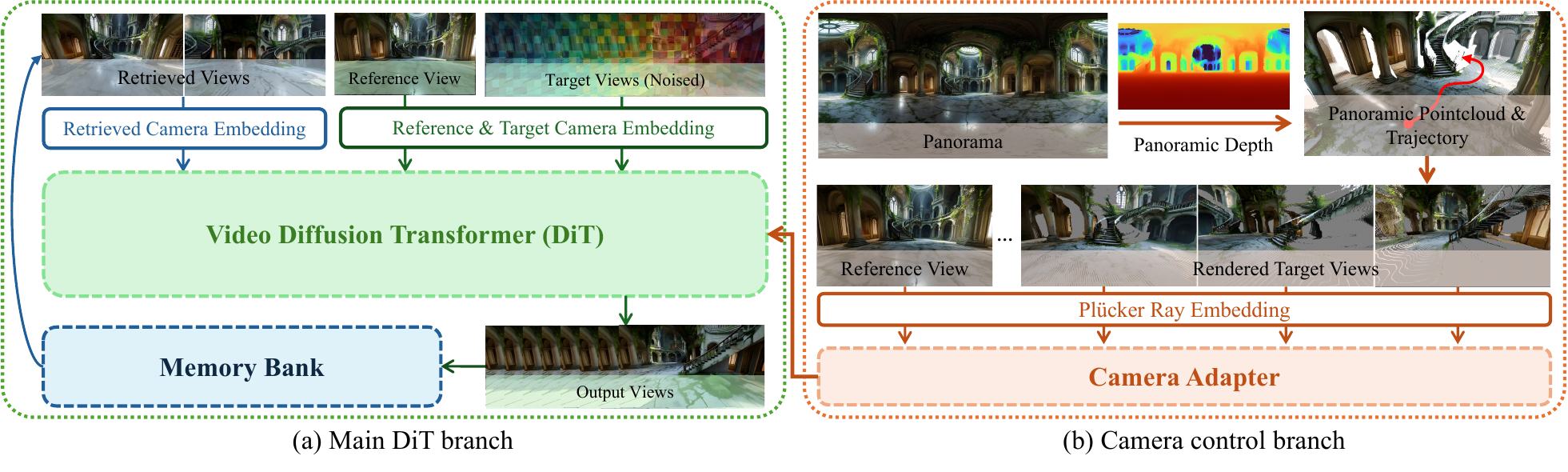}
    \caption{\textbf{Overall pipeline of WorldStereo 2.0.} (a) The main Video Diffusion Transformer (DiT) branch is enhanced by the retrieval-based improved Spatial-Stereo Memory (SSM++) for fine-grained consistency. (b) The camera control branch is guided by the panoramic point cloud, serving as Global-Geometric Memory (GGM) to confirm precise camera trajectory following and geometry-aware consistency. Here, we omit the VAE encoding/decoding for simplicity.
    \label{fig:worldexpand_overview}}
\end{figure}

\paragraph{Task Formulation.}
Building upon the high-quality panoramas (\Cref{sec:panorama_gen}) and broad-coverage camera trajectories (\Cref{sec:traj_plan}), we propose \textbf{WorldStereo 2.0}.
As an upgrade to WorldStereo 1.0~\cite{worldstereo2026}, it leverages camera-guided video generation to synthesize extensive novel views for world expansion.
As shown in \Cref{fig:worldexpand_training_stage}, the training process consists of three stages, designed to enable camera control, memory-based consistency, and efficient inference, respectively.

\paragraph{Overview of WorldStereo 2.0.}
WorldStereo 2.0 bridges camera-conditioned Video Diffusion Models (VDMs) and 3D scene reconstruction by enabling \emph{consistent multi-trajectory video generations with geometry-aware memories in the keyframe latent space}, as summarized in \Cref{tab:vdm_comparison} and visualized in \Cref{fig:worldexpand_overview}.
Specifically, we first rethink the limitations of the standard Video-VAE in \Cref{sec:keyframe_cam_control}, whose spatio-temporal compression often leads to artifacts that degrade downstream reconstruction---and instead formulate WorldStereo 2.0 in a keyframe latent space with precise camera control to preserve high-frequency appearance and geometric cues better.
To further ensure coherent expansion across trajectories, it incorporates two complementary memory modules in \Cref{sec:world_expand_memory}: \emph{Global-Geometric Memory} (GGM) that maintains globally consistent coarse scene structure, and \emph{Spatial-Stereo Memory} (SSM) that reinforces local correspondence and fine-grained details. Together, these designs enable visually faithful and geometrically consistent world expansion suitable for subsequent 3D reconstruction.
Finally, we introduce the acceleration of our model (\Cref{sec:dmd}).

\begin{table}
\centering
\caption{\textbf{Different video generation schemes for 3D reconstruction.} Native video diffusion models (VDMs) need to produce long trajectories in a single pass to cover diverse viewpoints as much as possible. Autoregressive (AR) models sequentially generate long videos. WorldStereo 2.0 achieves multiple consistent generations based on a high-fidelity keyframe latent space with complementary viewpoints and memory mechanisms for subsequent reconstruction.\label{tab:vdm_comparison}}
\small
\begin{tabular}{lccc}
\toprule
\textbf{Paradigms} & \textbf{Native VDM} & \textbf{AR} & \textbf{WorldStereo 2.0} \\
\midrule
Receptive Field & Bidirectional & Autoregressive & Bidirectional \\
Trajectory Length/Num & Long/Single & Long/Single & Medium/Multiple \\
Latent Space & Video Clip & Video Clip & Keyframe Image \\
\midrule
Frame Quality & \normal & \bad & \good \\
Frame Redundancy & \bad & \bad & \good \\
Precise Camera Control & \good & \normal & \good \\
Consistency & \normal & \normal & \good \\
Efficiency & \bad & \good & \good \\
\bottomrule
\end{tabular}
\vspace{-0.1in}
\end{table}

\subsection{Domain-Adaption:  Camera-Guided Keyframe Generation}
\label{sec:keyframe_cam_control}

During the phase of domain-adaption training, we tame the VDM into a camera-controlled keyframe generator to follow pre-defined camera trajectories.
We first introduce the keyframe latent space to confirm high-fidelity generation, followed by the explicit camera control with a unified point cloud and camera ray guidance.

\begin{figure}
    \centering
    \includegraphics[width=1.0\linewidth]{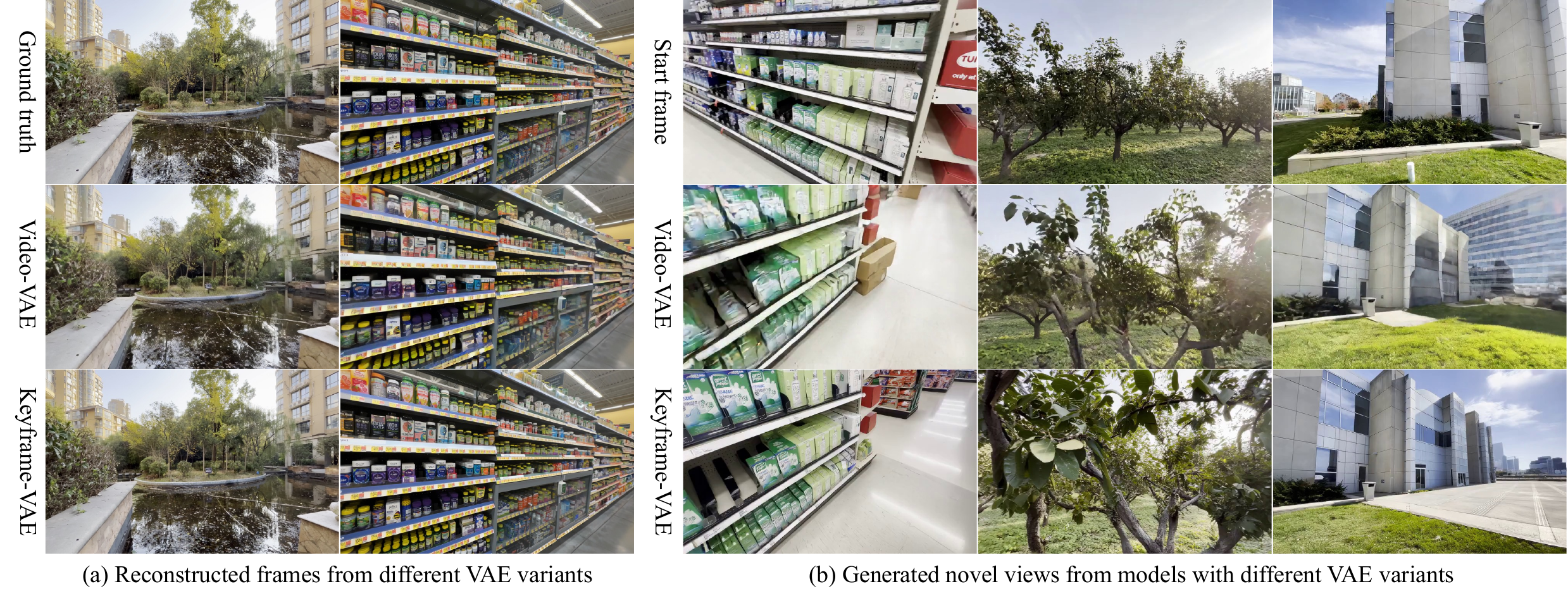}
    \caption{\textbf{Reconstruction and novel-view generation with different VAE variants.} Keyframe-VAE preserves appearance consistency in reconstructions and substantially improves the fidelity of generated novel views, particularly under large viewpoint changes. Please zoom in for details. \label{fig:vae-visual}}
\end{figure}

\begin{figure}
    \centering
    \includegraphics[width=0.95\linewidth]{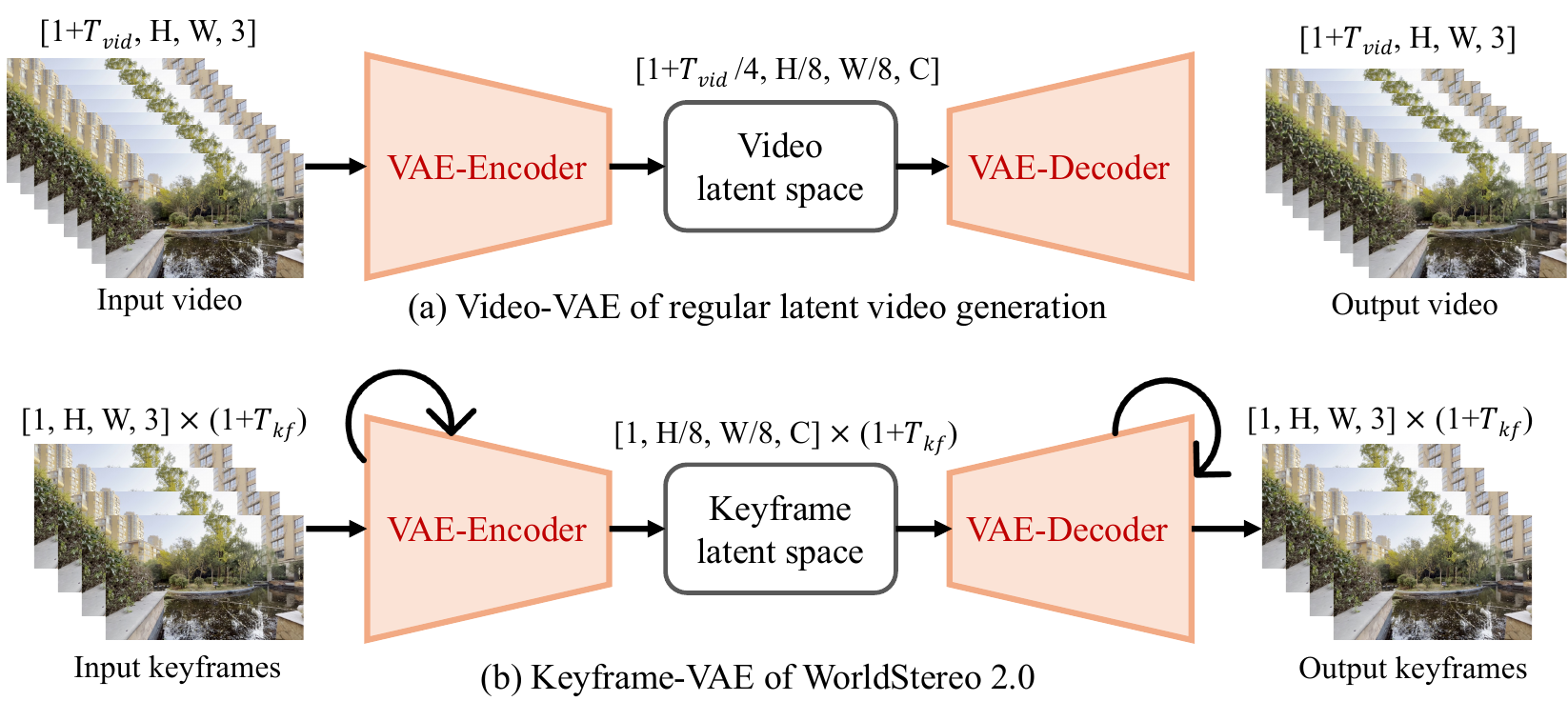}
    \caption{\textbf{Keyframe-VAE in WorldStereo 2.0 versus a standard Video-VAE~\cite{wang2025wan}.} 
    Unlike (a) Video-VAE, which performs spatio-temporal compression, (b) Keyframe-VAE applies \emph{spatial-only} compression to better preserve high-frequency details and reduce artifacts essentially caused by Video-VAE encoding (\eg, motion blur and geometric distortion). Specifically, Keyframe-VAE loops the causal padding-based \emph{image encoding} over $(1+T_{kf})$ times with a sparse frame set ($T_{kf}\ll T_{vid}$) that spans the same viewpoint changes by sampling at larger temporal intervals.
    \label{fig:keyframe-vae}}
\end{figure}

\paragraph{Keyframe-based Spatial Variational Autoencoder.}
\label{sec:keyframe}
Existing camera-guided VDMs often generate redundant frames when camera motion is slow or smooth, thereby failing to satisfy the requirements of \emph{broad} and \emph{diverse} viewpoints for reliable 3D reconstruction.
We attribute this issue largely to a common design choice in latent-based VDMs~\cite{yang2025cogvideox,wang2025wan,hunyuanvideo2025}: videos are compressed by a spatio-temporal Video-VAE.
In such spatio-temporally compressed latent spaces, fast camera motion tends to cause severe quality degradation in both generation and reconstruction, as shown in \Cref{fig:vae-visual}.
Inspired by FlashWorld~\cite{yang2025flash}, we rethink the importance of preserving the latent fidelity and propose to perform scene generation in a \emph{keyframe latent space} using Keyframe-VAE (see \Cref{fig:keyframe-vae}(b)).
Formally, given keyframes $\{\mathbf{V}_i\}_{i=1}^{1+T_{kf}}\in\mathbb{R}^{1\times H\times W\times 3}$, we apply the causal-padding image encoder independently to each keyframe to obtain latent features $\{\mathbf{F}_i\}_{i=1}^{1+T_{kf}}\in\mathbb{R}^{1\times \frac{H}{8}\times \frac{W}{8}\times C}$ for training WorldStereo 2.0, where $H,W,C$ indicate frame height, width, and latent feature channel, respectively. 
Thanks to the high-fidelity image preservation of most open-released Video-VAEs~\cite{yang2025cogvideox,kong2024hunyuanvideo,wang2025wan,hunyuanvideo2025}\footnote{Most Video-VAEs separately encode the first frame as image encoding via causal zero-padding to preserve high-fidelity information for image-to-video generation.}, we can directly inherit their model architectures by treating each keyframe as an image, \ie, applying iterative spatial compression without temporal compression.
A potential concern is that, for the same token length, keyframe latents contain fewer frames than standard video latents ($T_{kf}\ll T_{vid}$), which may reduce the viewpoint coverage available for camera control.
Empirically, we increase the keyframe sampling interval to maintain the same viewpoint coverages, while Keyframe-VAE achieves superior fidelity with comparable camera controllability as verified in \Cref{fig:vae-visual}(b) and \Cref{tab:camera_control_ablation}. 
Furthermore, we claimed that most discarded video frames are visually repetitive and thus largely \emph{redundant} for the subsequent reconstruction stage.
Additionally, the independent property of Keyframe-VAE enables good parallelizability, thereby largely strengthening both VAE encoding and decoding.

\paragraph{Explicit Camera Control.}
Following~\cite{cao2025uni3c,worldstereo2026}, WorldStereo 2.0 is built upon the pre-trained video DiT and integrated with a lightweight transformer-based camera adapter trained from scratch, as shown in \Cref{fig:worldexpand_overview}(b).
Formally, WorldStereo 2.0 incorporates both camera Plücker rays~\cite{sitzmann2021light} and point clouds as complementary camera guidance to enable explicit and precise camera control for subsequent 3D reconstruction.
In the domain-adaption, we only use the point cloud $\mathbf{P}^{ref}\in\mathbb{R}^{N\times 3}$ extracted from the reference view $\mathbf{I}^{ref}\in\mathbb{R}^{H\times W\times 3}$ ($N\leq HW$, after filtering floaters), instead of the panoramic point cloud.
We warp it into each target view to obtain $\{\mathbf{P}^{tar}_{i}\}_{i=1}^{T_{kf}}$, indicated as:
\begin{equation}
\mathbf{P}^{tar}_i(x) \simeq \mathbf{R}^{c \rightarrow w}_i \mathrm{D}(x) \mathbf{K}^{-1}_i \hat{x},
\label{eq_pcd}
\end{equation}
where $\mathbf{R}^{c \rightarrow w}_i$ and $\mathbf{K}_i$ denote the camera-to-world and intrinsic matrices of target view $i$; $\mathrm{D}(\cdot)$ is the monocular depth~\cite{wang2025moge} estimated on the reference view at pixel $x$, and $\hat{x}$ is the homogeneous pixel coordinate.
We then render the warped point clouds into view-wise keyframes~\cite{ravi2020pytorch3d} and encode them into latent features using the Keyframe-VAE.
Compared with Uni3C, which trains only the control branch, we also fine-tune a subset of the Diffusion Transformer (DiT) backbone~\cite{peebles2023scalable} to match the keyframe latent space better. 
Specifically, we freeze the cross-attention and feed-forward layers during the domain-adaption stage, which gives the best trade-off between performance and generalization in our ablations (see \Cref{tab:camera_control_ablation}).

\subsection{Middle-Training: Memory Mechanism}
\label{sec:world_expand_memory}

In the middle-training stage, we adapt the global-geometric and spatial-stereo memory mechanisms proposed in~\cite{worldstereo2026}, tailoring them for panoramic scenarios and the keyframe-based VDM to ensure frame consistency across diverse trajectories.

\begin{figure}
    \centering
    \includegraphics[width=1.0\linewidth]{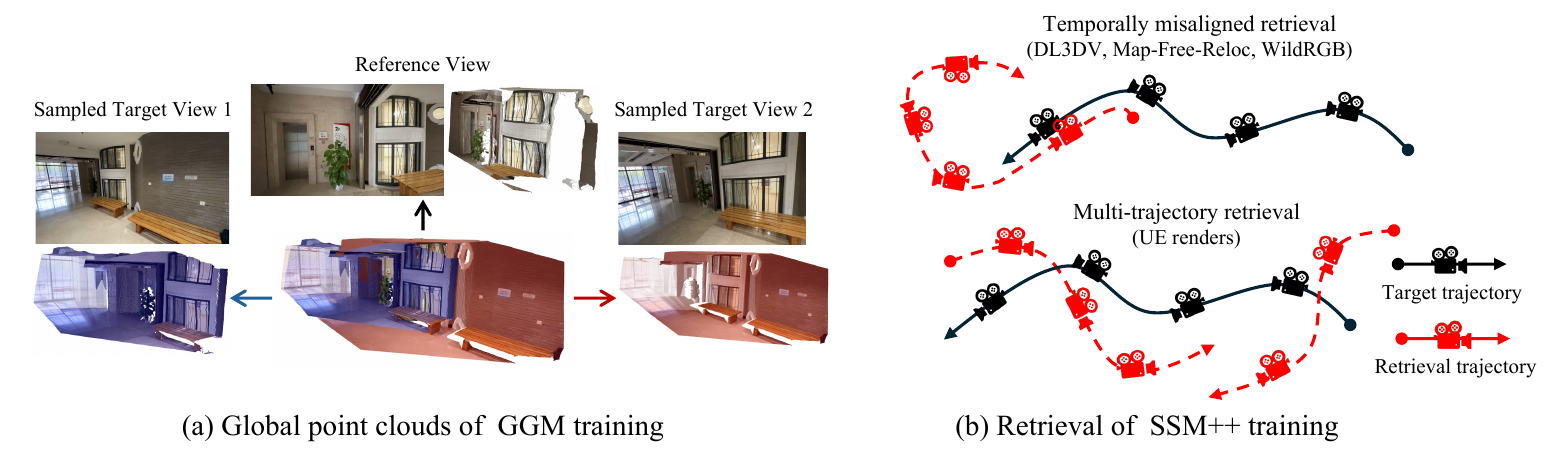}
    \caption{\textbf{Data construction of WorldStereo 2.0 memory training.} (a) The global point clouds built for the GGM training with one reference view and $T_g=2$ target views. (b) Trajectory retrieval strategies for SSM++ training tailored to dataset characteristics: temporally misaligned retrieval for existing multi-view data (top), and multi-trajectory retrieval for synthetic data (bottom).
    \label{fig:training_memory}}
\end{figure}

\subsubsection{Global-Geometric Memory}
\label{sec:ggm}

Global-Geometric Memory (\textbf{GGM}) renders extended point clouds into videos as global 3D priors to generate multiple consistent videos, as illustrated in \Cref{fig:worldexpand_overview}(b).
Particularly in panoramic scenes, GGM allows WorldStereo 2.0 to internalize 360$^{\circ}$ environmental structures, significantly improving geometric consistency.
Although point clouds have been used for camera control in WorldStereo 2.0, they previously served merely as \emph{soft camera guidance} rather than forcing the VDM to strictly adhere to these 3D representations~\cite{cao2025uni3c}.
While this behavior is beneficial for preserving the generalization of camera-guided VDMs against degradation caused by inferior monocular depth, it leads the model to ignore most geometric structures in the point clouds, even when the point clouds are perfectly reconstructed.
To overcome this, we fine-tune the WorldStereo 2.0 using videos rendered by extended global point clouds $\mathbf{P}^{glo}$ beyond the reference points $\mathbf{P}^{ref}\in\mathbb{R}^{N\times 3}$ as:
\begin{equation}
\mathbf{P}^{glo}=[\mathbf{P}^{ref}, \hat{\mathbf{P}}]\in\mathbb{R}^{(N+\hat{N})\times 3},
\label{eq_global_pcd}
\end{equation}
where $\hat{\mathbf{P}}\in\mathbb{R}^{\hat{N}\times 3}$ denotes the additional point clouds randomly sampled from $T_g$ novel views, as shown in \Cref{fig:training_memory}(a).
Furthermore, to prevent overfitting to the point clouds from novel views during training, we employ robust augmentation strategies, as detailed in \Cref{sec:mem_aug}.
For inference, we define the panoramic point cloud $\mathbf{P}^{pan}$ from \Cref{sec:geo_init} as the global point cloud, which covers 360$^{\circ}$ viewpoints' information as effective geometric guidance.

\subsubsection{Improved Spatial-Stereo Memory}
\label{sec:ssm++}

\begin{figure}
    \centering
    \includegraphics[width=0.75\linewidth]{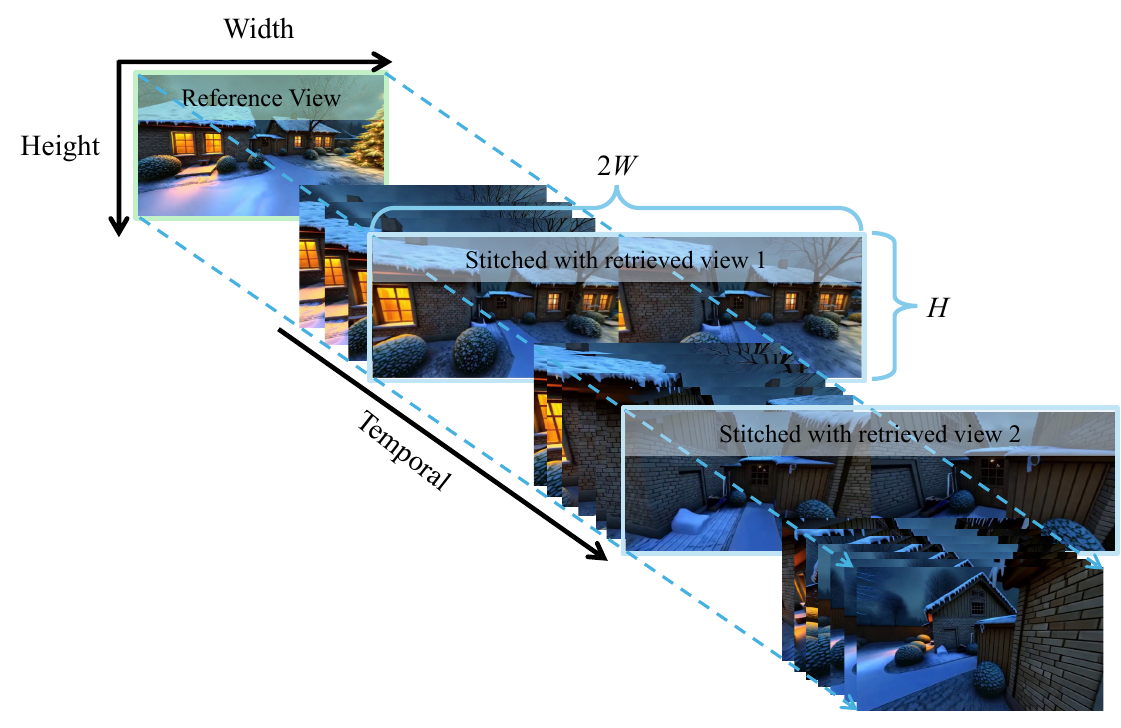}
    \caption{\textbf{Illustration of the RoPE~\cite{su2024roformer} modification in SSM++.} Target frames are spatially concatenated with their corresponding retrieved reference views along the horizontal axis (resulting in width $2W$). Crucially, each retrieved view inherits the temporal index of its paired target frame before being fed into the main DiT branch.
    \label{fig:ssm}}
\end{figure}

While GGM maintains global structural coherence using point clouds, it often struggles to preserve fine-grained details and is prone to accumulating errors.
Many previous studies~\cite{yu2025context,zhou2025stable,schneider2025worldexplorer,li2025vmem} retrieve historical reference frames and jointly model all frames via full-attention. 
However, we cannot guarantee the continuity of retrieved frames (\eg, panoramic scenarios). These disparate, unordered reference views further hinder the VDM learning process.
To overcome these issues, WorldStereo~\cite{worldstereo2026} draws inspiration from the traditional stereo matching~\cite{marr1976cooperative} and the reference-based inpainting~\cite{cao2024leftrefill} and proposes the Spatial-Stereo Memory (\textbf{SSM}), which discretely retrieves reference views and spatially stitches each with its corresponding target view. By constraining the attention receptive field to each retrieval-target pair and utilizing pointmap guidance, SSM effectively recovers details by establishing correspondence within the stitched pairs.

In the WorldStereo 2.0, we advance this design with \textbf{SSM++}, retaining the core concept of horizontal retrieval stitching while introducing significant improvements.
First, we discard the separate memory branch used in WorldStereo and instead directly incorporate retrieved keyframes into the main DiT branch (\Cref{fig:worldexpand_overview}a).
Second, as illustrated in \Cref{fig:ssm}, we modify the Rotary Positional Embedding (RoPE)~\cite{su2024roformer} to accommodate this integration.
Each target view is horizontally stitched with its retrieved counterpart, sharing the same temporal index. Unlike WorldStereo, which enforces a retrieval for every view, SSM++ selectively retrieves only the most relevant keyframes from the memory bank. This selective strategy significantly reduces redundant computation and memory overhead. 
Third, we transition from restricted attention to a full fine-tuning strategy. During the mid-training stage, we remove the constraints on attention receptive fields (except for cross-attention layers), enabling the model to learn global context across all target and retrieved features via full self-attention.
Finally, to enhance flexibility, we replace the explicit pointmap guidance of WorldStereo with implicit camera embeddings. Formally, we normalize all input camera poses to a unified world coordinate and represent them as 7-dimensional vectors (quaternion and translation). These vectors are then encoded by a 3-layer MLP into camera tokens, which are added to the target and retrieved keyframe features via zero-initialization to provide geometry-aware perception.

\paragraph{Memory Bank and Retrieval Strategies.}
We adopt distinct retrieval strategies during the mid-training stage to accommodate varying data properties, as illustrated in \Cref{fig:training_memory}(b).
Practically, training SSM++ requires multi-view videos to construct target-retrieval pairs, which is difficult to obtain in practice. 
To address this, we employ the \emph{temporally misaligned retrieval} to existing multi-view data~\cite {ling2024dl3dv,xia2024rgbd,tartanair2020iros,arnold2022map}.
Specifically, we randomly select frames from the retrieval trajectories with a specified temporal overlap (30\% to 90\%) relative to the target frames. Consequently, unlike simple interpolation, this strategy introduces several retrieved frames that lie outside the target trajectory, thereby increasing training difficulty and enhancing model robustness.
Additionally, we construct a synthetic dataset using UE, featuring multiple trajectories for each asset. For this synthetic data, we employ \emph{multi-trajectory retrieval}, which selects the most relevant frames from alternative trajectories based on 3D FoV similarity~\cite{worldstereo2026}. Furthermore, we apply data augmentation to the retrieved frames to further strengthen SSM++ generalization.
Furthermore, we apply data augmentation to the retrieved frames to further strengthen SSM++ generalization, as detailed in \Cref{sec:mem_aug}.

During inference, perspective views subdivided from the input panorama serve as the initial memory bank. Subsequently, the memory bank is incrementally updated with generated keyframes, storing both RGB images and camera parameters for 3D-FoV retrieval. 
To reduce the computational overhead for each video clip's generation, we limit the retrieval to a maximum of $T_r$ keyframes ($T_r < T_{kf}$).

\subsubsection{Memory Augmentation}
\label{sec:mem_aug}

To mitigate the potential error accumulation stemming from imperfect point clouds and the retrieved generation, we employ comprehensive data augmentations during the middle-training stage to improve the robustness of memory components.

For GGM, we employ specific strategies to degrade the training depth, thereby simulating the inference imperfections: 1) We apply bilinear interpolation to downsample 50\% of the depth maps, simulating the ``depth bleeding'' artifacts; 2) For 10\% of the samples, we apply a small Gaussian filter to the depth maps to create artificial floaters~\cite{yang2026neoverse}; 3) We retain the raw monocular aligned depth for 50\% of the real-world dataset samples~\cite{ling2024dl3dv,arnold2022map,xia2024rgbd} without any post-filtering, preserving native floaters and noises.
Notably, we empirically find that aggressive degradation strategies, like point cloud distortion used in~\cite{yang2026neoverse}, are not suitable for GGM.
Such strong augmentations excessively weaken the geometric guidance provided by the point clouds, resulting in inconsistent geometry across multiple generated videos.
For SSM++, we randomly perform the motion blur and color jitter on the retrieved frames. Moreover, we randomly crop the target and retrieved images to simulate varying visibility ranges and FoV overlaps in the inference scenarios.

\subsection{Post-Train: Model Distillation}
\label{sec:dmd}

During the post-distillation, we apply the modified Distribution Matching Distillation (DMD)~\cite{yin2024improved} to accelerate the inference of WorldStereo 2.0. DMD extends the idea of Variational Score Distillation (VSD)~\cite{wang2023prolificdreamer}, distilling a few-step diffusion student $G_\theta$ through the approximate Kullback-Liebler (KL) divergence built from the difference between the frozen real score function $s_{real}$ and the trainable fake score function $s_{fake}$.
The updating gradient of DMD can be written as:
\begin{equation}
\nabla \mathcal{L}_{\text{DMD}} = -\mathbb{E}_t \left( \int \left( s_{\text{real}}(x_t, t) - s_{\text{fake}}(x_t, t) \right) \frac{dx_t}{d\theta} dz \right),
\label{eq:dmd}
\end{equation}
where $x=G_\theta(z)$ denotes the student generation given random Gaussian noise $z\sim\mathcal{N}(0;\mathbf{I})$ and $t\sim\mathcal{U}(0,1)$, while $x_t\sim q_t(x_t|x,t)$ indicates the forward diffusion process.

The generator $G_\theta$ of WorldStereo 2.0 is distilled into a 4-step DiT.
$G_\theta$, $s_{real}$, $s_{fake}$ are all initialized from the same VDM after the middle-training phase: $s_{real}$ is frozen, while $G_\theta$ and $s_{fake}$ are fully trainable. 
Following~\cite{yin2024improved}, we train $s_{fake}$ 5 times per generator update. The stochastic gradient truncation~\cite{huang2025self} is employed to stabilize the training phase. We omit the GAN loss, as we found its impact to be insignificant while substantially slowing down training.
Different from WorldStereo~\cite{worldstereo2026}, which only distilled on the camera control task with a frozen memory branch due to a shortage of annotated memory data (specifically, the requirement for well-aligned depth for memory guidance). 
In contrast, benefiting from the flexible, explicit-guidance-free SSM++ and the abundance of high-quality UE rendering data, WorldStereo 2.0 enables full fine-tuning of the post-distillation within the memory-based training.
Although this choice is slightly more costly, we find that it simultaneously enhances both camera control precision and memory capability.

\begin{figure*}[t]
    \centering
    \includegraphics[width=1.0\linewidth]{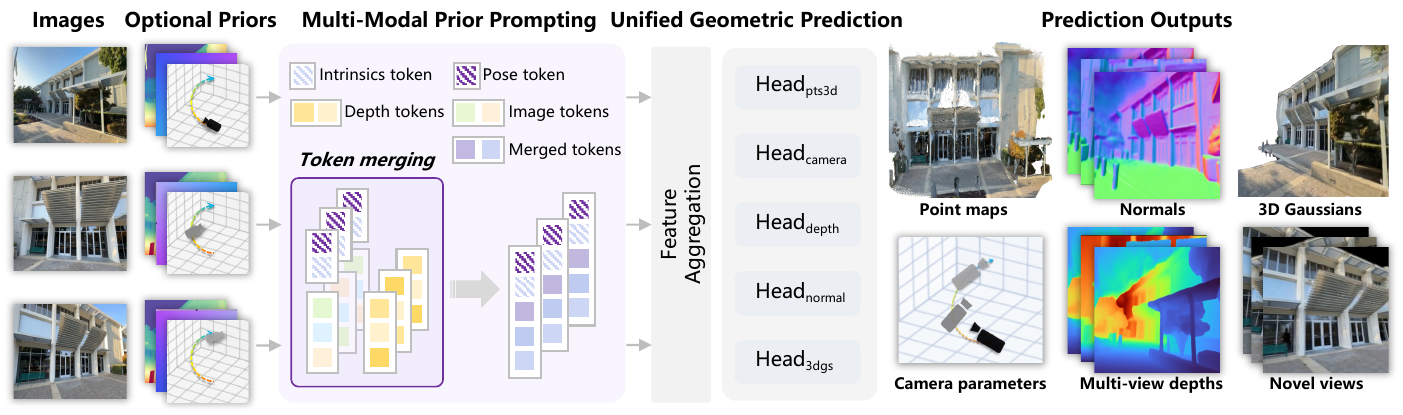}
    \caption{\textbf{Model architecture of WorldMirror 2.0,} which is a unified feed-forward model that takes multi-view images with optional geometric priors (camera poses, intrinsics, depth maps) as input, and simultaneously predicts dense point clouds, depth maps, surface normals, camera parameters, and 3DGS through a shared Transformer backbone with task-specific DPT decoder heads.}
    \label{fig:worldmirror_overview}
\end{figure*}

\section{World Reconstruction: WorldMirror 2.0}
\label{sec:world_recon}

Before detailing the final world composition stage (\Cref{sec:world_composition}), we first introduce our upgraded feed-forward 3D reconstruction model, \textbf{WorldMirror 2.0}, which serves as the crucial bridge between 2D keyframe generation and 3D world composition.
While \emph{world generation} aims to synthesize explorable 3D worlds from sparse inputs (\eg, single-view images or texts), \emph{world reconstruction} focuses on recovering geometrically accurate 3D spatial relationships from dense 2D visual observations (\ie, multi-view images or videos). 
In HY-World 2.0, we build this reconstruction capability upon WorldMirror~\cite{liu2025worldmirror}, a unified feed-forward model for comprehensive 3D geometric prediction. We address three key limitations of WorldMirror~1.0: (1)~degraded performance at non-training resolutions, (2)~limited depth geometric consistency due to the lack of explicit depth--normal coupling, and (3)~prohibitive memory and latency when scaling to large numbers of views. These are tackled through improvements in model architecture (\Cref{sec:model_improvements}), training data and supervision, and training strategy (\Cref{sec:training_improvements}), respectively. Consequently, WorldMirror 2.0 not only functions as a powerful standalone reconstruction foundation but also acts as the core geometry extractor for the generated views in our pipeline. \Cref{fig:worldmirror_overview} illustrates the overall model architecture, and \Cref{tab:wm_comparison} summarizes the key differences between WorldMirror~1.0 and WorldMirror~2.0.

\subsection{Revisiting WorldMirror 1.0}
\label{sec:worldmirror_revisit}

WorldMirror~\cite{liu2025worldmirror} is a unified feed-forward model for comprehensive 3D geometric prediction (see \Cref{fig:worldmirror_overview}). A core design is \emph{Any-Modal Tokenization}, which encodes all input modalities, including images, camera poses, intrinsics, and depth maps, as tokens within a unified sequence. During training, each prior modality is independently dropped with probability 0.5, enabling flexible control over input modalities at inference time. These tokens are jointly processed by a Transformer backbone with global-local attention mechanisms and decoded by multiple DPT heads~\cite{wang2025vggt} to produce 3D point maps, multi-view depth maps, surface normals, camera parameters, and pixel-wise 3D Gaussian Splatting attributes in a single forward pass. To decouple geometry learning from appearance modeling, WorldMirror employs a two-phase curriculum: geometry heads (point map, depth, camera, normal) are jointly trained in the first phase, then all geometry parameters are frozen while only the 3D Gaussian head is trained in the second phase.

\begin{table}[t]
\centering
\caption{\textbf{Comparison between WorldMirror~1.0 and WorldMirror~2.0.} Improvements are organized by model architecture (\Cref{sec:model_improvements}), training data, and training strategy (\Cref{sec:training_improvements}). The bottom section summarizes the resulting capability gains.}
\label{tab:wm_comparison}
\small
\setlength{\tabcolsep}{3pt}
\begin{tabular}{p{4.8cm}cc}
\toprule
\textbf{Component} & \textbf{WorldMirror 1.0} & \textbf{WorldMirror 2.0} \\
\midrule
\multicolumn{3}{l}{\cellcolor{catgray}\textit{Model Improvements (\Cref{sec:model_improvements})}} \\
Position Encoding & Absolute RoPE & Normalized RoPE \\
Depth Supervision & GT depth only & GT depth + GT/Pseudo normal \\
Invalid Pixel Modeling & Confidence only & Confidence + Depth mask head \\
Acceleration & None & Token/Frame SP + BF16 + FSDP \\
\midrule
\multicolumn{3}{l}{\cellcolor{catgray}\textit{Data Improvements (\Cref{sec:data_improvements})}} \\
 Data & Open-sourced & + Internal UE renderings \\
Pseudo-Label Enhancement & \xmark & Normal pseudo-labels\\
\midrule
\multicolumn{3}{l}{\cellcolor{catgray}\textit{Training Strategy (\Cref{sec:training_improvements})}} \\
Image Res./Num. Sampling & Independent & Token-budget dynamic \\
Curriculum Stages & 2 stages & 3 stages \\
Resolution Sampling & 100K--250K pixels & 50K--500K pixels \\
\midrule
\multicolumn{3}{l}{\cellcolor{catgray}\textit{Resulting Capabilities}} \\
Flexible Resolution Inference & \bad & \good \\
Depth Geometric Consistency & \bad & \good \\
Robust Invalid Pixel Handling & \normal & \good \\
Training Efficiency & \normal & \good \\
Inference Efficiency & \bad & \good \\
Overall Reconstruction Quality & \normal & \good \\
\bottomrule
\end{tabular}
\end{table}

\subsection{Model Improvements}
\label{sec:model_improvements}

As summarized in \Cref{tab:wm_comparison}, we introduce three key model-level improvements in WorldMirror~2.0: normalized position encoding for flexible resolution inference, explicit normal-based supervision for depth via a depth-to-normal loss, and a dedicated depth mask prediction head for robust handling of invalid pixels. We further describe data improvements (\Cref{sec:data_improvements}), inference efficiency optimizations (\Cref{sec:inference_efficiency}), and training strategy improvements (\Cref{sec:training_improvements}) in subsequent subsections.

\subsubsection{Normalized Position Encoding}
\label{sec:normalized_rope}

\paragraph{Motivation.}
WorldMirror~1.0 adopts the standard RoPE~\cite{su2024roformer} to inject 2D spatial awareness into multi-head self-attention. Each patch is assigned its absolute integer grid index $(i, j) \in \{0,\dots,H_p{-}1\}\times\{0,\dots,W_p{-}1\}$, which is used to compute position-dependent rotation angles. While effective at a fixed resolution, this scheme introduces a fundamental limitation for multi-resolution inference: when the test resolution exceeds the training resolution, a significant portion of patch indices fall outside the range observed during training (\emph{i.e.}, position extrapolation), leading to degraded predictions. Conversely, when the test resolution is lower, the index space is under-utilized, causing a distribution shift in the attention pattern.

\paragraph{Method.}
Inspired by DINOv3~\cite{simeoni2025dinov3}, we replace the absolute integer coordinates with normalized coordinates that map all patch positions into a fixed $[-1, 1]$ range regardless of the input resolution. Specifically, given an input image with a patch grid of size $H_p \times W_p$ (where $H_p = H/p$ and $W_p = W/p$ for patch size $p$), we compute the normalized coordinates for each patch $(i, j)$ as:
\begin{equation}
    \hat{x}_{i} = \frac{2i+1}{H_p} - 1, \quad \hat{y}_{j} = \frac{2j+1}{W_p} - 1,
\end{equation}
where $\hat{x}_{i}, \hat{y}_{j} \in [-1, 1]$. The $+1$ offset in the numerator ensures pixel-center alignment, preventing boundary patches from collapsing onto $\pm 1$. We normalize the height and width dimensions independently, which preserves aspect ratio information and generalizes better to non-square inputs. These normalized coordinates are then fed into the standard 2D RoPE computation to produce position-dependent rotations for each query and key token in attention.

\begin{figure}[t]
    \centering
    \includegraphics[width=\linewidth]{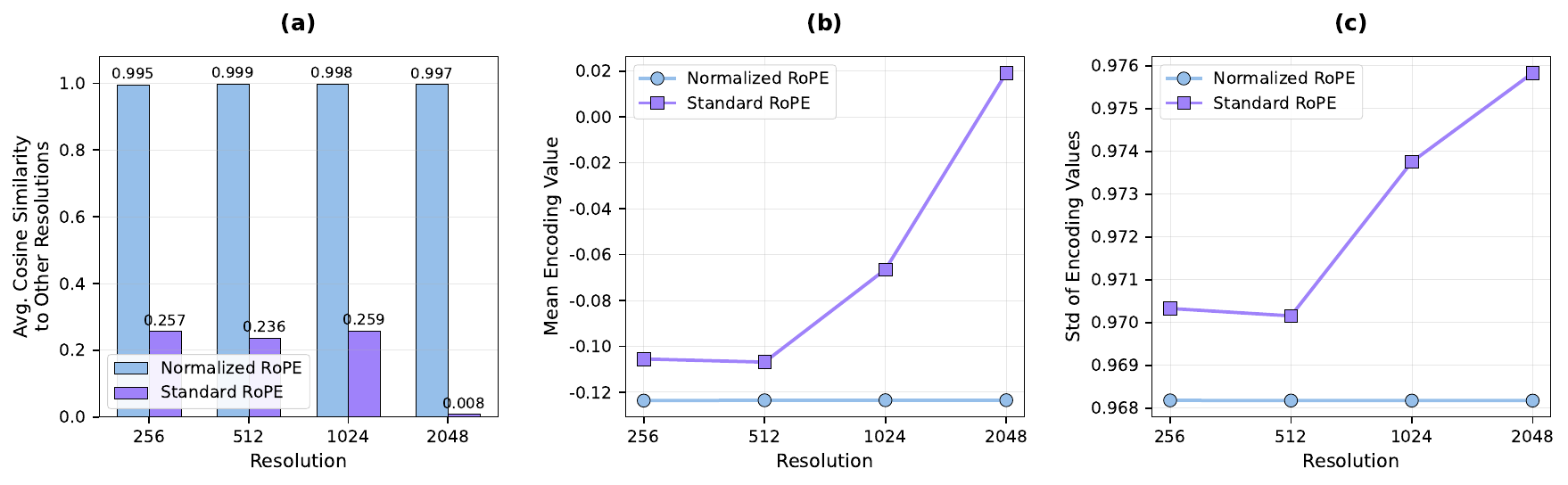}
    \caption{\textbf{Analysis of normalized position encoding.} \textbf{(a)}~Average cosine similarity of center-point RoPE encodings to other resolutions. Normalized RoPE maintains high cross-resolution consistency ($>0.95$), while standard RoPE degrades significantly. \textbf{(b)} and \textbf{(c)} show the mean and standard deviation of encoding values across resolutions, respectively. Normalized RoPE exhibits near-constant statistics, whereas standard RoPE shows systematic mean drift, confirming that normalization converts position extrapolation into interpolation.}
    \label{fig:normalized_rope}
\end{figure}

\paragraph{Analysis.}
The key advantage of normalized position encoding lies in converting resolution extrapolation into interpolation. With standard RoPE, an 8-patch training grid occupies integer indices $[0, 7]$; at inference on a 16-patch grid, indices $[8, 15]$ are entirely out-of-distribution. Normalized RoPE maps both grids into $[-1, 1]$, so inference-time coordinates are simply a denser sampling of the same range. We verify this in \Cref{fig:normalized_rope}: (a) normalized RoPE maintains consistently high cross-resolution cosine similarity ($>0.95$), whereas standard RoPE degrades significantly; (b, c) the mean and standard deviation of encoding values remain stable under normalized RoPE, while standard RoPE exhibits systematic mean drift.

\subsubsection{Explicit Normal Supervision for Depth Estimation}
\label{sec:depth2normal}

\paragraph{Motivation.}
In WorldMirror~1.0, the depth and normal prediction heads are independently supervised without explicit geometric coupling between the two quantities. Moreover, real-world multi-view datasets often contain noisy or incomplete depth annotations, while monocular depth pseudo labels suffer from multi-view inconsistencies. These challenges motivate us to introduce an alternative supervision pathway that explicitly couples depth with normals.

\paragraph{Method.}
We introduce a depth-to-normal loss ($\mathcal{L}_{\text{d2n}}$) that converts predicted depth into surface normals via back-projection and cross products, and supervises the derived normals against normal targets. Specifically, given a predicted depth map $\hat{\mathbf{D}}_i$ and camera intrinsics $\mathbf{K}$, we compute the derived normal $\tilde{\mathbf{N}}_i$ as:
\begin{equation}
    \tilde{\mathbf{N}}_i(x) = \text{normalize}\left(\frac{\partial \mathbf{P}_i}{\partial u} \times \frac{\partial \mathbf{P}_i}{\partial v}\right), \quad \mathbf{P}_i = \mathbf{K}^{-1} \hat{\mathbf{D}}_i \cdot [u, v, 1]^\top,
\end{equation}
where partial derivatives are approximated by finite differences from four quadrant directions and robustly aggregated. The loss is defined as the angular error between the derived and target normals:
\begin{equation}
    \mathcal{L}_{\text{d2n}} = \frac{1}{|\mathcal{V}|} \sum_{x \in \mathcal{V}} \arccos\left(\frac{\tilde{\mathbf{N}}_i(x) \cdot \hat{\mathbf{N}}_i(x)}{\|\tilde{\mathbf{N}}_i(x)\| \|\hat{\mathbf{N}}_i(x)\|}\right),
\end{equation}
where $\hat{\mathbf{N}}_i$ denotes the normal supervision target and $\mathcal{V}$ is the set of valid pixels. The choice of normal target depends on the data source:
\begin{itemize}[nosep,leftmargin=1.5em]
    \item \textbf{Synthetic datasets:} $\hat{\mathbf{N}}_i$ is obtained by applying the same depth-to-normal transform to the ground-truth depth, which provides clean and multi-view consistent supervision.
    \item \textbf{Real-world datasets:} $\hat{\mathbf{N}}_i$ is the pseudo normal predicted by a monocular normal estimation teacher model (see \Cref{sec:data_enhancement}), which offers dense and reliable surface orientation supervision without multi-view inconsistency.
\end{itemize}
Through this mechanism, the depth head receives effective geometric supervision from normals on \emph{all} datasets, even those lacking reliable depth ground truth.

\subsubsection{Depth Mask Prediction}
\label{sec:depth_mask}

Real-world depth data frequently contains invalid pixels due to sensor noise, occlusion boundaries, transparent surfaces, and sky regions. WorldMirror~1.0 handles pixel reliability through learned confidence weights that modulate the training loss, but does not produce an explicit per-pixel validity prediction at inference time, forcing downstream applications to rely on heuristic thresholds. To address this, we augment WorldMirror~2.0 with a dedicated depth mask prediction head that outputs a per-pixel validity logit $\hat{m}(x)$, trained with a binary cross-entropy loss:
\begin{equation}
    \mathcal{L}_{\text{mask}} = -\frac{1}{|\mathcal{M}|} \sum_{x \in \mathcal{M}} \left[ m^*(x) \log \sigma(\hat{m}(x)) + (1 - m^*(x)) \log (1 - \sigma(\hat{m}(x))) \right],
\end{equation}
where $m^*(x) \in \{0, 1\}$ denotes the ground-truth validity label and $\mathcal{M}$ is the set of pixels with known validity. For synthetic datasets, ground-truth masks are derived from rendering pipelines where invalid regions are precisely known. For real-world datasets, we generate pseudo labels by identifying pixels with extreme depth values, large depth discontinuities, or sky regions. At inference time, the predicted mask enables downstream applications to selectively filter invalid pixels, improving the robustness of point cloud fusion and 3D reconstruction.

\subsection{Data Improvements}
\label{sec:data_improvements}
We expand the training data of WorldMirror~2.0 with two key additions. First, we incorporate high-quality synthetic renderings from Unreal Engine scenes, which provide pixel-accurate ground-truth geometry in diverse indoor and outdoor environments. Second, we adopt a \emph{normal-only} pseudo-label enhancement strategy for real-world datasets.
\label{sec:data_enhancement}
A natural approach is to use monocular depth estimators to produce pseudo depth labels; however, we observe that independently predicted per-view depths introduce multi-view geometric inconsistencies (visible as point cloud layering artifacts). Surface normals, by contrast, describe local orientation without requiring global metric consistency, making them inherently more robust as pseudo labels in multi-view settings. We therefore employ a monocular normal estimation teacher model to predict dense normals per view and use them as pseudo supervision targets: directly for the normal head via an angular loss, and indirectly for the depth head through the depth-to-normal loss $\mathcal{L}_{\text{d2n}}$ (\Cref{sec:depth2normal}).

\subsection{Inference Efficiency Improvements}
\label{sec:inference_efficiency}

WorldMirror~1.0 runs on a single GPU with FP32 weights, which limits the maximum number of views and resolution at inference time. WorldMirror~2.0 introduces three complementary acceleration strategies to enable scalable multi-GPU deployment. First, we adopt \emph{sequence parallelism} at two granularities: token-level parallelism for the Transformer backbone, where the input token sequence is partitioned across GPUs and redistributed via All-to-All collectives at each attention layer, and \emph{frame-level} parallelism for the DPT decoder heads, whose convolutional layers operate independently on per-view feature maps and do not benefit from token-level partitioning---per-view features are instead redistributed so that each GPU decodes a disjoint subset of complete frames. Second, following VGGT-X~\cite{wang2025vggt}, we apply \emph{selective mixed-precision inference} by casting most parameters to BF16 while keeping a small set of precision-critical modules in FP32, halving the memory footprint with negligible accuracy loss. Third, we employ \emph{fully sharded data parallelism} (FSDP) to shard model parameters across GPUs, with each Transformer block and DPT head wrapped as an independent FSDP unit. These three strategies are complementary: sequence parallelism distributes computation and activation memory, mixed-precision reduces per-element cost, and FSDP shards weight memory. Together, they enable WorldMirror~2.0 to process substantially larger inputs while reducing per-GPU memory consumption and wall-clock time (\Cref{sec:recon_experiments}).

\subsection{Training Strategy Improvements}
\label{sec:training_improvements}

\textbf{Token-based Dynamic Batch Sizing.}
\label{sec:token_batch}
WorldMirror~1.0 independently samples the per-image resolution and the number of views at each training iteration. Since GPU memory must accommodate the worst-case joint maximum (\ie, highest resolution $\times$ maximum view count), this independent sampling strategy leads to substantial GPU memory under-utilization in practice, as most sampled configurations fall well below this ceiling.

We address this with a \emph{token-budget-first} strategy. Concretely, we fix a maximum token budget $T_{\text{max}}$ per GPU (\eg, 25{,}000 tokens). At each iteration, we first sample the per-image resolution (pixel count from a configurable range, \eg, 50K--500K) and aspect ratio, then compute the per-image token count $t = \frac{H}{p} \times \frac{W}{p}$. The maximum number of views is then derived as:
\begin{equation}
    N_{\text{max}} = \min\!\left(N_{\text{cap}},\;\left\lfloor \frac{T_{\text{max}}}{t} \right\rfloor\right),
\end{equation}
where $N_{\text{cap}}$ is the architectural view-count cap (\eg, 48). The actual view count is uniformly sampled from $[N_{\text{min}}, N_{\text{max}}]$. When the sampled view count is smaller than $N_{\text{max}}$, multiple samples are packed into the same GPU to fill the token budget, ensuring the tightly bounded token count for each GPU:
\begin{equation}
    T_{\text{total}} = N \times \frac{H}{p} \times \frac{W}{p} \leq T_{\text{max}},
\end{equation}
where $N$ is the total number of images on one GPU, including multiple samples. This design consistently achieves near-full GPU memory utilization regardless of the sampled resolution, exposes the model to more diverse resolution--view-count combinations during training, and eliminates out-of-memory errors without conservative memory provisioning.

\textbf{Multi-Stage Curriculum Learning.}
\label{sec:multi_stage}
WorldMirror~1.0 employs a two-phase curriculum: geometry heads are jointly trained first, then all geometry parameters are frozen while only the Gaussian head is trained. In WorldMirror~2.0, we further decompose the geometry training into two sub-stages, yielding a three-stage pipeline: \textbf{Stage~1} trains all geometry heads using native annotations without pseudo-label enhancement or the depth-to-normal loss; \textbf{Stage~2} introduces  the depth-to-normal loss (\Cref{sec:depth2normal}), while significantly increasing the proportion of synthetic data to improve geometric precision; \textbf{Stage~3} freezes the backbone and all geometry heads, training only the 3DGS head initialized from the depth head weights.

\section{World Generation Stage IV: World Composition}
\label{sec:world_composition}

\paragraph{Task Formulation.}
We define the input for this stage as a tuple containing the initial panorama $\mathbf{I}^{pan}$ (\Cref{sec:panorama_gen}), its corresponding panoramic point cloud $\mathbf{P}^{pan}$ (\Cref{sec:geo_init}), and the whole set of $T_{ex}$ novel keyframes $\{\mathbf{V}_i\}_{i=1}^{T_{ex}}$ generated from WordExpand (\Cref{sec:world_expansion}) based on pre-defined trajectories $\{\mathbf{C}_i\}_{i=1}^{T_{ex}}$ (\Cref{sec:traj_plan}).
The goal of \textit{World Composition} is to integrate these inputs into a unified, navigable 3D representation. This process consists of two sequential steps:

1) Point cloud expansion (\Cref{sec:pcd_expansion}): constructing a globally aligned point cloud $\mathbf{\tilde{P}}$ by expanding $\mathbf{P}^{pan}$ with generated keyframes.

2) 3D scene optimization (\Cref{sec:3dgs}): training a 3DGS, initialized with the expanded point cloud $\mathbf{\tilde{P}}$, to synthesize the complete high-fidelity 3D world.

\subsection{Point Cloud Expansion}
\label{sec:pcd_expansion}

\begin{figure}
    \centering
    \includegraphics[width=1.0\linewidth]{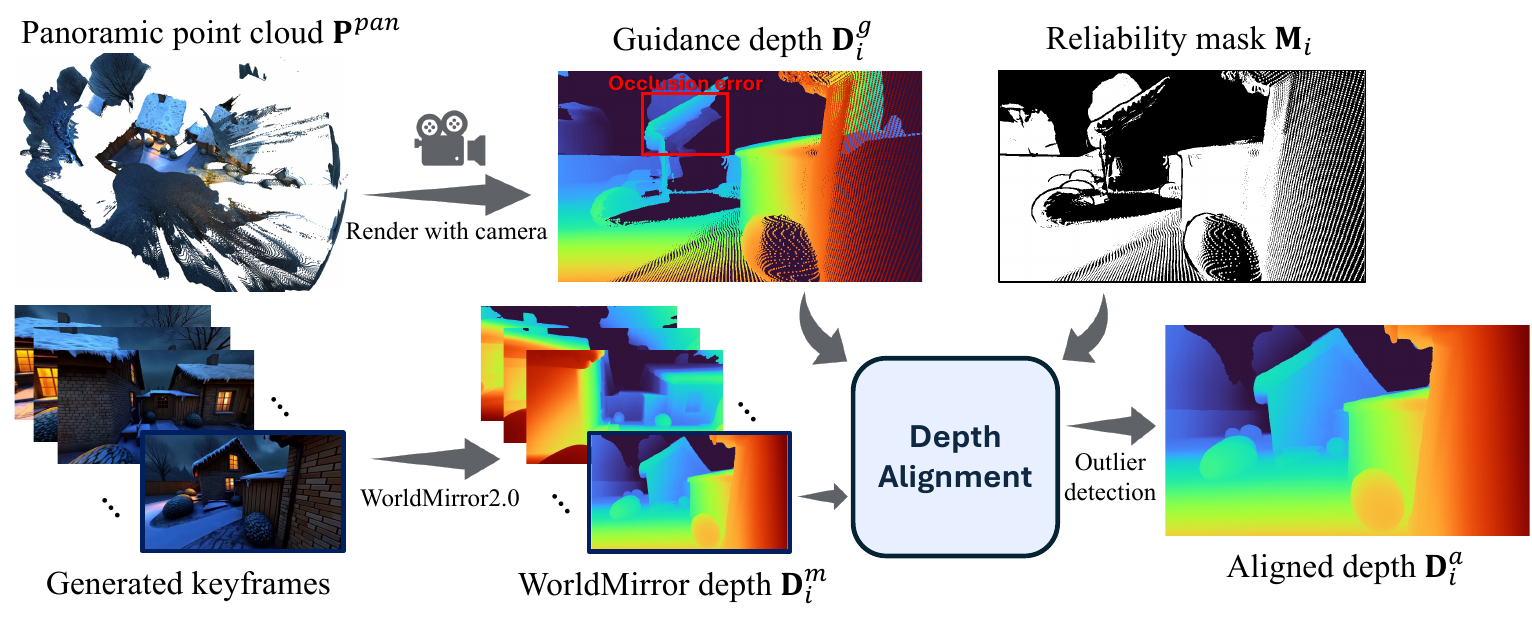}
    \caption{\textbf{The pipeline of depth alignment.} We align the WorldMirror depth estimated from generated keyframes with the panoramic point cloud. The point cloud is rendered as geometric guidance, identifying reliable regions to supervise the linear alignment. The outlier detection process effectively eliminated and corrected the alignment coefficients based on global statistics.
    \label{fig:depth_align}}
\end{figure}

\subsubsection{Reconstruction via WorldMirror 2.0}
\begin{figure}
    \centering
    \includegraphics[width=1.0\linewidth]{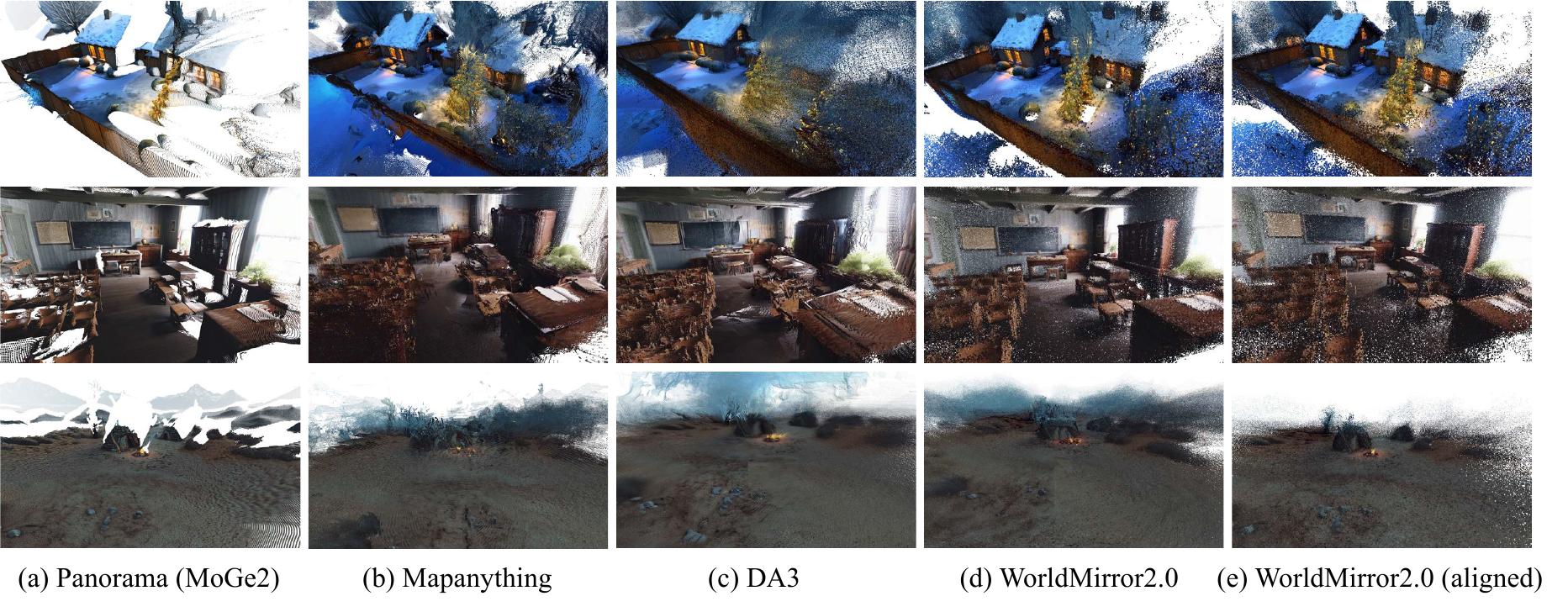}
    \caption{\textbf{Comparison of reconstructed point clouds.} (a) Reference panoramic point cloud estimated by MoGe2~\cite{wang2025moge}.
    (b-d) Results from state-of-the-art feedforward reconstruction methods: Mapanything~\cite{keetha2025mapanything}, DepthAnything3 (DA3)~\cite{lin2025depth}, and WorldMirror 2.0. (e) WorldMirror 2.0 with the proposed depth alignment. Note that for all feedforward methods (b-d), the 10\% of points with the lowest confidence are filtered out, and sky points are manually cropped for better visualization.
    \label{fig:gen_pcd_compare}}
\end{figure}

We employ the state-of-the-art feedforward reconstruction model, WorldMirror 2.0 (a core component of HY-World 2.0, as detailed in \Cref{sec:world_recon}), to reconstruct globally aligned point clouds and depth maps for point cloud expansion, as illustrated in \Cref{fig:depth_align}. 
Specifically, we first downsample a subset of $T'_{ex}$ frames from the fully generated sequence of $T_{ex}$ frames. Subsequently, WorldMirror 2.0 is applied to estimate the per-frame depth and normal maps for this subset, conditioned on their respective camera poses as geometric priors:
\begin{equation}
\{\mathbf{D}^m_i, \mathbf{N}^m_i\}_{i=1}^{T'_{ex}} = \Phi \left( \{\mathbf{V}_j, \mathbf{C}_j\}_{j=1}^{T_{pan}}, \{\mathbf{V}_i, \mathbf{C}_i\}_{i=1}^{T'_{ex}} \right),
\label{eq_wm_align}
\end{equation}
where $\Phi(\cdot)$ denotes the WorldMirror 2.0 network; $\{\mathbf{V}_j, \mathbf{C}_j\}_{j=1}^{T_{pan}}$ represents the perspective views and their corresponding camera parameters subdivided from the initial panorama $\mathbf{I}^{pan}$.
Although WorldMirror 2.0 is not explicitly tailored for panoramic reconstruction, it performs well when combined with our generated keyframe sequences. 
Furthermore, we empirically demonstrate that WorldMirror 2.0 benefits significantly from camera conditions, outperforming other state-of-the-art feedforward reconstruction methods~\cite{keetha2025mapanything,lin2025depth} under identical conditional settings as verified in \Cref{fig:gen_pcd_compare}.

\subsubsection{Depth Alignment}
Although WorldMirror 2.0 generates high-quality depth maps $\{\mathbf{D}^{m}_i\}_{i=1}^{T'_{ex}}$, they suffer from scale ambiguity and fail to align with the world coordinate of the panoramic point cloud $\mathbf{P}^{pan}$.
Moreover, while WorldMirror outperforms other feed-forward reconstruction methods under camera conditions, it still struggles in highly challenging outdoor scenes, as illustrated in the third row of \Cref{fig:gen_pcd_compare}.
Therefore, we propose a robust alignment strategy to rectify WorldMiror depth $\mathbf{D}^{m}_i$ into an aligned depth map $\mathbf{D}^{a}_i$ using the panoramic point cloud $\mathbf{P}^{pan}$ as the geometric guidance.

Formally, we render $\mathbf{P}^{pan}$ from the viewpoint of $\mathbf{C}_i$ to obtain the sparse guidance depth $\mathbf{D}^{g}_i$, as illustrated in \Cref{fig:depth_align}.
The alignment process is formulated as:
\begin{equation}
\mathbf{D}^a_i=\varphi_{align}\left(\mathbf{D}^m_i,\mathbf{D}^g_i,\mathbf{M}_i\right),
\label{eq_alignment}
\end{equation}
where $\mathbf{M}_i$ denotes the reliability mask for view $i$, indicating valid overlapping regions where the alignment should be enforced.
We define $\mathbf{M}_i$ as the intersection across several empirical masks:
\begin{equation}
\mathbf{M}_i=\mathbf{M}^{m}_i\cap\mathbf{M}^{g}_i\cap\mathbf{M}^n_i\cap\mathbf{M}^p_i\cap\overline{\mathbf{M}}^{sky}_i,
\label{eq_valid_mask}
\end{equation}
where $\mathbf{M}^{m}_i$ and $\mathbf{M}^{g}_i$ represent the valid projection regions of the WorldMirror confidence and the panoramic guidance, respectively, with edge floaters removed.
$\mathbf{M}^{n}_i$ enforces normal consistency, excluding regions where the angular deviation between the WorldMirror normal $\mathbf{N}^m_i$ and the derived panoramic normal $\mathbf{N}^g_i$ exceeds 90 degrees.
To mitigate occlusion artifacts shown in the guidance depth of \Cref{fig:depth_align}, we employ a percentile-based statistical filter $\mathbf{M}^{p}_i$ to discard outliers with significant relative depth discrepancies.
Finally, we omit the sky regions using the non-sky mask $\overline{\mathbf{M}}^{sky}_i$ identified by SAM3~\cite{carion2025sam} in video mode.

Subsequently, we perform a \textit{RANSAC-based linear alignment} over the valid regions defined by $\mathbf{M}_i$ to estimate a scale $\gamma_i$ and shift $\beta_i$, yielding a transformation as $\mathbf{D}^a_i=\gamma_i\mathbf{D}^m_i+\beta_i$\footnote{In practice, we apply the alignment in the disparity space instead of the depth space for better foreground alignment. To avoid confusion, we omit the disparity transformation for simplicity.}.
Due to the high-quality initial depth provided by WorldMirror 2.0, we empirically find that per-frame linear alignment is sufficient for our scenarios, thus obviating the need for complex non-linear refinements~\cite{ren2025gen3c,hollein2026world}.
However, erroneous alignment coefficients may still occur, particularly when the valid guidance masks are overly sparse or the camera trajectories are highly challenging. 
To address this, we propose an outlier detection and revision strategy based on the global distribution of the alignment coefficients $\{\gamma_i,\beta_i\}_{i=1}^{T'_{ex}}$. 
Specifically, we set $Q=9$ anchor depth values $\{A_q\}_{q=1}^{Q}$ uniformly distributed across the scene's depth range. For each frame $i$, we compute the transformed anchor values $\mathcal{V}_{i,q}=\gamma_i A_q+\beta_i$.
The maximum relative deviation for each coefficient pair $(\gamma_i,\beta_i)$ is then formulated as:
\begin{equation}
\mathcal{V}_i^{\max}=\max\limits_{q}\left(\left|\frac{\mathcal{V}_{i,q}-\hat{\mathcal{V}}_q}{\hat{\mathcal{V}}_q}\right|\right),\quad
\hat{\mathcal{V}}_q=\mathrm{median}_{j\in\{1,...,T'_{ex}\}}(\mathcal{V}_{j,q}),
\label{eq_max_rel_dev}
\end{equation}
where $\hat{\mathcal{V}}_q$ represents the median transformed value of anchor $q$ across all frames.
Any coefficient pair $(\gamma_i,\beta_i)$ whose maximum relative deviation $\mathcal{V}_{i}^{max}$ exceeds the 90-th percentile is regarded as an outlier.
These outliers are then replaced by the nearest inlier coefficient pairs within the same video sequence.
If an entire sequence is detected as an outlier, we discard all its depth maps.
Finally, upon obtaining the aligned depth maps $\{\mathbf{D}^a_i\}_{i=1}^{T_{ex}'}$,
we back-project them into 3D space to construct the extended point cloud $\mathbf{P}^{ex}$. The union of this extension and the original panorama, $\mathbf{P}^{pan}\cup\mathbf{P}^{ex}$, is then further voxel-downsampled to yield the final expanded point cloud $\mathbf{\tilde{P}}\in\mathbb{R}^{N\times 3}$.

\subsection{3D Gaussian Splatting}
\label{sec:3dgs}

\paragraph{Task Formulation.} 
Given the panorama $\mathbf{I}^{pan}$, expanded point cloud $\mathbf{\tilde{P}}$, and and a set of $T_{ex}$ generated novel keyframes along with their corresponding camera parameters $\{\mathbf{V}_i, \mathbf{C}_i\}_{i=1}^{T_{ex}}$, we optimize a 3DGS model~\citep{kerbl20233d} to serve as the final scene representation.


\paragraph{Initialization.}
We initialize the 3DGS model using the expanded point cloud $\mathbf{\tilde{P}}$. Each 3D Gaussian is parameterized by a set of learnable attributes, including an opacity $\sigma_k \in [0,1]$, a center position $\boldsymbol{\mu}_k \in \mathbb{R}^{3 \times 1}$, and a 3D covariance matrix $\mathbf{\Sigma}_k \in \mathbb{R}^{3 \times 3}$. 
Following \citep{kerbl20233d}, to ensure the covariance matrix remains positive semi-definite during optimization, $\mathbf{\Sigma}_k$ is decomposed into a scaling matrix $\mathbf{S}_k$ and a rotation matrix $\mathbf{R}_k$, formulated as $\mathbf{\Sigma}_k = \mathbf{R}_k \mathbf{S}_k \mathbf{S}_k^{T} \mathbf{R}_k^{T}$. 
Furthermore, we empirically observe that the generated scenes exhibit negligible view-dependent effects. Therefore, instead of employing Spherical Harmonics (SH) for appearance modeling, we adopt view-independent RGB colors $\mathbf{c}_k \in \mathbb{R}^{3}$ as the color features for each Gaussian, reducing both redundancy and complexity.


\paragraph{Growth Strategy and MaskGaussian.}
During the 3DGS training on generated data, we observe a dilemma regarding the adaptive densification mechanism~\cite{kerbl20233d}.
On the one hand, relying solely on the initial point cloud $\mathbf{\tilde{P}}$ without densification leads to a conflict between rendering efficiency and detail preservation.
Specifically, $\mathbf{\tilde{P}}$ exhibits a spatially uneven distribution, over-populating low-frequency regions (\eg, sky and flat surfaces) with redundant Gaussians that degrade real-time rendering efficiency. While applying a uniform voxel downsampling (with a voxel size $v$) can mitigate this redundancy, it severely undermines the reconstruction quality in high-frequency regions, which inherently require denser Gaussian coverage to faithfully capture fine-grained textures. On the other hand, enabling the standard growth strategy, which periodically densifies Gaussians via cloning and splitting based on view-space positional gradients, successfully recovers these high-frequency details but inevitably introduces severe floating artifacts (floaters). These artifacts predominantly originate from sky regions, where the generated depth supervision is unavailable.

To resolve this dilemma and simultaneously achieve rendering efficiency and high-fidelity detail reconstruction, we adopt a two-pronged approach. 
First, we segment the initial point cloud $\mathbf{\tilde{P}}$ into sky and scene subsets, denoted as $\mathbf{\tilde{P}}_{\mathrm{sky}}$ and $\mathbf{\tilde{P}}_{\mathrm{scene}}$, respectively. The standard growth strategy is applied exclusively to $\mathbf{\tilde{P}}_{\mathrm{scene}}$, enabling necessary densification in texture-rich regions while strictly preventing the sky from spawning floaters.
To further eliminate redundant Gaussians in over-populated areas and suppress residual floating artifacts, we integrate MaskGaussian~\citep{liu2025maskgaussian}.
Instead of relying on heuristic hard-pruning, MaskGaussian models the existence of each Gaussian as a probabilistic entity.
Concretely, for the $k$-th Gaussian, a binary mask $M_k \in \{0, 1\}$ is sampled via Gumbel-Softmax~\citep{jang2016categorical} from learnable mask logits. 
This mask is then incorporated into the tile-based rasterizer through a \emph{masked-rasterization} scheme. 
For a given pixel~$\mathbf{x}$, the rendered color $\mathbf{c}(\mathbf{x})$ and transmittance evolution $T_{k+1}$ are reformulated as:
\begin{equation}
  \mathbf{c}(\mathbf{x}) = \sum_{k=1}^{N} M_k \, \mathbf{c}_k \, \sigma_k \, T_k, \quad
  T_{k+1} = T_k ( 1 - M_k \sigma_k ),
  \label{eq:mask_raster}
\end{equation}
where $\sigma_k$ denotes the opacity and $T_k$ is the accumulated transmittance of the $k$-th Gaussian in depth order ($T_1 = 1$). When $M_k = 0$, the Gaussian's color contribution is negligible, and it consumes no transmittance. Crucially, thanks to the Gumbel-Softmax relaxation, it still receives gradients during the backward pass, allowing for a dynamic reassessment of its importance as the scene optimization evolves. To encourage sparsity, a squared loss regularizes the average mask activation:
\begin{equation}
  \mathcal{L}_{\mathrm{mask}} = \lambda_m \left(\frac{1}{N}\sum_{k=1}^{N}M_k\right)^{\!2},
  \label{eq:L_mask}
\end{equation}
which is added to the overall training objective $\mathcal{L}_{\mathrm{GS}}$. During training, Gaussians whose activation probabilities consistently remain near zero are permanently pruned. This adaptive mechanism preferentially eliminates redundant Gaussians in over-populated low-frequency areas while preserving essential primitives in detail-rich regions. Consequently, it simultaneously accelerates rendering speed and suppresses floaters through the implicit regularization induced by probabilistic masking.

\paragraph{Optimization and Losses.}
For the $i$-th training view, the 3DGS renderer produces an RGB image $\hat{\mathbf{I}}_i$ and a depth map $\hat{\mathbf{D}}_i$. The corresponding surface normal $\hat{\mathbf{N}}_i$ is derived analytically as the normalized spatial gradient of $\hat{\mathbf{D}}_i$. The photometric objective is defined as:
\begin{equation}
\mathcal{L}_{\text{color}} = (1 - \lambda_{c1}) \, \mathcal{L}_1(\hat{\mathbf{I}}_i, \mathbf{I}_i) 
+ \lambda_{c1} \, \text{SSIM}(\hat{\mathbf{I}}_i, \mathbf{I}_i) 
+ \lambda_{c2} \, \text{LPIPS}(\hat{\mathbf{I}}_i, \mathbf{I}_i),
\label{eq:L_c}
\end{equation}
where the ground truth images $\mathbf{I}_i$ are sampled from the union of views divided from the panorama and the generated keyframes. Here, $\text{SSIM}$ and $\text{LPIPS}$ denote the structural similarity and perceptual loss~\cite{johnson2016perceptual}, respectively. 
To enforce geometric consistency, we introduce a geometric loss:
\begin{equation}
\mathcal{L}_{\text{geo}} = \lambda_d \, \mathcal{L}_1(\hat{\mathbf{D}}_i, \mathbf{D}^a_i) 
+ \lambda_n \, \big(1 - \text{cos}(\hat{\mathbf{N}}_i, \mathbf{N}_i)\big),
\label{eq:L_g}
\end{equation}
where $\text{cos}(\cdot)$ denotes the pixel-wise cosine similarity. To mitigate computational overhead, depth supervision is applied sparsely, restricted to the partially aligned depth maps $\{\mathbf{D}^a_i\}_{i=1}^{T_{ex}'}$ (\Cref{sec:pcd_expansion}). 
In contrast, the high-quality normal maps $\{\mathbf{N}_i\}_{i=1}^{T_{ex}}$ estimated by MoGe2~\cite{wang2025moge} are inherently alignment-free.
This property enables them to be applied across all frames, providing a dense and robust geometric constraint. 
Furthermore, following~\cite{xie2024physgaussian}, we incorporate a scale regularization term $\mathcal{L}_{\text{reg}}$ to penalize excessively sharp Gaussians, encouraging more isotropic shapes. The overall 3DGS training objective is thus given by:
\begin{equation}
\mathcal{L}_{\text{GS}} = \mathcal{L}_{\text{color}} + \mathcal{L}_{\text{geo}} + \mathcal{L}_{\text{reg}} + \mathcal{L}_{\text{mask}}.
\end{equation}

\paragraph{Mesh Extraction.}
To support downstream applications such as collision detection and physics simulation, we further extract a mesh from the optimized 3DGS representation. Specifically, we render RGB images and depth maps from all training views and integrate them into a Truncated Signed Distance Function (TSDF) volume. The final mesh is extracted via the marching cube algorithm~\cite{lorensen1998marching}. To improve mesh quality, we remove small disconnected components and apply mesh simplification, which effectively suppresses floating artifacts and reduces storage overhead.

\section{Results: Multi-Modal World Creation}
\label{sec:all_exp}

\subsection{World Generation from Text or Single Image}
\label{sec:exp_world_gen}
  
In this section, we comprehensively evaluate the world generation pipeline of HY-World 2.0. We first assess its individual components: panorama generation (\Cref{sec:exp_pano}), trajectory planning (\Cref{sec:exp_traj}), world expansion (\Cref{sec:exp_video_gen}), and 3DGS (\Cref{sec:exp_3dgs}). Then, the final outputs of the integrated system will be showcased in \Cref{sec:exp_overall_world_gen}.

\begin{figure}
    \centering
    \includegraphics[width=1.0\linewidth]{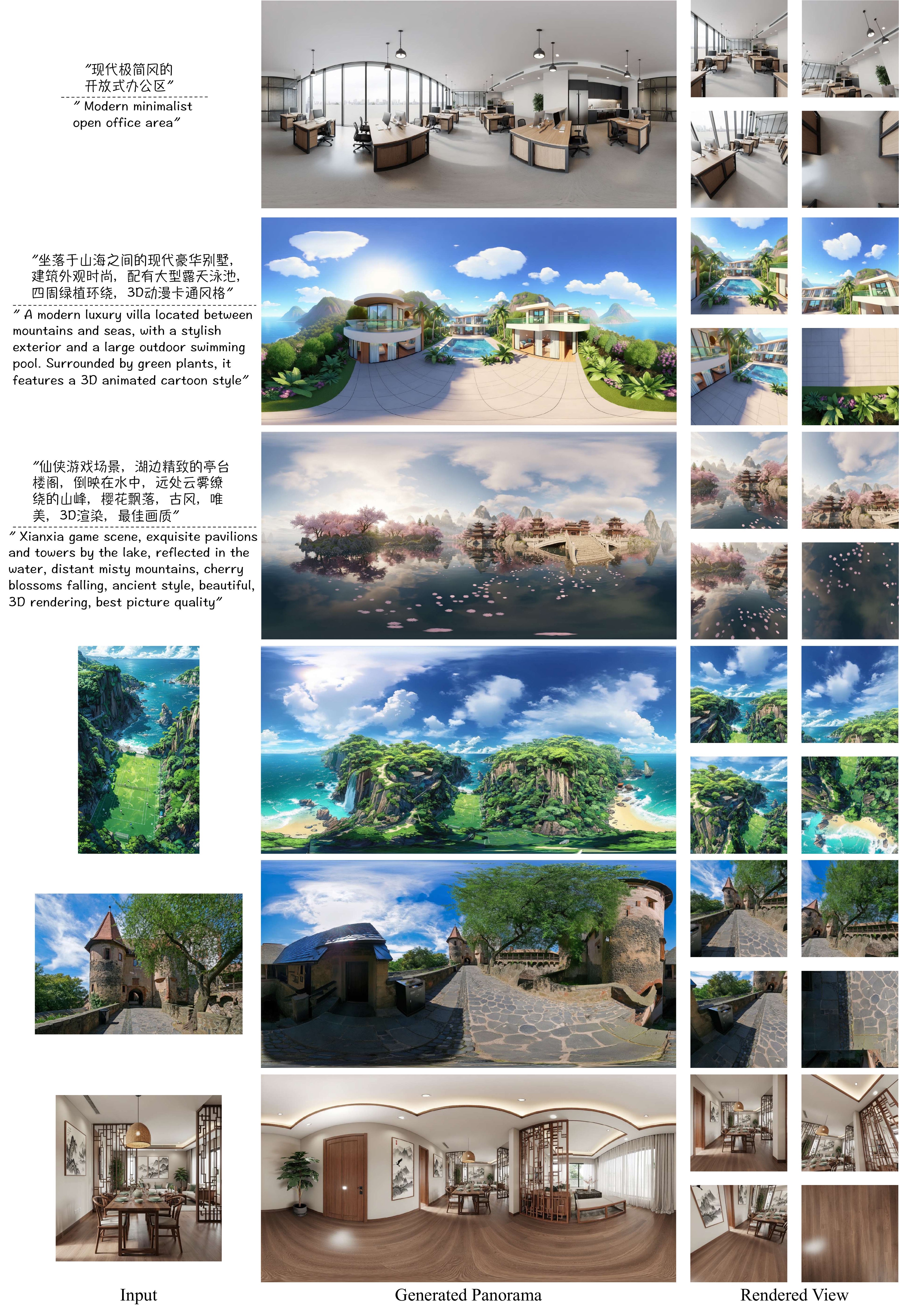}
    \caption{\textbf{Visual results of panorama generation by HY-Pano 2.0.} Our model supports both text and images of various resolutions as inputs.
    \label{fig:pano_results}}
\end{figure}

\begin{figure*}[t]
\centering
\includegraphics[width=1.045\linewidth]{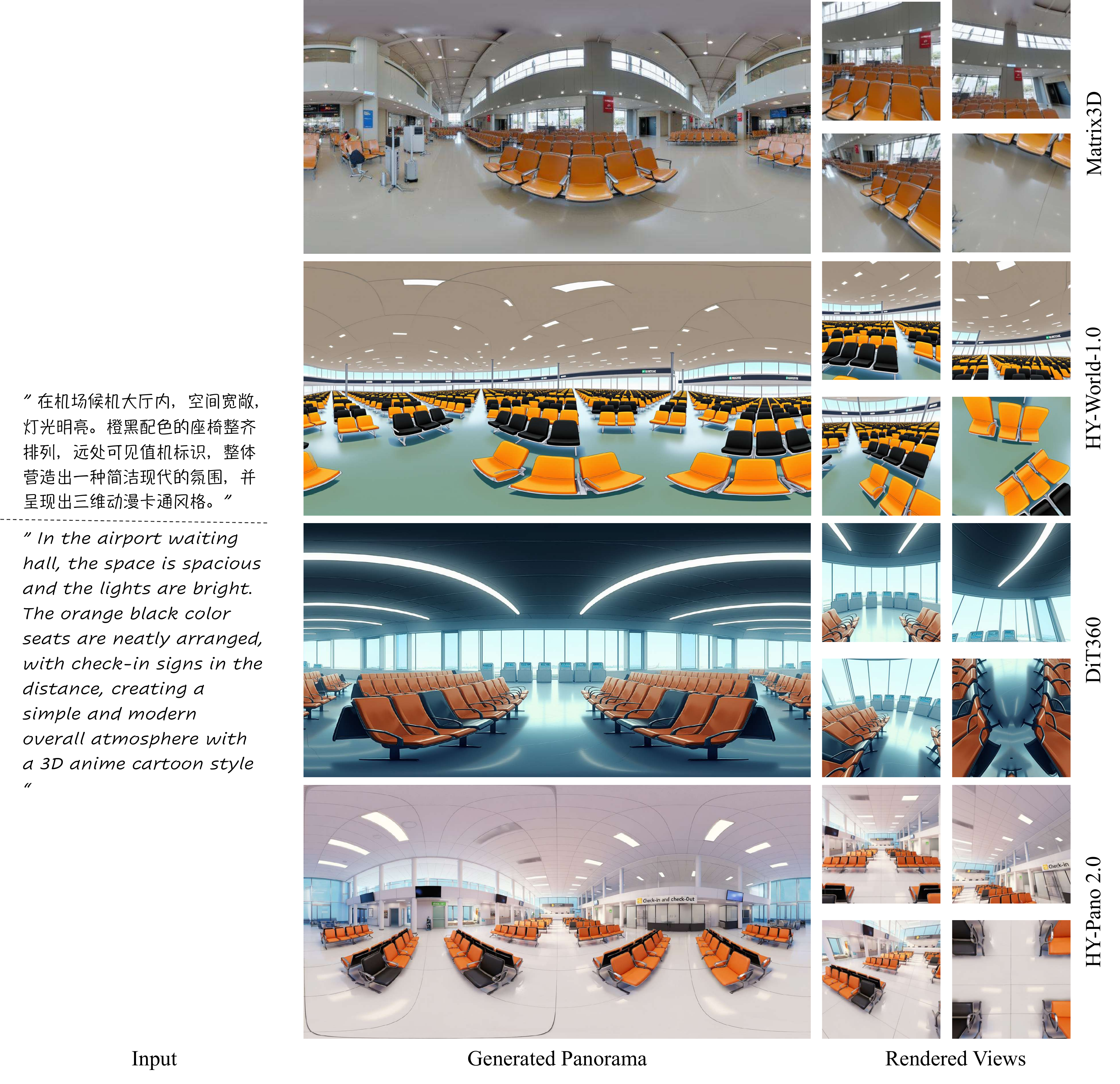}
\vspace{-0.15in}
\caption{\textbf{Qualitative comparison on the text-to-panorama (T2P) task.} Our method outperforms previous approaches in terms of layout coherence, fine-grained details, and overall visual aesthetics.\label{fig:t2p}}
\vspace{-0.15in}
\end{figure*}

\begin{figure*}[t]
\centering
\includegraphics[width=1.045\linewidth]{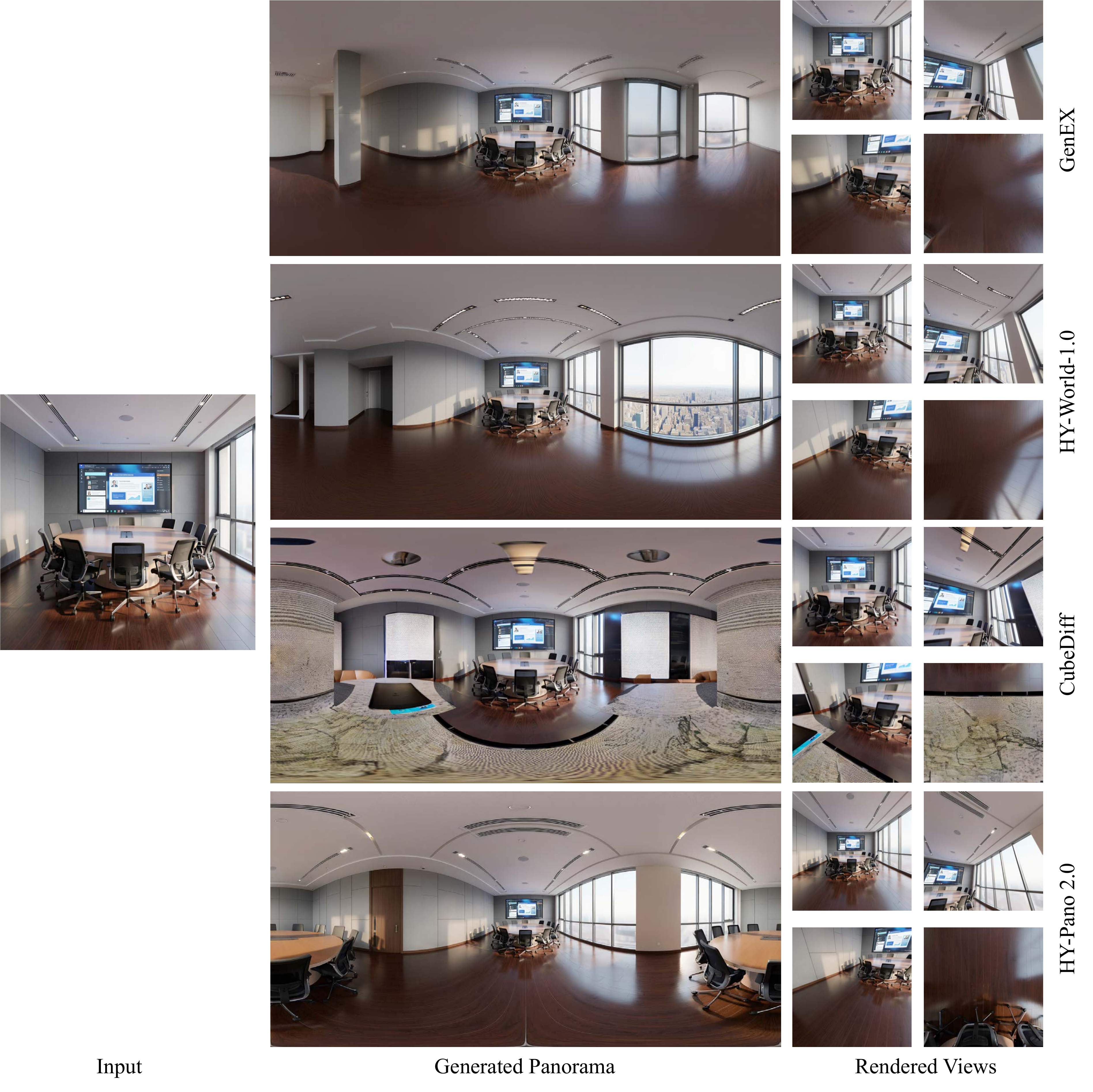}
\vspace{-0.15in}
\caption{\textbf{Qualitative comparison on the image-to-panorama (I2P) task.} Our method outperforms previous approaches in extension plausibility, content richness, and overall quality.\label{fig:i2p}}
\vspace{-0.15in}
\end{figure*}

\subsubsection{Results \& Analysis of HY-Pano 2.0}
\label{sec:exp_pano}

For both qualitative and quantitative comparisons, we evaluate the panorama generation of HY-Pano 2.0 against several state-of-the-art approaches across text-to-panorama (T2P) and image-to-panorama (I2P). 
For T2P, we compare with DiT360~\cite{feng2025dit360}, Matrix3D~\cite{yang2025matrix}, and HY-World 1.0~\cite{hunyuanworld2025tencent}. 
For I2P, we compare with CubeDiff~\cite{kalischek2025cubediff}, GenEx~\cite{lu2024genex}, and HY-World 1.0~\cite{hunyuanworld2025tencent}.

\paragraph{Quantitative Results.}
\Cref{tab:main} presents the quantitative comparison for both T2P and I2P tasks.
We evaluate generated panoramas using multiple complementary metrics. CLIP-T~\cite{radford2021learning} (T2P) and CLIP-I~\cite{radford2021learning} (I2P) measure text-image and image-image alignment, respectively. Q-Align~\cite{wu2023qalign} provides both perceptual quality (Qual) and aesthetic (Aes) scores based on a large multi-modal model aligned with human ratings. For all applicable metrics, we report results on both the equirectangular (Equi) panorama and averaged perspective (Persp) projections, where each panorama is projected onto 12 perspective faces.
As shown in \Cref{tab:main}, HY-Pano 2.0 achieves the best scores on the majority of metrics across both tasks. For T2P, it obtains the highest CLIP-T score and leads on most Q-Align quality and aesthetics measures. For I2P, it ranks first on all five metrics, with notable improvements over HY-World 1.0 in both perceptual quality and aesthetics. These results demonstrate that HY-Pano 2.0 exhibits stronger adherence to input signals (text prompts or reference images), improved fine-grained detail quality, and enhanced aesthetic score compared to prior methods.

\noindent\textbf{Qualitative Results.}
We first show some generated panoramas conditioned on image and text inputs in \Cref{fig:pano_results}.
Then, we present qualitative comparisons for T2P and I2P in \Cref{fig:t2p} and \Cref{fig:i2p}, respectively. 
Compared to existing methods, HY-Pano 2.0 generates panoramas with more structurally coherent layouts, exhibiting plausible spatial arrangements and consistent geometric structures across the full 360$^\circ$ field of view. In terms of visual aesthetics, our results demonstrate superior color harmony, lighting consistency, and overall artistic quality. Furthermore, HY-Pano 2.0 produces notably finer details, including sharper textures, cleaner object boundaries, and richer high-frequency content, leading to more realistic and visually appealing panoramas.

\begin{table*}[t]
\centering
\caption{\textbf{Quantitative comparisons on text-to-panorama (T2P) and image-to-panorama (I2P).}}
\label{tab:main}
\resizebox{\linewidth}{!}{
\begin{tabular}{l|cccc|cccc}
\toprule
 & \multicolumn{4}{c|}{\textbf{Text-to-Panorama (T2P)}} & \multicolumn{4}{c}{\textbf{Image-to-Panorama (I2P)}} \\
Metric & DiT360 & Matrix3D & HY-World 1.0 & HY-Pano 2.0 & CubeDiff & GenEx & HY-World 1.0 & HY-Pano 2.0 \\
\midrule
CLIP-T / CLIP-I $\uparrow$ & 0.248 & 0.238 & 0.250 & \textbf{0.258} & 0.828 & 0.831 & 0.831 & \textbf{0.844} \\
Q-Align Qual (Persp) $\uparrow$ & 3.788 & 2.983 & 3.992 & \textbf{4.103} & 2.938 & 2.917 & 3.317 & \textbf{4.026} \\
Q-Align Qual (Equi) $\uparrow$ & 4.436 & 4.258 & \textbf{4.493} & 4.403 & 3.814 & 3.868 & 4.076 & \textbf{4.277} \\
Q-Align Aes (Persp) $\uparrow$ & 2.882 & 2.126 & \textbf{3.404} & 3.376 & 2.319 & 2.445 & 2.638 & \textbf{3.208} \\
Q-Align Aes (Equi) $\uparrow$ & 4.072 & 3.880 & 4.186 & \textbf{4.247} & 3.645 & 3.646 & 3.767 & \textbf{4.056} \\
\bottomrule
\end{tabular}}
\end{table*}

\subsubsection{Results \& Analysis of WorldNav}
\label{sec:exp_traj}

\begin{figure}
    \centering
    \includegraphics[width=1.0\linewidth]{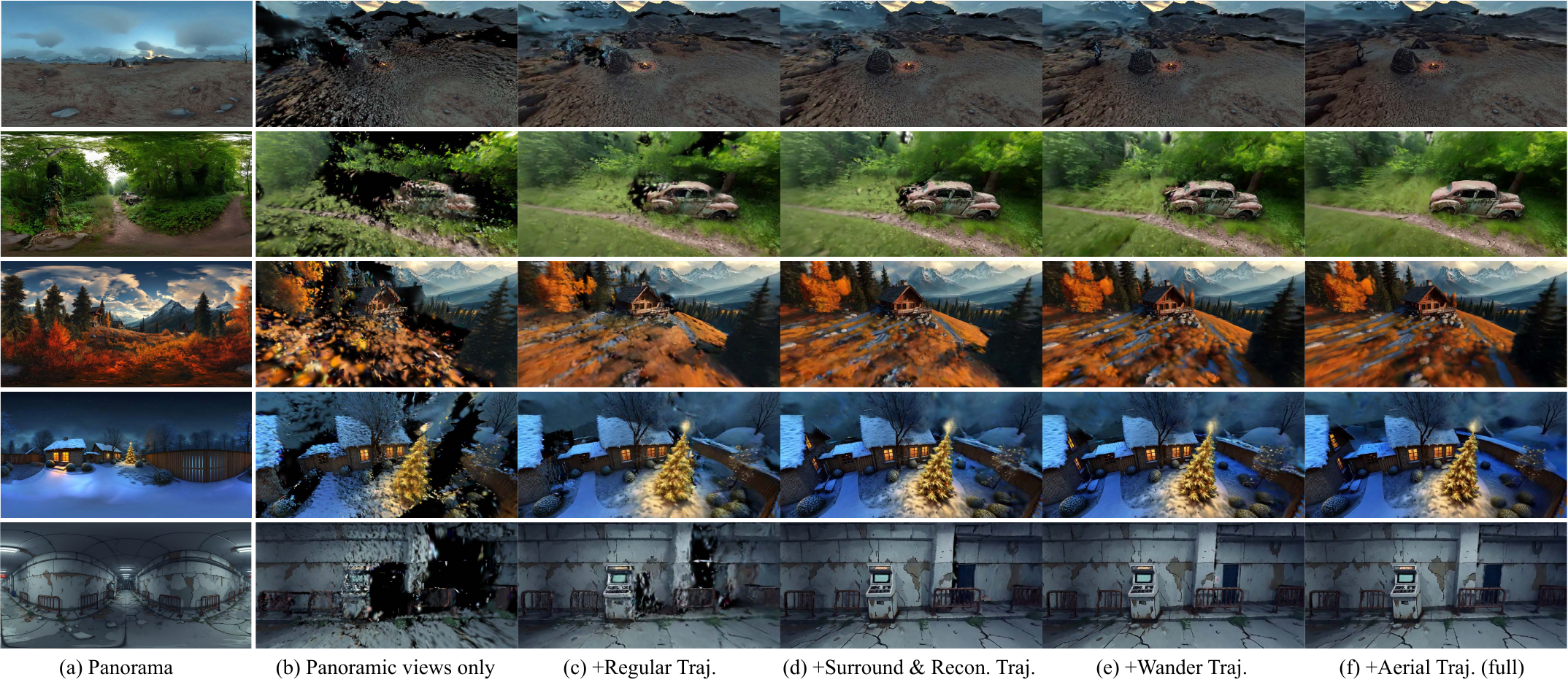}
    \caption{\textbf{Qualitative ablation results of trajectory planning.} Relying solely on panoramic views (b) results in severe artifacts and incomplete geometry. By sequentially integrating views generated from (c) regular, (d) surrounding \& reconstruction, (e) wandering, and (f) aerial trajectories, our method progressively eliminates blind spots, refines complex object structures, and enhances overall scene completeness. Please zoom in for details.
    \label{fig:traj_ablation}}
    \vspace{-0.15in}
\end{figure}

We present qualitative comparisons in \Cref{fig:traj_ablation} to intuitively demonstrate the necessity of each trajectory planning component. 
Training 3DGS solely on panoramic views (\Cref{fig:traj_ablation}b) inevitably suffers from massive geometric voids and poor rendering quality.
By sequentially integrating views from different trajectories, the scene completeness progressively improves.
Specifically, regular trajectories (\Cref{fig:traj_ablation}c) break the limitation of a fixed viewpoint, providing expanded observations that eliminate most large-scale artifacts. 
However, regular paths often fail to cover occluded structures, leaving the sides and backs of specific objects (\eg, the car, cabin, and arcade machine) incomplete.
This limitation is effectively resolved by introducing surrounding and reconstruction-aware trajectories (\Cref{fig:traj_ablation}d), which explicitly target and complete these complex structures. 
Furthermore, wandering trajectories (\Cref{fig:traj_ablation}e) enhance the textural details of distant walls and floors, enabling good roaming experiences. 
Finally, aerial trajectories (\Cref{fig:traj_ablation}f) incorporate additional bird's-eye view (BEV) observations, improving the freedom of the 3D world's viewpoint changing.

\subsubsection{Results \& Analysis of WorldStereo 2.0}
\label{sec:exp_video_gen}

\begin{table}
\centering
\caption{\textbf{Point cloud results of one-view-generated 3D reconstruction.} 
Precision measures the percentage of generated points that fall within a distance threshold of ground-truth points, while Recall measures the percentage of ground-truth points covered by generated points. The F1-score is their harmonic mean, and AUC denotes the area under the curve across varying distance thresholds.
\label{tab:single_view_recon_exp}}
\fontsize{8.0}{9}\selectfont
\setlength{\tabcolsep}{5pt}
\begin{tabular}{lcccccccc}
\toprule
\multirow{2}{*}{Methods} & \multicolumn{4}{c}{Tanks-and-Temples~\cite{knapitsch2017tanks}} & \multicolumn{4}{c}{MipNeRF360~\cite{barron2022mip}}\tabularnewline
\cmidrule(lr){2-5}\cmidrule(lr){6-9}
 & Precision$\uparrow$ & Recall$\uparrow$ & F1-Score$\uparrow$ & AUC$\uparrow$ & Precision$\uparrow$ & Recall$\uparrow$ & F1-Score$\uparrow$ & AUC$\uparrow$\tabularnewline
\midrule
SEVA~\cite{zhou2025stable} & 33.59 & 35.34 & 36.73 & 51.03 & 22.38 & 55.63 & 28.75 & 46.81\tabularnewline
Gen3C~\cite{ren2025gen3c} & \underline{46.73} & 25.51 & 31.24 & 42.44 & 23.28 & \textbf{75.37} & 35.26 & 52.10\tabularnewline
Lyra~\cite{bahmani2025lyra} & \textbf{50.38} & 28.67 & 32.54 & 43.05 & 30.02 & 58.60 & 36.05 & 49.89\tabularnewline
FlashWorld~\cite{yang2025flash} & 26.58 & 20.72 & 22.29 & 30.45 & 35.97 & 53.77 & 42.60 & 53.86\tabularnewline
WorldStereo 2.0 & 43.62 & \underline{41.02} & \underline{41.43} & \underline{58.19} & \textbf{43.19} & \underline{65.32} & \textbf{51.27} & \textbf{65.79}\tabularnewline
WorldStereo 2.0 (DMD) & 40.41 & \textbf{44.41} & \textbf{43.16} & \textbf{60.09} & \underline{42.34} & 64.83 & \underline{50.52} & \underline{65.64}\tabularnewline
\bottomrule
\end{tabular}
\end{table}
 
\begin{table}
\centering
\vspace{-0.15in}
\caption{\textbf{Quantitative results of camera control capability.} 
We evaluate performance across camera metrics and visual quality. Methods labeled with $^{*}$ indicate that the model is only trained with camera control (domain-adaption) without memory capabilities.
\label{tab:camera_control_exp}}
\fontsize{8.5}{9}\selectfont
\setlength{\tabcolsep}{5.5pt}
\begin{tabular}{lccccccc}
\toprule
\multirow{2}{*}{Methods} & \multicolumn{3}{c}{Camera Metrics} & \multicolumn{4}{c}{Visual Quality}\tabularnewline
\cmidrule(lr){2-4}\cmidrule(lr){5-8}
 & RotErr$\downarrow$ & TransErr$\downarrow$ & ATE$\downarrow$ & Q-Align$\uparrow$ & CLIP-IQA+$\uparrow$ & Laion-Aes$\uparrow$ & CLIP-I$\uparrow$\tabularnewline
\midrule
SEVA~\cite{zhou2025stable} & 1.690 & 1.578 & 2.879 & 3.232 & 0.479 & 4.623 & 77.16\tabularnewline
Gen3C~\cite{ren2025gen3c} & 0.944 & 1.580 & 2.789 & 3.353 & 0.489 & 4.863 & 82.33 \tabularnewline
WorldPlay~\cite{sun2025worldplay} & 3.481 & 1.288 & 2.722 & 3.628 & \textbf{0.552} & 5.103 & 86.79\tabularnewline
WorldCompass~\cite{wang2026worldcompass} & 3.452 & \underline{1.068} & 2.379 & 3.615 & \underline{0.548} & 5.111 & 85.51\tabularnewline
WorldStereo~\cite{worldstereo2026}$^{*}$ & \underline{0.762} & 1.245 & \underline{2.141} & \underline{4.149} & 0.547 & \underline{5.257} & \underline{89.05}\tabularnewline
WorldStereo 2.0$^{*}$ & \textbf{0.492} & \textbf{0.968} & \textbf{1.768} & \textbf{4.205} & 0.544 & \textbf{5.266} & \textbf{89.43}\tabularnewline
\bottomrule
\end{tabular}
\end{table}

\paragraph{Results of Single-View Scene Reconstruction.}
Following WorldStereo~\cite{worldstereo2026}, we evaluate WorldStereo 2.0 on the single-view scene reconstruction benchmark in \Cref{tab:single_view_recon_exp}, utilizing the out-of-distribution Tanks-and-Temples~\cite{knapitsch2017tanks} and MipNeRF360~\cite{barron2022mip} datasets.
For quantitative evaluation, we compare our results against pseudo ground-truth point clouds reconstructed via Multi-View-Stereo~\cite{cao2024mvsformer++} from real multi-view images.
To rigorously test our method, we introduce more challenging camera trajectories than the original benchmark: closed-loop paths for object-centric scenes and manually designed, explorable routes for large-scale forward-facing scenes. This significantly increases the difficulty of maintaining multi-view consistency. 
As demonstrated in \Cref{tab:single_view_recon_exp}, WorldStereo 2.0 achieves the highest point cloud F1 and AUC scores, surpassing all video-based and 3D-based competitors. Although single-view generative reconstruction inherently suffers from high uncertainty, these superior geometric metrics confirm that our approach successfully synthesizes highly consistent and physically plausible 3D structures.

\begin{table}
\centering
\vspace{-0.15in}
\caption{\textbf{Camera control ablation studies of WorldStereo 2.0.} We evaluate performance across camera metrics, visual quality, and user study. Given the misalignment between the commonly used metrics and human perception, we prioritize user study results for model selection. 
The \textcolor{red}{red row} indicates the baseline (camera control of WorldStereo~\cite{worldstereo2026}), while the \textcolor{blue}{blue row} denotes our final domain-adaption setting.
\label{tab:camera_control_ablation}}
\fontsize{8.0}{9}\selectfont
\setlength{\tabcolsep}{2.2pt}
\begin{tabular}{ll ccc ccc cc}
\toprule
\multirow{2}{*}{Frozen Parts} & \multirow{2}{*}{VAE Types} &
\multicolumn{3}{c}{Camera Metrics} &
\multicolumn{3}{c}{Visual Quality} &
\multicolumn{2}{c}{User Study} \\
\cmidrule(lr){3-5}\cmidrule(lr){6-8}\cmidrule(lr){9-10}
 &  &
RotErr$\downarrow$ & TransErr$\downarrow$ & ATE$\downarrow$ &
Q-Align$\uparrow$ & CLIP-IQA+$\uparrow$ & Laion-Aes$\uparrow$ &
Camera$\uparrow$ & Quality$\uparrow$ \\
\midrule
\rowcolor{red!10}
Main DiT & Video-VAE
& 0.762 & 1.245 & 2.141
& 4.149 & \underline{0.547} & 5.257
& 84.85\% & 46.46\% \\
Main DiT & Keyframe-VAE
& 0.768 & 1.149 & \underline{2.027}
& 4.060 & 0.520 & 5.210
& -- & -- \\
None & Keyframe-VAE
& \underline{0.578} & \underline{1.115} & 2.245
& \textbf{4.237} & \textbf{0.554} & \textbf{5.278}
& \textbf{93.81\%} & 60.61\% \\
Cross-Attn & Keyframe-VAE
& 0.684 & 1.243 & 2.111
& 4.181 & 0.538 & 5.235
& \underline{93.13\%} & \underline{60.95\%} \\
\rowcolor{blue!10}
Cross-Attn + FFN & Keyframe-VAE
& \textbf{0.492} & \textbf{0.968} & \textbf{1.768}
& \underline{4.205} & 0.544 & \underline{5.266} & 92.44\% & \textbf{64.39\%} \\
\bottomrule
\end{tabular}
\end{table}

\paragraph{Results of Camera Control Capability.}
We quantitatively evaluate the camera control capability of WorldStereo 2.0 in \Cref{tab:camera_control_exp}, while ablation studies are performed in \Cref{tab:camera_control_ablation}. Both evaluations are applied with 100 out-of-distribution images selected from~\cite{duan2025worldscore} with challenging trajectories.
Notably, WorldStereo 2.0 outperforms all video-based competitors by achieving the lowest errors across all camera metrics. Furthermore, it also delivers superior visual quality and semantic alignment. 
For the ablation study in \Cref{tab:camera_control_ablation}, since Keyframe-VAE introduces significant changes to the latent representations, directly applying it without training the main network is unfair and yields limited improvements. Therefore, we unlock the main DiT for full training (freezing ``None'').
Compared with the Video-VAE baseline, the fully trained Keyframe-VAE significantly improves visual quality, user-perceived camera control, and most camera metrics.
Moreover, we provide qualitative comparisons in \Cref{fig:vae-visual}, which further support this conclusion.
However, we observe an obvious trade-off between performance and generalization for the trainable parts of the main DiT.
While full model training maximizes visual metrics, it leads to inferior camera precision and suboptimal user study quality.
We find that this is due to overfitting issues, where the global image style slightly drifts during video generation. To address this, we selectively freeze specific layers. As shown in the blue row, freezing the cross-attention and FFN layers achieves the best balance.
It effectively mitigates the overfitting, yielding the most precise camera control with the lowest RotErr, TransErr, and ATE, while attaining the highest user preference for visual quality (64.39\%).

\begin{table}
\centering
\caption{\textbf{Memory and distillation ablation studies of WorldStereo 2.0.} 
PSNR$_m$ and SSIM$_m$ are computed within \textit{valid warping mask} regions to evaluate consistency.
The configuration A$^*$ denotes a variant where the SSM incorporates reference features via temporal concatenation rather than spatial concatenation. 
Settings of our final memory and distilled models are colorized in \textcolor{softgreen}{green} and \textcolor{softyellow}{yellow}.
\label{tab:memory_ablation}}
\fontsize{8.0}{9}\selectfont
\setlength{\tabcolsep}{2.5pt}
\begin{tabular}{llcccccccc}
\toprule
\multirow{2}{*}{} & \multirow{2}{*}{Configuration} & \multicolumn{3}{c}{Photometric Metrics} & \multicolumn{2}{c}{Consistency} & \multicolumn{3}{c}{Camera Metrics}\tabularnewline
\cmidrule(lr){3-5}\cmidrule(lr){6-7}\cmidrule(lr){8-10}
 &  & PSNR$\uparrow$ & SSIM$\uparrow$ & LPIPS$\downarrow$ & PSNR$_{m}\uparrow$ & SSIM$_{m}\uparrow$ & RotErr$\downarrow$ & TransErr$\downarrow$ & ATE$\downarrow$\tabularnewline
 \midrule
 & Camera control baseline & 16.13 & 0.474 & 0.349 & 28.81 & 0.448 & 0.396 & 0.053 & 0.071\tabularnewline
 \midrule
A & GGM and SSM++ (A) & 20.94 & 0.640 & 0.170 & 30.27 & 0.623 & 0.407 & 0.047 & 0.046\tabularnewline
B & Trainable FFN (A,B) & 21.56 & \underline{0.667} & \underline{0.162} & 30.44 & 0.624 & 0.351 & \textbf{0.036} & \textbf{0.035}\tabularnewline
C & Pointcloud augmentation (A,B,C) & 21.36 & 0.632 & 0.163 & 30.72 & 0.619 & 0.360 & 0.050 & 0.053\tabularnewline
D & Reference augmentation (A,B,C,D) & 20.86 & 0.639 & 0.165 & 30.66 & 0.636 & 0.322 & 0.049 & 0.067\tabularnewline
E & Camera embedding (A,B,C,D,E) & 21.06 & 0.639 & 0.164 & 30.58 & 0.617 & 0.329 & \underline{0.042} & 0.048\tabularnewline
A$^*$ & Temporal-concated SSM (A$^*$,B,C,D,E) & 19.83 & 0.581 & 0.219 & 29.77 & 0.571 & 0.545 & 0.087 & 0.114\tabularnewline
\rowcolor{rowgreen}
F & Doubled batch-size (64) (A,B,C,D,E,F) & \underline{21.63} & \textbf{0.669} & \textbf{0.156} & \underline{30.76} & \underline{0.647} & \textbf{0.296} & \textbf{0.036} & \underline{0.041}\tabularnewline
\midrule
\rowcolor{rowyellow}
G & After post-distillation (A,B,C,D,E,F,G) & \textbf{21.84} & \textbf{0.669} & 0.165 & \textbf{30.93} & \textbf{0.656} & \underline{0.316} & 0.052 & 0.072\tabularnewline
\bottomrule
\end{tabular}
\end{table}

\paragraph{Ablation Studies of Memory Training and Distillation.}
We comprehensively evaluate the memory training and post-distillation in \Cref{tab:memory_ablation}.
Incorporating GGM and SSM++ (Config A) substantially improves the photometric quality and multi-trajectory consistency.
Furthermore, unfreezing the FFN (Config B) significantly enhances camera control precision.
To mitigate overfitting to point cloud guidance and retrieved reference views, we introduce several augmentation strategies (Configs C and D, detailed in \Cref{sec:mem_aug}) to these conditions.
While these regularizations cause a little gap in clean-data metrics, they are crucial for overall robustness and maintain highly competitive performance.
Moreover, we validate our spatial-stereo stitching design in SSM. Replacing it with temporal concatenation (Config A$^*$) severely degrades performance across all metrics.
Additionally, scaling the training batch size to 64 (Config F) stabilizes training, yielding consistent improvements.
Finally, after applying DMD post-distillation (Config G), the model not only retains comparable camera control but even slightly improves photometric and consistency metrics.

\subsubsection{Results \& Analysis of World Composition}
\label{sec:exp_rec_align}

\begin{figure}
    \centering
    \includegraphics[width=\linewidth]{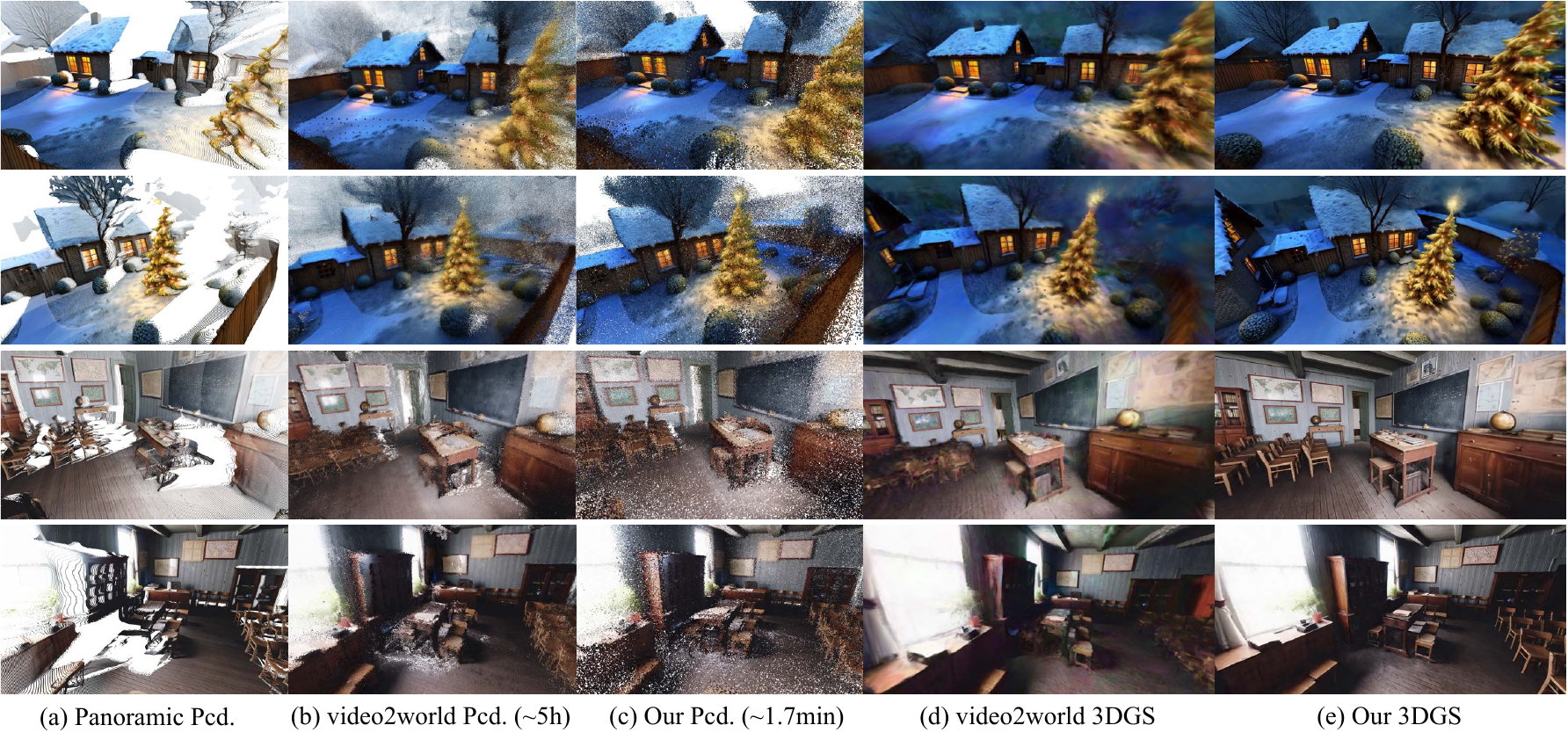}
    \caption{\textbf{Point clouds and 3DGS comparisons with video2world~\cite{hollein2026world}.} ``Pcd.'' denotes point cloud. 
    For each scene, both methods are evaluated within 300 images generated by WorldStereo 2.0.
    \label{fig:world_recon_compare}}
\end{figure}

\textbf{Reconstruction and Alignment.}
While \Cref{sec:pcd_expansion} establishes the effectiveness of WorldMirror 2.0 in point cloud expansion with known camera poses, we further evaluate our overall composition pipeline against the concurrent world reconstruction method, video2world~\cite{hollein2026world} in \Cref{fig:world_recon_compare}.
To ensure a fair comparison, both methods are evaluated on 300-view images generated by WorldStereo 2.0, which reaches the memory limit of an NVIDIA H20 GPU for video2world.
As illustrated in \Cref{fig:world_recon_compare}, although video2world produces impressive point clouds via feature-matched Iterative Closest Point (ICP), this process is inherently difficult to parallelize, resulting in a prohibitive computational overhead of approximately 5 hours per scene. In contrast, our lightweight linear alignment fully leverages camera pose priors to achieve comparable reconstruction quality in less than 2 minutes. 
Furthermore, our final 3DGS reconstructions exhibit superior geometric and textural details, largely attributed to the tailored optimization strategies proposed in \Cref{sec:3dgs}. Notably, because our WorldStereo 2.0 generates sequences with significantly higher visual consistency than SEVA~\cite{zhou2025stable}, the complex non-rigid aware 3DGS required by video2world~\cite{hollein2026world} becomes unnecessary in our pipeline. Additionally, we observe that the SH optimization often leads to undesirable color artifacts rendered in novel views (see \Cref{fig:world_recon_compare}(d)). Consequently, our pipeline adopts a direct RGB optimization, which proves to be more robust and effective for generative scenarios.

\begin{table}
\centering
\caption{\textbf{3DGS ablation studies averaged over 10 scenes.} $\dagger$ denotes that the adaptive densification is restricted to non-sky regions.}
\label{tab:ablation_gs}
\small
\setlength{\tabcolsep}{4.5pt}
\begin{tabular}{cccccccc}
\toprule
Voxel Downsample & Adaptive Densification & MaskGaussian & GS Number & PSNR$\uparrow$ & SSIM$\uparrow$ & LPIPS$\downarrow$ \\
\midrule
           &                        &            & 6.000M & 25.176 & 0.751 & 0.209 \\
\checkmark &                        &            & 1.000M & 24.504 & 0.720 & 0.276 \\
\checkmark & \checkmark             &            & 5.254M & 25.158 & 0.750 & 0.210 \\
\checkmark & \checkmark             & \checkmark & 1.383M & 25.017 & 0.747 & 0.216 \\
\rowcolor{gray!15}
\checkmark & \checkmark$^{\dagger}$ & \checkmark & 1.381M & 25.023 & 0.747 & 0.215 \\
\bottomrule
\end{tabular}
\end{table}

\textbf{Gaussian Splattings.}
\label{sec:exp_3dgs}
We ablate each component of the proposed 3DGS pipeline across 10 scenes, evaluating each on a 20-view validation set (\Cref{tab:ablation_gs}).
The baseline initializes from 6M Gaussians randomly sampled from the expanded point cloud~$\mathbf{\tilde{P}}$, yielding the highest quality (PSNR 25.176) but incurring substantial rendering overhead.
Applying voxel downsampling alone reduces the Gaussian count to 1M, but at a severe cost to quality---a 0.68 dB drop in PSNR and a 32\% increase in LPIPS---confirming that uniform decimation disproportionately degrades detail-rich regions.
Enabling adaptive densification restores the quality to near-baseline levels (PSNR 25.158), yet inflates the count to 5.254M, largely negating the efficiency gains of downsampling. 
Integrating MaskGaussian resolves this trade-off: redundant Gaussians in low-frequency areas are pruned, reducing the count by 73.7\% (from 5.254M$\to$1.383M) with only $-$0.14\,dB PSNR degradation.
Further restricting densification to non-sky regions suppresses floaters where depth supervision is unavailable.
The full configuration retains comparable visual quality while reducing 77\% Gaussian count compared to the baseline.

\subsubsection{Full Results \& Comparison with Marble}
\label{sec:exp_overall_world_gen}

\begin{figure}
    \centering
    \includegraphics[width=1.0\linewidth]{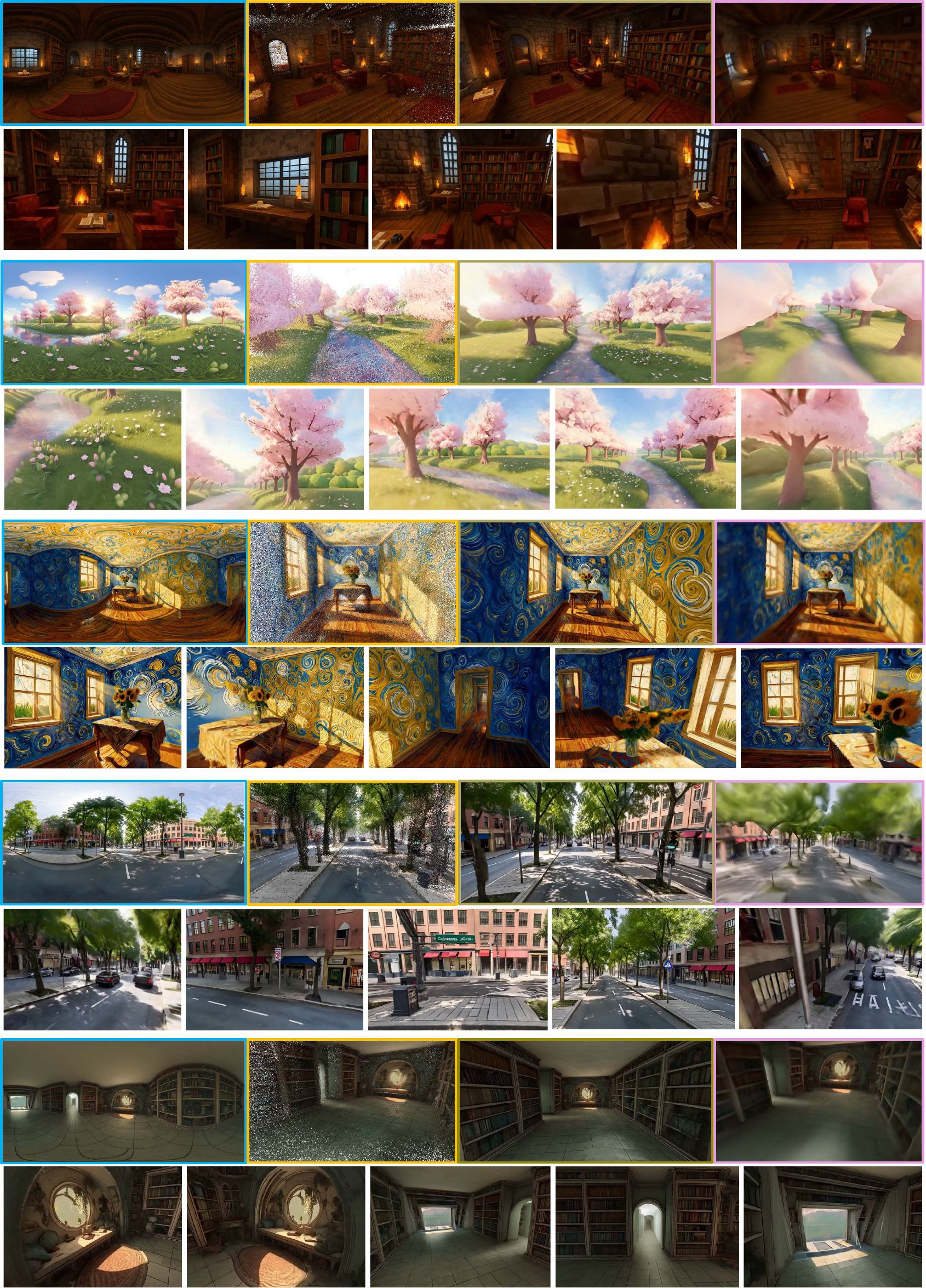}
    \caption{\textbf{Results of the overall world generation pipeline.} Each scene is visualized across two rows. The top row displays, from left to right: the generated \textcolor{cyan}{panorama}, the aligned \textcolor{orange}{point clouds}, a global overview of \textcolor{olive}{splattings}, and the extracted \textcolor{magenta}{coarse mesh}. The bottom row showcases novel views rendered from various viewpoints.}
    \label{fig:overall_result}
\end{figure}

\begin{figure}
    \centering
    \includegraphics[width=1.0\linewidth]{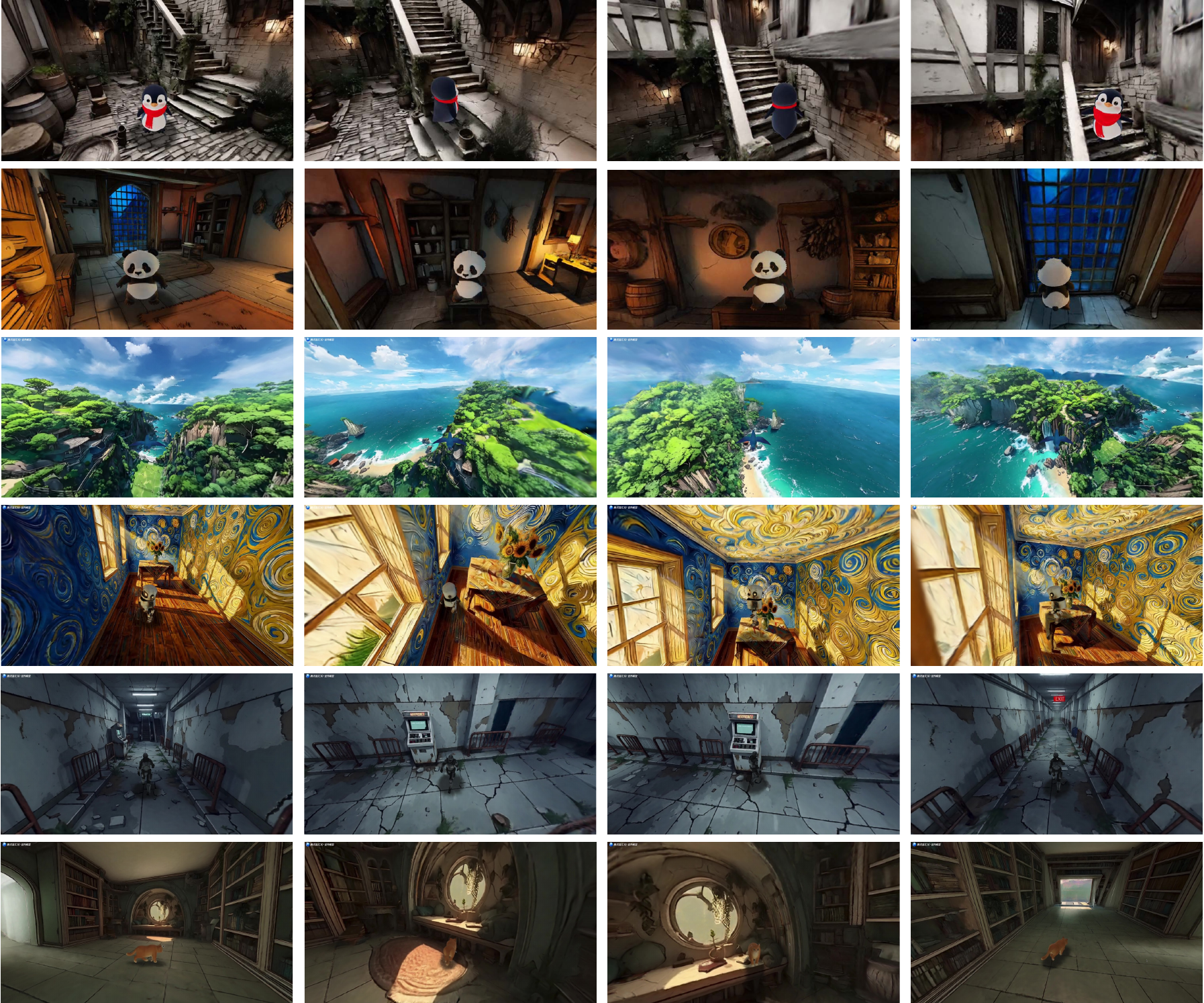}
    \caption{\textbf{Interactive exploration within the generated 3D worlds of HY-World 2.0.} 
    By controlling virtual agents, users can navigate complex geometric structures (\eg, stairs and indoor layouts) with real-time collision detection and physically plausible feedback, demonstrating the readiness of our results for interactive applications.}
    \label{fig:physical}
\end{figure}

\paragraph{Explorable and Interactive Worlds.} 
As illustrated in \Cref{fig:overall_result}, HY-World 2.0 yields comprehensive multi-modal 3D assets, encompassing panoramas, aligned point clouds for 3DGS initialization, high-fidelity 3DGS renderings, and extracted geometric meshes. 
Crucially, these rich 3D representations transcend static visualization, serving as foundational environments for explorable and interactive 3D worlds (see \Cref{fig:physical}). 
By leveraging the meshes extracted from 3DGS as underlying collision proxies, our system supports real-time physical feedback and spatial interactions.
To ensure seamless user experiences and rapid scene loading, we optimize these meshes into lightweight topological structures. This deliberate design strikes a balance between physical plausibility and efficiency, paving the way for downstream applications in gaming, virtual reality, and embodied AI.

\begin{figure}
    \centering
    \includegraphics[width=1.035\linewidth]{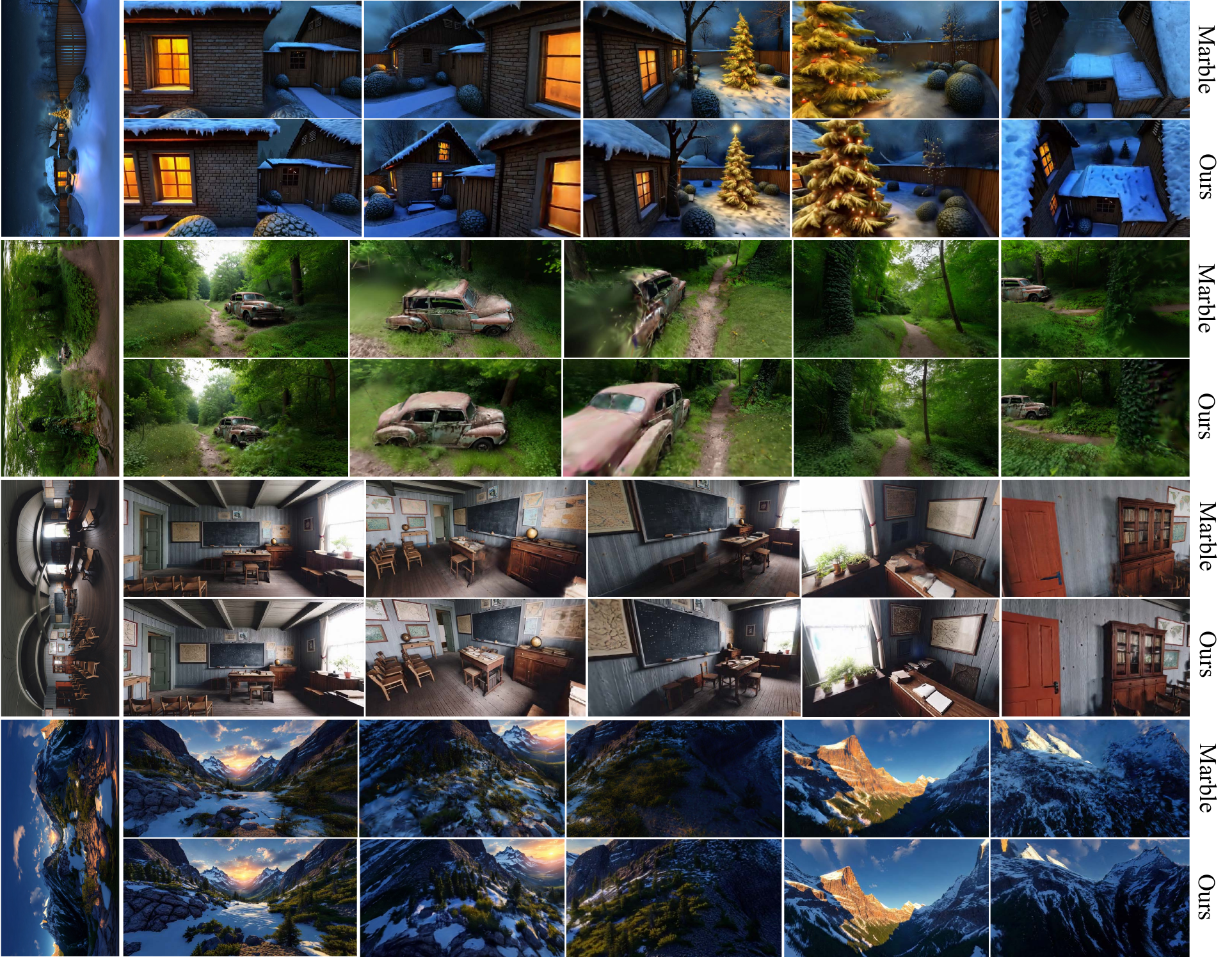}
    \caption{\textbf{Qualitative comparison with Marble~\cite{marble_worldlabs_2026} using the same panoramic inputs.} The input panoramas are displayed on the left. For each scene, we present novel view renderings from the generated 3DGS models of both methods. Compared to Marble, our approach achieves higher fidelity to the input conditions, sharper textures, and superior geometric consistency across diverse viewpoints. Please zoom in for details.}
    \label{fig:compare_to_marble0}
\end{figure}

\begin{figure}
    \centering
    \includegraphics[width=1.03\linewidth]{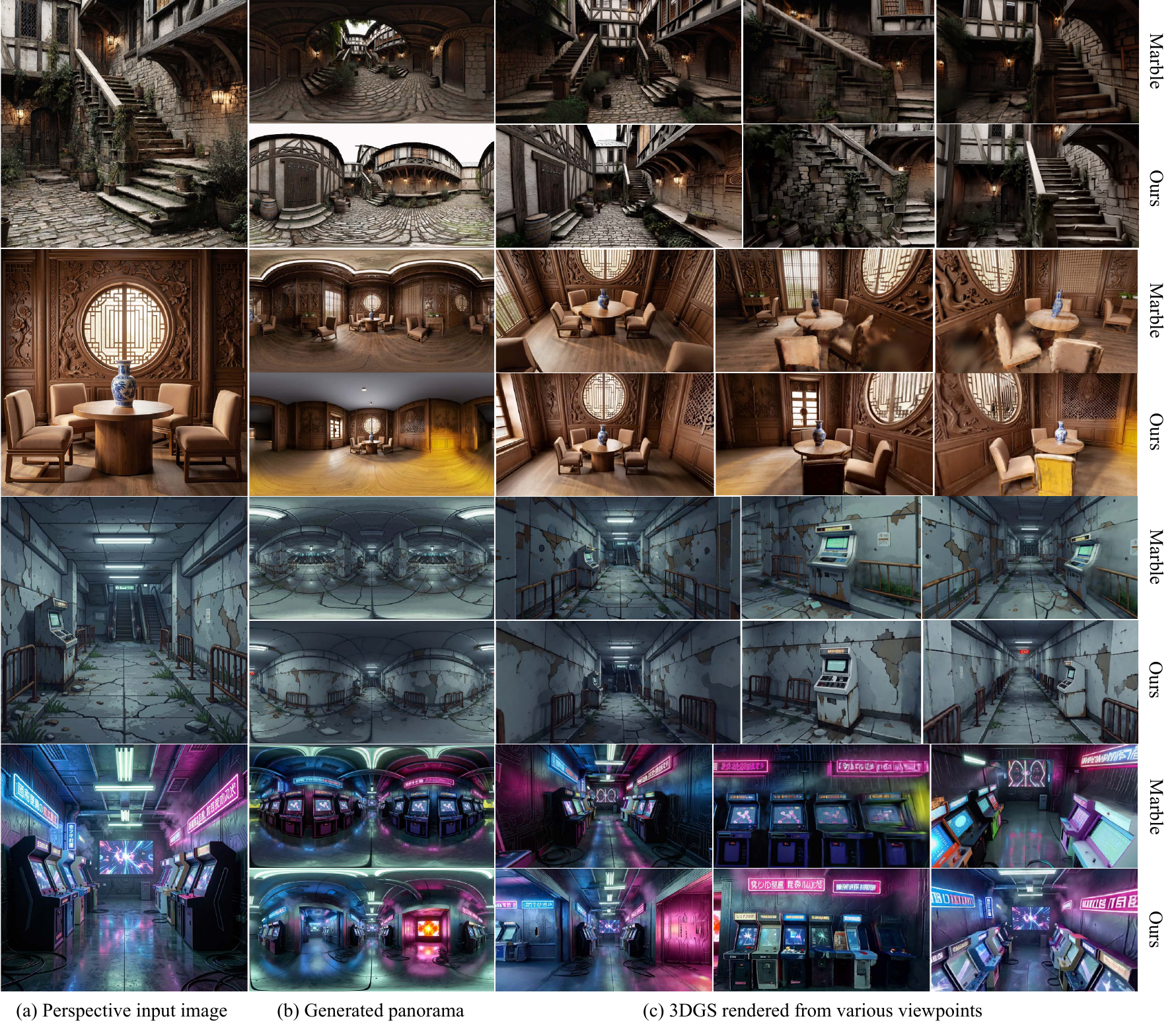}
    \caption{\textbf{Qualitative comparison with Marble~\cite{marble_worldlabs_2026} using the same input image.} (a) displays input perspective images. For each scene, we compare both (b) generated panoramas and (c) 3DGS renderings. Compared to Marble, our approach better adheres to the input views while achieving comparable quality and superior completeness in 3DGS. Please zoom in for details.}
    \label{fig:compare_to_marble1}
\end{figure}

\paragraph{Comparisons with the State-of-the-art.}
We compare our approach against the closed-source commercial world model, Marble~\cite{marble_worldlabs_2026}\footnote{In this report, we compare our method with Marble 1.0 (as of March 30, 2026).}.
The comparison is conducted under two settings: using identical panoramic inputs (\Cref{fig:compare_to_marble0}) and using the same perspective conditions (\Cref{fig:compare_to_marble1}).
While Marble can produce impressive 3DGS results, it usually deviates from the input guidance, resulting in noticeable discrepancies and lower fidelity in regions explicitly covered by the panoramic or perspective conditions.
In contrast, our method achieves high-fidelity results that strictly adhere to the provided conditions. Furthermore, our generation outperforms Marble in terms of detail preservation and geometric consistency of novel views. 
As illustrated from \Cref{fig:compare_to_marble0} and \Cref{fig:compare_to_marble1}, our results maintain superior structural integrity and smoother textures across fences, cars, furniture, mountains, and arcade machines, whereas Marble suffers from severe blurring and geometric missing under large viewpoint changes.

\begin{table}
\centering
\caption{\textbf{Runtime of each component in HY-World 2.0 for one single world generation.} ``Recon'' means reconstruction. All efficiency evaluations are performed using NVIDIA H20 GPUs. \label{tab:time_cost_worldgen}}
\small
\begin{tabular}{lcccccc}
\toprule
Stage & Panorama & Trajectory Plan & World Expansion & Recon and Align & 3DGS & Total\tabularnewline
\midrule
Time (sec) & 15s & 182s & 286s & 102s & 127s & 712s\tabularnewline
\bottomrule
\end{tabular}
\end{table}

\paragraph{Runtime Analysis.}
We evaluate the overall runtime of HY-World 2.0 on NVIDIA H20 GPUs, as detailed in \Cref{tab:time_cost_worldgen}. By integrating systematic efficiency optimizations, the end-to-end pipeline for generating a complete 3D world is accelerated, requiring only 10 minutes. Specifically, we employ Sequence Parallelism (SP) to distribute all model inference stages, including panorama generation, Keyframe-VAE, WorldStereo 2.0, and WorldMirror 2.0. Furthermore, the overall efficiency is improved by incorporating other acceleration techniques, such as SageAttention2~\cite{zhang2025sageattention2}, FP8 mixed-precision inference, and step caching mechanisms~\cite{liu2025survey}.

\subsection{World Reconstruction from Multi-View Images or Video}
\label{sec:recon_experiments}

We evaluate WorldMirror~2.0 as a standalone reconstruction foundation model on comprehensive benchmarks covering point map reconstruction (\Cref{tab:wm_pointmap}), camera pose estimation, depth estimation, novel view synthesis (\Cref{tab:wm_pose_depth_nvs}), and surface normal estimation (\Cref{tab:wm_normal}). All tasks are evaluated at three inference resolutions, \ie, low~($189\!\times\!259$), medium~($378\!\times\!518$, the default of WorldMirror~1.0), and high~($756\!\times\!1036$), to validate the resolution generalization enabled by normalized position encoding (\Cref{sec:normalized_rope}).

A consistent finding across all tasks is that WorldMirror~1.0 suffers from severe performance degradation at high resolution due to position extrapolation (\eg, camera pose AUC@30 drops from 86.13 to 66.29; 7-Scenes point map accuracy degrades from 0.043 to 0.079), whereas WorldMirror~2.0 maintains, and often improves, performance from medium to high resolution across every benchmark. Beyond the multi-resolution improvements, we further evaluate the effectiveness of flexible geometric prior injection (\Cref{sec:prior_guidance}) and the inference efficiency optimizations introduced in \Cref{sec:inference_efficiency}.

\subsubsection{Results \& Analysis of WorldMirror 2.0}

\paragraph{Point Map Reconstruction.}
We evaluate point map reconstruction on scene-level datasets (7-Scenes, NRGBD) and an object-level dataset (DTU), following the same sequence mappings as~\cite{wang2025pi}. As shown in \Cref{tab:wm_pointmap}, WorldMirror~1.0 at medium resolution already surpasses all baselines. WorldMirror~2.0 further improves at every resolution: at medium, it reduces the 7-Scenes accuracy error from 0.043 to 0.033; at high, the gap is even more pronounced (0.079$\to$0.037). Incorporating geometric priors yields additional gains, with WorldMirror~2.0 at high resolution with all priors achieving the best overall results on 7-Scenes and DTU.

\begin{table*}[t]
    \centering
    \caption{\textbf{Point map reconstruction on 7-Scenes, NRGBD, and DTU.} Both Acc.\,($\downarrow$) and Comp.\,($\downarrow$) are
  lower-is-better. Baseline methods are evaluated at M resolution. WorldMirror 2.0 generalizes across multiple resolutions.
  ``+\,all priors'' denotes additionally providing camera pose, intrinsics, and depth as input. ``L/M/H'': low ($182{\times}252$), 
  medium ($378{\times}518$), high ($756{\times}1036$) resolution. \colorbox{catgray}{Best} results are highlighted.}
    \setlength{\tabcolsep}{5pt}
    \renewcommand{\arraystretch}{1.15}
    \resizebox{1.0\textwidth}{!}{
    \begin{tabular}{l @{\hskip 12pt} cccc @{\hskip 12pt} cccc @{\hskip 12pt} cccc}
        \toprule[0.12em]
        \multirow{3}{*}{\textbf{Method}} &
        \multicolumn{4}{c}{\textbf{7-Scenes} {\small(scene)}} &
        \multicolumn{4}{c}{\textbf{NRGBD} {\small(scene)}} &
        \multicolumn{4}{c}{\textbf{DTU} {\small(object)}} \\
        \cmidrule(lr){2-5} \cmidrule(lr){6-9} \cmidrule(lr){10-13}
        & \multicolumn{2}{c}{Acc.\,$\downarrow$} & \multicolumn{2}{c}{Comp.\,$\downarrow$}
        & \multicolumn{2}{c}{Acc.\,$\downarrow$} & \multicolumn{2}{c}{Comp.\,$\downarrow$}
        & \multicolumn{2}{c}{Acc.\,$\downarrow$} & \multicolumn{2}{c}{Comp.\,$\downarrow$} \\
        \cmidrule(lr){2-3} \cmidrule(lr){4-5} \cmidrule(lr){6-7} \cmidrule(lr){8-9} \cmidrule(lr){10-11} \cmidrule(lr){12-13}
        & Mean & Med. & Mean & Med.
        & Mean & Med. & Mean & Med.
        & Mean & Med. & Mean & Med. \\
        \midrule[0.08em]
        Fast3R~\cite{yang2025fast3r}
            & 0.096 & 0.065 & 0.145 & 0.093
            & 0.135 & 0.091 & 0.163 & 0.104
            & 3.340 & 1.919 & 2.929 & 1.125 \\
        CUT3R~\cite{wang2025continuous}
            & 0.094 & 0.051 & 0.101 & 0.050
            & 0.104 & 0.041 & 0.079 & 0.031
            & 4.742 & 2.600 & 3.400 & 1.316 \\
        FLARE~\cite{zhang2025flare}
            & 0.085 & 0.058 & 0.142 & 0.104
            & 0.053 & 0.024 & 0.051 & 0.025
            & 2.541 & 1.468 & 3.174 & 1.420 \\
        VGGT~\cite{wang2025vggt}
            & 0.046 & 0.026 & 0.057 & 0.034
            & 0.051 & 0.029 & 0.066 & 0.038
            & 1.338 & 0.779 & 1.896 & 0.992 \\
        $\pi^3$~\cite{wang2025pi}
            & 0.048 & 0.028 & 0.072 & 0.047
            & 0.026 & 0.015 & 0.028 & 0.014
            & 1.198 & 0.646 & 1.849 & 0.607 \\

        \midrule[0.08em]
        \multicolumn{13}{l}{\textit{WorldMirror 1.0}} \\[2pt]
        \quad L
            & 0.043 & 0.029 & 0.055 & 0.029
            & 0.046 & 0.027 & 0.049 & 0.026
            & 1.476 & 0.889 & 1.768 & 0.917 \\
        \quad L\,+\,all priors
            & 0.021 & 0.014 & 0.026 & 0.016
            & 0.022 & 0.015 & 0.020 & 0.014
            & 1.347 & 0.854 & 1.392 & 0.865 \\[3pt]
        \quad M
            & 0.043 & 0.026 & 0.049 & 0.028
            & 0.041 & 0.020 & 0.045 & 0.019
            & 1.017 & 0.564 & 1.780 & 0.690 \\
        \quad M\,+\,all priors
            & 0.018 & 0.011 & 0.023 & 0.014
            & 0.016 & 0.011 & 0.014 & 0.010
            & 0.735 & 0.461 & 0.935 & 0.550 \\[3pt]
        \quad H
            & 0.079 & 0.052 & 0.087 & 0.051
            & 0.077 & 0.047 & 0.093 & 0.051
            & 2.271 & 1.083 & 2.113 & 0.825 \\
        \quad H\,+\,all priors
            & 0.042 & 0.024 & 0.041 & 0.024
            & 0.078 & 0.053 & 0.082 & 0.051
            & 1.773 & 0.792 & 1.478 & 0.782 \\

        \midrule[0.04em]
        \multicolumn{13}{l}{\textit{WorldMirror 2.0}} \\[2pt]
        \quad L
            & 0.041 & 0.027 & 0.052 & 0.027
            & 0.047 & 0.028 & 0.058 & 0.035
            & 1.352 & 0.824 & 2.009 & 0.880 \\
        \quad L\,+\,all priors
            & 0.019 & 0.012 & 0.024 & 0.014
            & 0.017 & 0.011 & 0.015 & 0.010
            & 1.100 & 0.748 & 1.201 & 0.774 \\[3pt]
        \quad M
            & 0.033 & 0.020 & 0.046 & 0.026
            & 0.039 & 0.024 & 0.047 & 0.027
            & 1.005 & 0.545 & 1.892 & 0.681 \\
        \quad M\,+\,all priors
            & 0.013 & \colorbox{catgray}{0.008} & 0.017 & 0.011
            & \colorbox{catgray}{0.013} & \colorbox{catgray}{0.009} & \colorbox{catgray}{0.013} & \colorbox{catgray}{0.009}
            & 0.690 & 0.458 & 0.876 & 0.506 \\[3pt]
        \quad H
            & 0.037 & 0.025 & 0.040 & 0.023
            & 0.046 & 0.026 & 0.053 & 0.030
            & 0.845 & 0.426 & 1.904 & 0.632 \\
        \quad H\,+\,all priors
            & \colorbox{catgray}{0.012} & {0.008} & \colorbox{catgray}{0.016} & \colorbox{catgray}{0.010}
            & 0.015 & 0.010 & 0.016 & 0.010
            & \colorbox{catgray}{0.554} & \colorbox{catgray}{0.343} & \colorbox{catgray}{0.771} & \colorbox{catgray}{0.398} \\
        \bottomrule[0.12em]
    \end{tabular}
    }
    \label{tab:wm_pointmap}
\end{table*}

\paragraph{Camera Pose, Depth, and Novel View Synthesis.}
In~\Cref{tab:wm_pose_depth_nvs}, we jointly report camera pose estimation and depth estimation on RealEstate10K, and novel view synthesis averaged across RealEstate10K and DL3DV, following the protocol of~\cite{wang2025pi}. For camera pose, WorldMirror~2.0 improves AUC@30 over WorldMirror~1.0 at every resolution: low~(80.55$\to$83.43), medium~(86.13$\to$86.48), and high~(66.29$\to$86.89). The high-resolution gain of over 20 points being particularly striking. For depth, WorldMirror~2.0 consistently reduces AbsRel (medium: 0.178$\to$0.167; high: 0.195$\to$0.162) and achieves the best $\delta{<}1.25$ accuracy (0.815 at high resolution). For novel view synthesis, WorldMirror~1.0 at high resolution suffers a dramatic quality collapse (PSNR from 21.34 to 17.78), whereas WorldMirror~2.0 maintains stable performance across resolutions (PSNR of 20.14/20.07/19.98 at low/medium/high) and achieves the best SSIM (0.726 at high resolution), confirming that higher-resolution inference improves structural fidelity.

\begin{table*}[t]
  \centering
  \caption{\textbf{Results of camera pose estimation and depth estimation (left); Results of novel view synthesis (right).}
  Camera pose and depth are evaluated on RealEstate10K; novel view synthesis is averaged across RealEstate10K and DL3DV.
  $\uparrow$\,/\,$\downarrow$ indicate higher-/lower-is-better.                                                                    
  Baseline methods are evaluated at M resolution. WorldMirror 2.0 generalizes across multiple resolutions.
  ``L/M/H'' denote low / medium / high inference resolution. \colorbox{catgray}{Best} results are highlighted.}
  \begin{minipage}[t]{0.522\textwidth}
  \centering
  \resizebox{\textwidth}{!}{
  \begin{tabular}{lcccc}
  \toprule[0.14em]
  \multicolumn{1}{l}{\multirow{3}{*}{\textbf{Method}}} &
  \multicolumn{2}{c}{\textbf{Camera Pose} (Avg.)} &
  \multicolumn{2}{c}{\textbf{Depth} (Avg.)} \\
  \cmidrule(r){2-3} \cmidrule(r){4-5}
  & AUC@30$\uparrow$ & RTA@30$\uparrow$
  & AbsRel$\downarrow$ & $\delta{<}1.25\uparrow$ \\
  \midrule[0.08em]
  Fast3R~\cite{yang2025fast3r}      & 61.68 & 81.86 & 0.353 & 0.666 \\
  CUT3R~\cite{wang2025continuous}    & 81.47 & 95.10 & 0.260 & 0.704 \\
  FLARE~\cite{zhang2025flare}       & 80.01 & 95.23 & 0.445 & 0.551 \\
  VGGT~\cite{wang2025vggt}          & 77.62 & 93.13 & 0.256 & 0.789 \\
  $\pi^3$~\cite{wang2025pi}         & 85.90 & \colorbox{catgray}{95.62} & \colorbox{catgray}{0.151} & 0.805 \\
  \midrule[0.08em]
  WorldMirror 1.0 (L)               & 80.55 & 93.68 & 0.225 & 0.751 \\
  WorldMirror 1.0 (M)               & 86.13 & 95.47 & 0.178 & 0.812 \\
  WorldMirror 1.0 (H)               & 66.29  & 89.62 & 0.195 & 0.797 \\
  WorldMirror 2.0 (L)               & 83.43  & 94.79 & 0.199 & 0.770 \\
  WorldMirror 2.0 (M)               & 86.48 & 95.55 & 0.167 & 0.806 \\
  WorldMirror 2.0 (H)               & \colorbox{catgray}{86.89} & 95.34 & 0.162 & \colorbox{catgray}{0.815} \\
  \bottomrule[0.14em]
  \end{tabular}
  }
  \end{minipage}
  \hfill
  \begin{minipage}[t]{0.468\textwidth}
  \centering
  \resizebox{\textwidth}{!}{
  \begin{tabular}{lccc}
  \toprule[0.14em]
  \multicolumn{1}{l}{\multirow{3}{*}{\textbf{Method}}} &
  \multicolumn{3}{c}{\textbf{NVS} (Avg.)} \\
  \cmidrule(r){2-4}
  & PSNR$\uparrow$ & SSIM$\uparrow$ & LPIPS$\downarrow$ \\
  \midrule[0.08em]
  FLARE~\cite{zhang2025flare}       & 15.84 & 0.545 & 0.500 \\
  AnySplat~\cite{jiang2025anysplat} & 18.57 & 0.626 & 0.255 \\
  \midrule[0.08em]
  WorldMirror 1.0 (L)              & 20.38 & 0.658 & 0.163 \\
  WorldMirror 1.0 (M)              & \colorbox{catgray}{21.34} & 0.709 & 0.181 \\
  WorldMirror 1.0 (H)              & 17.78 & 0.659 & 0.379 \\
  WorldMirror 2.0 (L)              & 20.14 & 0.679 & \colorbox{catgray}{0.149} \\
  WorldMirror 2.0 (M)              & 20.07 & 0.680 & 0.186 \\
  WorldMirror 2.0 (H)              & 19.98 & \colorbox{catgray}{0.726} & 0.235 \\
  \bottomrule[0.14em]
  \end{tabular}
  }
  \end{minipage}
  \label{tab:wm_pose_depth_nvs}
  \end{table*}

\paragraph{Surface Normal Estimation.}
Following~\cite{bae2024rethinking}, we evaluate surface normal estimation on ScanNet~\cite{dai2017scannet}, NYUv2~\cite{silberman2012indoor}, and iBims-1~\cite{koch2018evaluation}. As shown in~\Cref{tab:wm_normal}, WorldMirror~2.0 achieves the best results across all three benchmarks at medium resolution, surpassing dedicated single-task methods. The improvements of both depth and normal estimation are consistent with the explicit depth-to-normal supervision (\Cref{sec:depth2normal}) and pseudo-normal enhancement (\Cref{sec:data_enhancement}), which jointly strengthen geometric coupling between depth and normal predictions. 
The resolution generalization remains consistent: WorldMirror~2.0 at high resolution (ScanNet mean error 12.5) closely matches its medium-resolution optimum (12.3), whereas WorldMirror~1.0 degrades from 13.8 to 17.6.

\begin{table*}[t]
  \centering
  \caption{\textbf{Surface normal estimation on ScanNet, NYUv2, and iBims-1.} We compare with both regression-based and
  diffusion-based approaches. $\uparrow$\,/\,$\downarrow$ indicate higher-/lower-is-better. Baseline methods are evaluated at M
  resolution. WorldMirror 2.0 generalizes across multiple resolutions. ``L/M/H'' denote low / medium / high inference resolution. \colorbox{catgray}{Best} results are highlighted.}
  \resizebox{1.0\textwidth}{!}{
      \begin{tabular}{lccccccccc}
      \toprule[0.14em]
      \multicolumn{1}{l}{\multirow{3}{*}{\textbf{Method}}} &
      \multicolumn{3}{c}{\textbf{ScanNet}} &
      \multicolumn{3}{c}{\textbf{NYUv2}} &
      \multicolumn{3}{c}{\textbf{iBims-1}} \\
      \cmidrule(r){2-4} \cmidrule(r){5-7} \cmidrule(r){8-10}
      \multicolumn{1}{c}{} &
      mean $\downarrow$ & med $\downarrow$ & $22.5^{\circ}\uparrow$ &
      mean $\downarrow$ & med $\downarrow$ & $22.5^{\circ}\uparrow$ &
      mean $\downarrow$ & med $\downarrow$ & $22.5^{\circ}\uparrow$ \\
      \midrule[0.08em]
      OASIS~\cite{chen2020oasis} & 32.8 & 28.5 & 38.5 & 29.2 & 23.4 & 48.4 & 32.6 & 24.6 & 46.6 \\
      EESNU~\cite{bae2021estimating} & -- & -- & -- & 16.2 & 8.5 & 77.2 & 20.0 & 8.4 & 73.4 \\
      Omnidata v1~\cite{eftekhar2021omnidata} & 22.9 & 12.3 & 66.1 & 23.1 & 12.9 & 66.3 & 19.0 & 7.5 & 76.1 \\
      Omnidata v2~\cite{kar20223d} & 16.2 & 8.5 & 79.5 & 17.2 & 9.7 & 76.5 & 18.2 & 7.0 & 77.4 \\
      DSine~\cite{bae2024rethinking} & 16.2 & 8.3 & 78.7 & 16.4 & 8.4 & 77.7 & 17.1 & 6.1 & 79.0 \\
      GeoWizard~\cite{fu2024geowizard} & 16.7 & 9.5 & 78.3 & 19.5 & 11.7 & 74.5 & 20.4 & 9.4 & 76.4 \\
      StableNormal~\cite{ye2024stablenormal} & 16.0 & 9.9 & 81.5 & 18.5 & 11.2 & 77.5 & 17.9 & 8.5 & 80.4 \\
      \midrule
      WorldMirror 1.0 (L) & 14.4 & 7.4 & 81.5 & 16.0 & 8.2 & 78.7 & 19.0 & 7.2 & 76.3 \\
      WorldMirror 1.0 (M) & 13.8 & 7.3 & 82.5 & 15.1 & 8.0 & 80.1 & 16.6 & 6.4 & 80.1 \\
      WorldMirror 1.0 (H) & 17.6 & 12.5 & 76.2 & 19.1 & 13.3 & 72.4 & 19.2 & 10.9 & 76.9 \\
      WorldMirror 2.0 (L) & 12.7 & 6.8 & 83.7 & 14.4 & 7.8 & 80.4 & 15.6 & 6.2 & 80.4 \\
      WorldMirror 2.0 (M) & \colorbox{catgray}{12.3} & \colorbox{catgray}{6.5} & \colorbox{catgray}{84.3} & \colorbox{catgray}{13.9} & \colorbox{catgray}{7.6} & \colorbox{catgray}{81.4} & \colorbox{catgray}{14.2} & \colorbox{catgray}{5.6} & \colorbox{catgray}{82.4} \\
      WorldMirror 2.0 (H) & 12.5 & 6.8 & 84.2 & 14.0 & 7.8 & 81.4 & 14.5 & 6.1 & 82.0 \\
      \bottomrule[0.14em]
      \end{tabular}
  }
  \label{tab:wm_normal}
  \end{table*}

\begin{figure}
    \centering
    \includegraphics[width=1.02\linewidth]{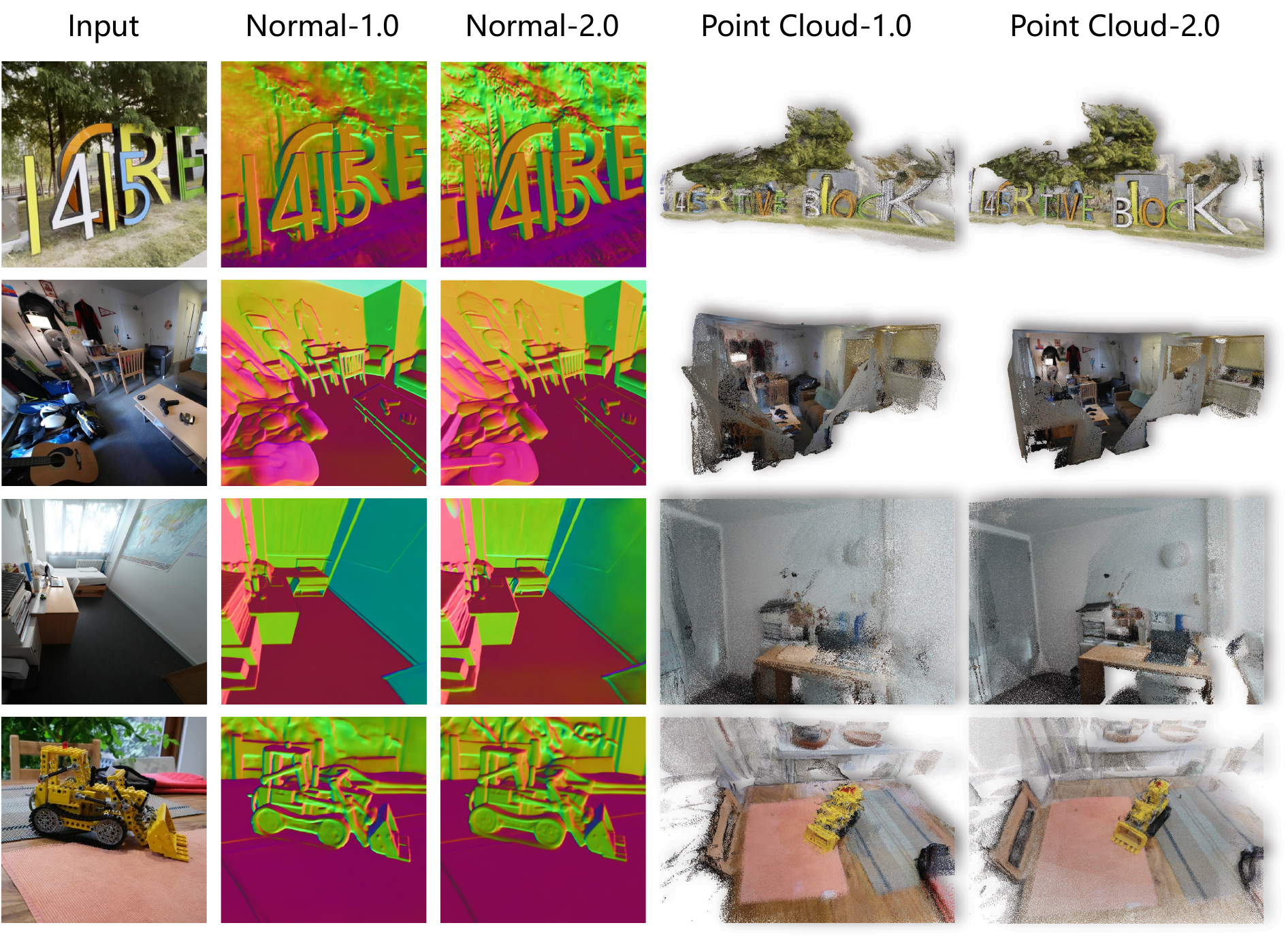}
    \caption{\textbf{Visual comparison of WorldMirror~1.0 and 2.0.} We compare predicted surface normals and reconstructed point clouds. WorldMirror~2.0 produces more accurate normals with finer structural details and more consistent multi-view point clouds.\label{fig:wm_visual_comparison}}
\end{figure}

\begin{figure}
    \centering
    \includegraphics[width=1.02\linewidth]{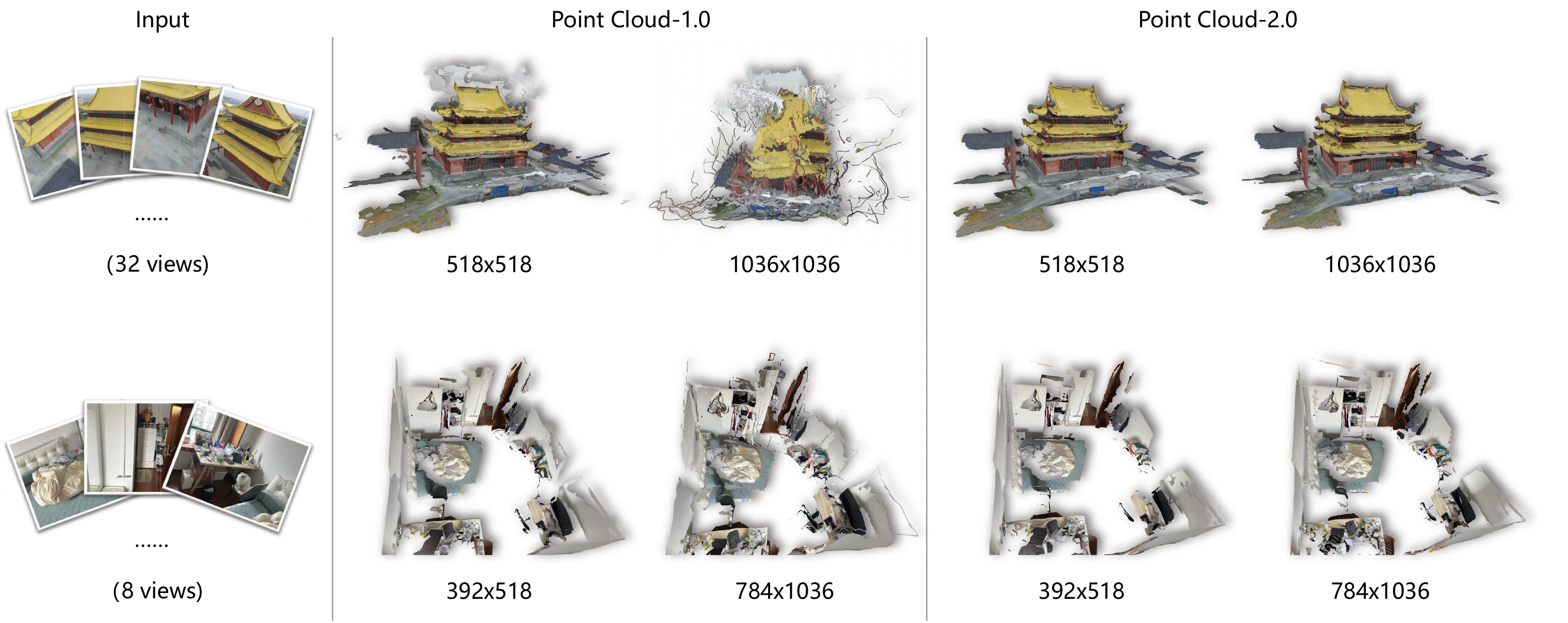}
    \caption{\textbf{Multi-resolution point cloud comparison of WorldMirror~1.0 and 2.0.} We evaluate under dense (32 views, top) and sparse (8 views, bottom) settings at different inference resolutions. WorldMirror~1.0 degrades severely at high resolution, while WorldMirror~2.0 maintains consistent reconstruction quality across all resolutions.}
    \label{fig:wm_resolution_comparison}
\end{figure}

\paragraph{Qualitative Results.}
We present visual comparisons between WorldMirror~1.0 and 2.0 in \Cref{fig:wm_visual_comparison} and \Cref{fig:wm_resolution_comparison}. As shown in \Cref{fig:wm_visual_comparison}, WorldMirror~2.0 produces sharper and more geometrically coherent surface normals, with finer structural details and fewer artifacts in complex regions. The reconstructed point clouds of WorldMirror~2.0 also exhibit tighter multi-view consistency, reflecting the geometric coupling enforced by the depth-to-normal supervision (\Cref{sec:depth2normal}) and more robust invalid-pixel handling via the depth mask prediction head (\Cref{sec:depth_mask}).

\Cref{fig:wm_resolution_comparison} further examines multi-resolution robustness under both dense (32 views) and sparse (8 views) input configurations. WorldMirror~1.0 produces reasonable point clouds at medium resolution ($518\!\times\!518$), but suffers from severe geometric degradation at high resolution ($1036\!\times\!1036$); under the dense 32-view setting, the point cloud structure collapses entirely. In contrast, WorldMirror~2.0 maintains stable and coherent reconstructions across all tested resolutions, directly validating the effectiveness of normalized position encoding (\Cref{sec:normalized_rope}) for flexible resolution inference.

\subsubsection{Inference-Time Evaluation}

\begin{figure}
    \centering
    \includegraphics[width=\linewidth]{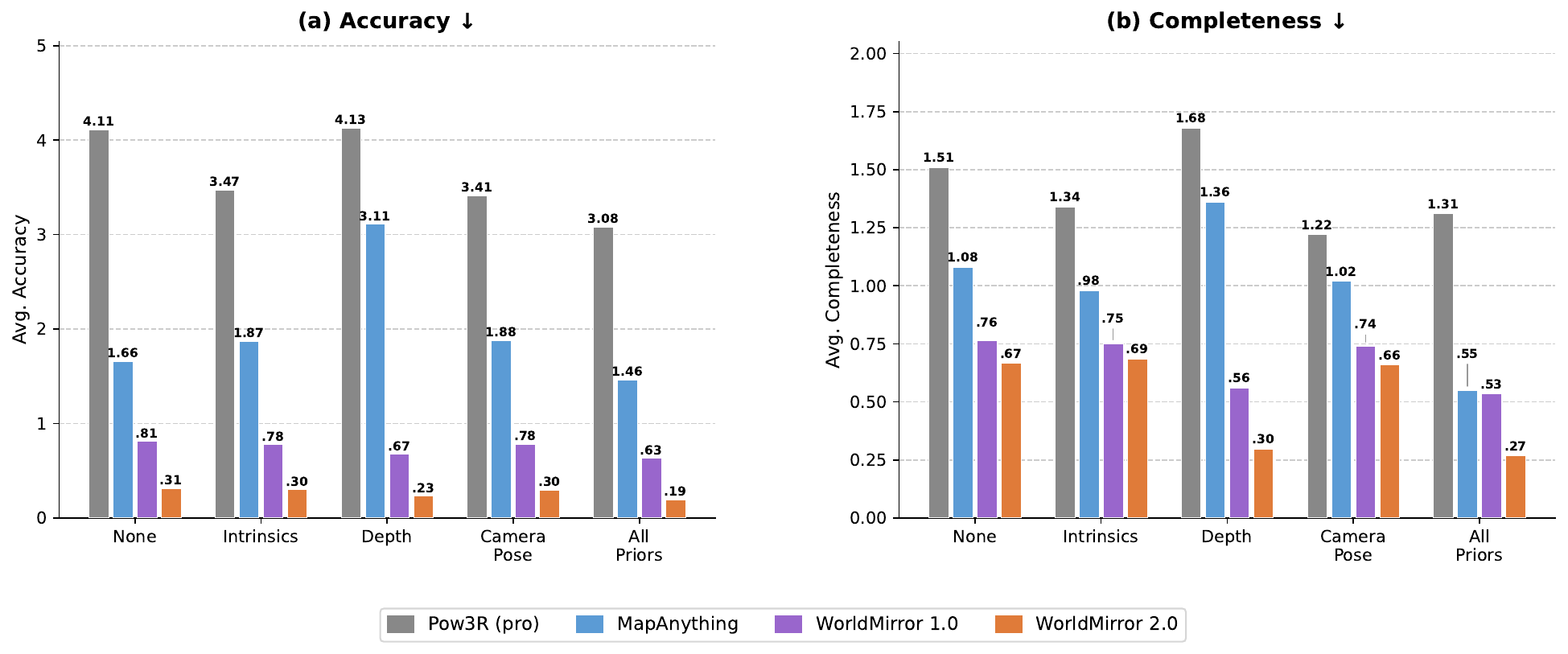}
    \caption{\textbf{Comparison with Pow3R and MapAnything under different prior conditions.} All methods are evaluated at high resolution ($756{\times}1036$). Results are averaged on 7-Scenes, NRGBD, and DTU datasets. Pow3R (pro) refers to the original Pow3R with Procrustes alignment. WorldMirror 2.0 demonstrates stronger geometric reasoning and 3D prior integration capabilities
   at high resolution.
    }
    \label{fig:wm_prior_comparison}
\end{figure}

\paragraph{Geometric Prior Injection.}
\label{sec:prior_guidance}
A distinctive feature of WorldMirror is its ability to flexibly incorporate geometric priors, including camera poses, intrinsics, and depth maps, through Any-Modal Tokenization (\Cref{sec:worldmirror_revisit}). We compare WorldMirror~1.0 and 2.0 with prior-guided methods Pow3R~\cite{jang2025pow3r} and MapAnything~\cite{keetha2025mapanything} under different prior conditions at high resolution (\Cref{fig:wm_prior_comparison}). WorldMirror~2.0 consistently outperforms all alternatives, with the largest improvements appearing in the camera-conditioned and all-priors settings. Camera poses capture the global geometric layout, calibrated intrinsics resolve metric scale ambiguity, and depth priors provide pixel-level constraints; combining all priors yields synergistic gains.

The practical benefit of this prior integration is further demonstrated in our world generation pipeline. As illustrated in \Cref{fig:gen_pcd_compare}, when conditioned on the same camera poses, WorldMirror~2.0 produces significantly more coherent and globally consistent point clouds than MapAnything~\cite{keetha2025mapanything} and DepthAnything3~\cite{lin2025depth}, both of which exhibit noticeable structural inconsistencies. This confirms that WorldMirror~2.0's learned multi-modal tokenization enables more effective utilization of geometric cues, making it well-suited as the reconstruction backbone for world generation.

\begin{table*}
    \centering                                                              
    \caption{\textbf{Inference efficiency of WorldMirror~2.0.} We report GPU memory (GB) and wall-clock time (s) for different
    numbers of input views at $518\!\times\!378$ resolution. ``Baseline'' refers to single-GPU FP32 inference (WorldMirror~1.0 setting). SP denotes Token/Frame Sequence Parallelism. All measurements on NVIDIA H20 GPUs.}                                   
    \label{tab:wm_inference_efficiency}
    \resizebox{1.0\textwidth}{!}{
    \begin{tabular}{lccccccccc}                                                                                  
    \toprule[0.14em]
    \multirow{2}{*}{\textbf{Configuration}} & \multirow{2}{*}{\textbf{\#GPUs}} &                                 
    \multicolumn{2}{c}{\textbf{32 views}} & \multicolumn{2}{c}{\textbf{64 views}} &
    \multicolumn{2}{c}{\textbf{128 views}} & \multicolumn{2}{c}{\textbf{256 views}}\\
    \cmidrule(r){3-4} \cmidrule(r){5-6} \cmidrule(r){7-8} \cmidrule(r){9-10}
    & & Mem. (GB) & Time (s) & Mem. (GB) & Time (s) & Mem. (GB) & Time (s) & Mem. (GB) & Time (s) \\
    \midrule[0.08em]
    Baseline (FP32, 1 GPU) & 1 & 24.95 & 2.45 & 38.56 & 6.27 & 59.26 & 18.00 & OOM & OOM \\
    + BF16 & 1 & 15.10 & 2.11 & 25.06 & 5.65 & 41.73 & 16.96 & 75.05 & 56.96 \\
    + SP ($\times$2 GPUs) & 2 & 26.32 & 1.59 & 41.31 & 3.73 & 64.75 & 10.53 & OOM & OOM \\
    + SP ($\times$4 GPUs) & 4 & 25.55 & 1.09 & 39.71 & 2.38 & 61.53 & 6.27 & OOM & OOM \\
    + SP + BF16 ($\times$4 GPUs) & 4 & 15.81 & 0.96 & 26.44 & 2.21 & 44.47 & 5.65 & 80.54 & 17.69 \\
    + SP + BF16 + FSDP ($\times$4 GPUs) & 4 & 14.04 & 0.93 & 24.67 & 2.20 & 42.71 & 5.60 & 78.78 & 17.52 \\
    \bottomrule[0.14em]
    \end{tabular}
    }
    \end{table*}

\paragraph{Inference Efficiency.}
\label{sec:infer_efficiency_exp}
We benchmark the inference efficiency optimizations of WorldMirror~2.0 introduced in \Cref{sec:inference_efficiency}. \Cref{tab:wm_inference_efficiency} reports per-GPU memory consumption and wall-clock inference time across different view counts at $518\!\times\!378$ resolution, measured on NVIDIA H20 GPUs. The single-GPU FP32 baseline runs out of memory (OOM) at 256 views. BF16 mixed-precision inference reduces per-GPU memory by approximately 40\% (\eg, 128 views: 59.26\,GB$\to$41.73\,GB) and critically enables 256-view inference (75.05\,GB) that is infeasible under FP32. Sequence parallelism (SP) provides substantial speedups by distributing computation across GPUs (\eg, 128 views: 18.00s$\to$6.27s with 4-GPU SP). The full combination of SP, BF16, and FSDP on 4 GPUs achieves the best trade-off: 128 views in 5.60s with 42.71\,GB per GPU (a $3.2\times$ speedup and 28\% memory reduction over the baseline), and 256 views in 17.52s with 78.78\,GB per GPU. These complementary strategies enable WorldMirror~2.0 to scale to substantially larger input configurations within practical memory and latency constraints.

\section{Conclusion}

In this report, we present \textbf{HY-World 2.0}, a comprehensive multi-modal world model framework that bridges the longstanding gap between 3D world generation and reconstruction. By dynamically adapting to diverse input modalities---ranging from sparse texts and single images to dense multi-view videos---our framework establishes a unified paradigm for offline 3D world modeling.
To achieve this, we introduced a four-stage pipeline. We scaled up panorama generation (\textbf{HY-Pano 2.0}) for high-fidelity world initialization and designed a semantic-aware trajectory planning (\textbf{WorldNav}) to guide optimal, collision-free routes for scene exploration. 
Furthermore, we significantly upgraded our generative world expansion (\textbf{WorldStereo 2.0}) by operating in a keyframe-latent space with spatially consistent memory.
Finally, we employ the world composition via our enhanced 3D reconstruction foundation (\textbf{WorldMirror 2.0}) to produce geometrically accurate and navigable 3DGS assets. We also propose a high-performance 3DGS rendering platform (\textbf{WorldLens}) to enable interactive exploration of 3D worlds with character support and lighting control.
Extensive evaluations demonstrate that HY-World 2.0 achieves state-of-the-art performance among open-source approaches, delivering visual quality, geometric consistency, and exploratory capabilities that are highly competitive with leading closed-source commercial models.

\clearpage

\definecolor{leadcolor}{HTML}{4285F4}
\definecolor{corecolor}{HTML}{EA4335}
\definecolor{contribcolor}{HTML}{34A853}
\definecolor{sponsorcolor}{HTML}{FBBC05}
\definecolor{boxbg}{HTML}{F8F9FA}
\definecolor{boxframe}{HTML}{DADCE0}

\section*{Contribution\textsuperscript{*}}
\let\thefootnote\relax
\footnotetext{* All in Alphabetical Order by First Name.}
\addtocounter{footnote}{-1}  
\addcontentsline{toc}{section}{Contribution}

\noindent{\faUserTie\;\textcolor{leadcolor}{\textbf{Project Lead}}} \textit{(Alphabetical Order)} \\[4pt]
\hspace{1.2em} Chunchao Guo, Tengfei Wang

\vspace{0.4cm}

\noindent{\faStar\;\textcolor{corecolor}{\textbf{Core Contributors}}} \textit{(Alphabetical Order)} \\[4pt]
\hspace{1.2em} Chenjie Cao, Junta Wu, Tengfei Wang, Xuhui Zuo, Yang Liu, Yisu Zhang, \\
\hspace{1.2em} Yuning Gong, Zhenwei Wang, Zhenyang Liu

\vspace{0.4cm}

\noindent{\faUsers\;\textcolor{contribcolor}{\textbf{Contributors}}} \textit{(Alphabetical Order)} 

\vspace{0.2cm}

\hspace{0.6em}{\faCogs\;\textbf{Engineering \& Infra:}} \\[3pt]
\hspace{1.8em} Bo Yuan, Coopers Li, Fan Yang, Haiyu Zhang, Jianchen Zhu, Jie Xiao, \\
\hspace{1.8em} Lei Wang, Minghui Chen, Penghao Zhao, Qi Chen, Wangchen Qin, \\
\hspace{1.8em} Xiang Yuan, Yifu Sun, Yihang Lian, Yuyang Yin, Zhiyuan Min

\vspace{0.25cm}

\hspace{0.6em}{\faPaintBrush\;\textbf{Data \& Art Design:}} \\[3pt]
\hspace{1.8em} Chao Zhang, Dongyuan Guo, Hang Cao, Jiaxin Lin, Jihong Zhang, \\
\hspace{1.8em} Junlin Yu, Lifu Wang, Lilin Wang, Peng He, Rui Chen, Rui Shao, \\
\hspace{1.8em} Sicong Liu, Xiaochuan Niu, Yi Sun, Yifei Tang, Yonghao Tan, Yuhong Liu

\vspace{0.4cm}

\noindent{\faHandshake\;\textcolor{black}{\textbf{Project Sponsors}}} \\[4pt]
\hspace{1.2em} Linus

\clearpage
\bibliographystyle{plain}
\bibliography{references}

@String(ECCV= {Eur. Conf. Comput. Vis.})

@String(TOG= {ACM Trans. Graph.})

@String(ECCV  = {ECCV})

@String(TOG   = {ACM TOG})

@article{hyworld15,
  title={HY-World 1.5: A Systematic Framework for Interactive World Modeling with Real-Time Latency and Geometric Consistency},
  author={Team HunyuanWorld},
  journal={arXiv preprint},
  year={2025}
}

@inproceedings{yang2025cogvideox,
  title={Cogvideox: Text-to-video diffusion models with an expert transformer},
  author={Yang, Zhuoyi and Teng, Jiayan and Zheng, Wendi and Ding, Ming and Huang, Shiyu and Xu, Jiazheng and Yang, Yuanming and Hong, Wenyi and Zhang, Xiaohan and Feng, Guanyu and others},
  booktitle={International Conference on Learning Representations},
  year={2025}
}

@article{kong2024hunyuanvideo,
  title={Hunyuanvideo: A systematic framework for large video generative models},
  author={Kong, Weijie and Tian, Qi and Zhang, Zijian and Min, Rox and Dai, Zuozhuo and Zhou, Jin and Xiong, Jiangfeng and Li, Xin and Wu, Bo and Zhang, Jianwei and others},
  journal={arXiv preprint arXiv:2412.03603},
  year={2024}
}

@article{wang2025wan,
  title={Wan: Open and Advanced Large-Scale Video Generative Models},
  author={Wang, Ang and Ai, Baole and Wen, Bin and Mao, Chaojie and Xie, Chen-Wei and Chen, Di and Yu, Feiwu and Zhao, Haiming and Yang, Jianxiao and Zeng, Jianyuan and others},
  journal={arXiv preprint arXiv:2503.20314},
  year={2025}
}

@inproceedings{cao2024mvsformer++,
  title={MVSFormer++: Revealing the Devil in Transformer's Details for Multi-View Stereo},
  author={Cao, Chenjie and Ren, Xinlin and Fu, Yanwei},
  booktitle={International Conference on Learning Representations},
  year={2024}
}

@article{hunyuanworld2025tencent,
    title={HunyuanWorld 1.0: Generating Immersive, Explorable, and Interactive 3D Worlds from Words or Pixels},
    author={Team HunyuanWorld},
    year={2025},
    journal={arXiv preprint}
}

@inproceedings{ren2025gen3c,
  title={Gen3c: 3d-informed world-consistent video generation with precise camera control},
  author={Ren, Xuanchi and Shen, Tianchang and Huang, Jiahui and Ling, Huan and Lu, Yifan and Nimier-David, Merlin and M{\"u}ller, Thomas and Keller, Alexander and Fidler, Sanja and Gao, Jun},
  booktitle={Proceedings of the IEEE/CVF Conference on Computer Vision and Pattern Recognition},
  year={2025}
}

@article{wu2023qalign,
  title={Q-Align: Teaching LMMs for Visual Scoring via Discrete Text-Defined Levels},
  author={Wu, Haoning and Zhang, Zicheng and Zhang, Weixia and Chen, Chaofeng and Li, Chunyi and Liao, Liang and Wang, Annan and Zhang, Erli and Sun, Wenxiu and Yan, Qiong and Min, Xiongkuo and Zhai, Guangtai and Lin, Weisi},
  journal={arXiv preprint arXiv:2312.17090},
  year={2023},
  institution={Nanyang Technological University and Shanghai Jiao Tong University and Sensetime Research},
  note={Equal Contribution by Wu, Haoning and Zhang, Zicheng. Project Lead by Wu, Haoning. Corresponding Authors: Zhai, Guangtai and Lin, Weisi.}
}

@article{lu2024genex,
  title={Genex: Generating an explorable world},
  author={Lu, Taiming and Shu, Tianmin and Xiao, Junfei and Ye, Luoxin and Wang, Jiahao and Peng, Cheng and Wei, Chen and Khashabi, Daniel and Chellappa, Rama and Yuille, Alan and others},
  journal={arXiv preprint arXiv:2412.09624},
  year={2024}
}

@article{feng2025dit360,
  title={Dit360: High-fidelity panoramic image generation via hybrid training},
  author={Feng, Haoran and Zhang, Dizhe and Li, Xiangtai and Du, Bo and Qi, Lu},
  journal={arXiv preprint arXiv:2510.11712},
  year={2025}
}

@article{yang2025matrix,
  title={Matrix-3d: Omnidirectional explorable 3d world generation},
  author={Yang, Zhongqi and Ge, Wenhang and Li, Yuqi and Chen, Jiaqi and Li, Haoyuan and An, Mengyin and Kang, Fei and Xue, Hua and Xu, Baixin and Yin, Yuyang and others},
  journal={arXiv preprint arXiv:2508.08086},
  year={2025}
}

@inproceedings{kalischek2025cubediff,
  title={Cubediff: Repurposing diffusion-based image models for panorama generation},
  author={Kalischek, Nikolai and Oechsle, Michael and Manhardt, Fabian and Henzler, Philipp and Schindler, Konrad and Tombari, Federico},
  booktitle={The Thirteenth International Conference on Learning Representations},
  year={2025}
}

@inproceedings{radford2021learning,
  title={Learning transferable visual models from natural language supervision},
  author={Radford, Alec and Kim, Jong Wook and Hallacy, Chris and Ramesh, Aditya and Goh, Gabriel and Agarwal, Sandhini and Sastry, Girish and Askell, Amanda and Mishkin, Pamela and Clark, Jack and others},
  booktitle={International conference on machine learning},
  pages={8748--8763},
  year={2021},
  organization={PmLR}
}

@inproceedings{peebles2023scalable,
  title={Scalable diffusion models with transformers},
  author={Peebles, William and Xie, Saining},
  booktitle={Proceedings of the IEEE/CVF international conference on computer vision},
  pages={4195--4205},
  year={2023}
}

@inproceedings{ling2024dl3dv,
  title={Dl3dv-10k: A large-scale scene dataset for deep learning-based 3d vision},
  author={Ling, Lu and Sheng, Yichen and Tu, Zhi and Zhao, Wentian and Xin, Cheng and Wan, Kun and Yu, Lantao and Guo, Qianyu and Yu, Zixun and Lu, Yawen and others},
  booktitle={Proceedings of the IEEE/CVF Conference on Computer Vision and Pattern Recognition},
  pages={22160--22169},
  year={2024}
}

@article{tartanair2020iros,
  title =   {TartanAir: A Dataset to Push the Limits of Visual SLAM},
  author =  {Wang, Wenshan and Zhu, Delong and Wang, Xiangwei and Hu, Yaoyu and Qiu, Yuheng and Wang, Chen and Hu, Yafei and Kapoor, Ashish and Scherer, Sebastian},
  booktitle = {2020 IEEE/RSJ International Conference on Intelligent Robots and Systems (IROS)},
  year =    {2020}
}

@inproceedings{arnold2022map,
  title={Map-free visual relocalization: Metric pose relative to a single image},
  author={Arnold, Eduardo and Wynn, Jamie and Vicente, Sara and Garcia-Hernando, Guillermo and Monszpart, Aron and Prisacariu, Victor and Turmukhambetov, Daniyar and Brachmann, Eric},
  booktitle={European Conference on Computer Vision},
  pages={690--708},
  year={2022},
  organization={Springer}
}

@inproceedings{xia2024rgbd,
  title={RGBD objects in the wild: scaling real-world 3D object learning from RGB-D videos},
  author={Xia, Hongchi and Fu, Yang and Liu, Sifei and Wang, Xiaolong},
  booktitle={Proceedings of the IEEE/CVF Conference on Computer Vision and Pattern Recognition},
  pages={22378--22389},
  year={2024}
}

@article{knapitsch2017tanks,
  title={Tanks and temples: Benchmarking large-scale scene reconstruction},
  author={Knapitsch, Arno and Park, Jaesik and Zhou, Qian-Yi and Koltun, Vladlen},
  journal={ACM Transactions on Graphics (ToG)},
  volume={36},
  number={4},
  pages={1--13},
  year={2017},
  publisher={ACM New York, NY, USA}
}

@inproceedings{barron2022mip,
  title={Mip-nerf 360: Unbounded anti-aliased neural radiance fields},
  author={Barron, Jonathan T and Mildenhall, Ben and Verbin, Dor and Srinivasan, Pratul P and Hedman, Peter},
  booktitle={Proceedings of the IEEE/CVF conference on computer vision and pattern recognition},
  pages={5470--5479},
  year={2022}
}

@article{zhou2025stable,
  title={STABLE VIRTUAL CAMERA: Generative View Synthesis with Diffusion Models},
  author={Zhou, Jensen Jinghao and Gao, Hang and Voleti, Vikram and Vasishta, Aaryaman and Yao, Chun-Han and Boss, Mark and Torr, Philip and Rupprecht, Christian and Jampani, Varun},
  journal={arXiv e-prints},
  pages={arXiv--2503},
  year={2025}
}

@inproceedings{wang2025vggt,
  title={VGGT: Visual Geometry Grounded Transformer},
  author={Wang, Jianyuan and Chen, Minghao and Karaev, Nikita and Vedaldi, Andrea and Rupprecht, Christian and Novotny, David},
  booktitle={Proceedings of the IEEE/CVF conference on computer vision and pattern recognition},
  year={2025}
}

@article{yu2025context,
  title={Context as memory: Scene-consistent interactive long video generation with memory retrieval},
  author={Yu, Jiwen and Bai, Jianhong and Qin, Yiran and Liu, Quande and Wang, Xintao and Wan, Pengfei and Zhang, Di and Liu, Xihui},
  journal={arXiv preprint arXiv:2506.03141},
  year={2025}
}

@article{cao2025uni3c,
  title={Uni3C: Unifying Precisely 3D-Enhanced Camera and Human Motion Controls for Video Generation},
  author={Cao, Chenjie and Zhou, Jingkai and Li, Shikai and Liang, Jingyun and Yu, Chaohui and Wang, Fan and Xue, Xiangyang and Fu, Yanwei},
  booktitle={SIGGRAPH Asia 2025 Conference Papers},
  year={2025}
}

@article{marr1976cooperative,
  title={Cooperative Computation of Stereo Disparity: A cooperative algorithm is derived for extracting disparity information from stereo image pairs.},
  author={Marr, David and Poggio, Tomaso},
  journal={Science},
  volume={194},
  number={4262},
  pages={283--287},
  year={1976},
  publisher={American Association for the Advancement of Science}
}

@article{yin2024improved,
  title={Improved distribution matching distillation for fast image synthesis},
  author={Yin, Tianwei and Gharbi, Micha{\"e}l and Park, Taesung and Zhang, Richard and Shechtman, Eli and Durand, Fredo and Freeman, Bill},
  journal={Advances in neural information processing systems},
  volume={37},
  pages={47455--47487},
  year={2024}
}

@inproceedings{yang2025fast3r,
  title={Fast3r: Towards 3d reconstruction of 1000+ images in one forward pass},
  author={Yang, Jianing and Sax, Alexander and Liang, Kevin J and Henaff, Mikael and Tang, Hao and Cao, Ang and Chai, Joyce and Meier, Franziska and Feiszli, Matt},
  booktitle={Proceedings of the Computer Vision and Pattern Recognition Conference},
  pages={21924--21935},
  year={2025}
}

@inproceedings{wang2025continuous,
  title={Continuous 3d perception model with persistent state},
  author={Wang, Qianqian and Zhang, Yifei and Holynski, Aleksander and Efros, Alexei A and Kanazawa, Angjoo},
  booktitle={Proceedings of the Computer Vision and Pattern Recognition Conference},
  pages={10510--10522},
  year={2025}
}

@article{keetha2025mapanything,
  title={MapAnything: Universal feed-forward metric 3D reconstruction},
  author={Keetha, Nikhil and M{\"u}ller, Norman and Sch{\"o}nberger, Johannes and Porzi, Lorenzo and Zhang, Yuchen and Fischer, Tobias and Knapitsch, Arno and Zauss, Duncan and Weber, Ethan and Antunes, Nelson and others},
  journal={arXiv preprint arXiv:2509.13414},
  year={2025}
}

@article{huang2025self,
  title={Self Forcing: Bridging the Train-Test Gap in Autoregressive Video Diffusion},
  author={Huang, Xun and Li, Zhengqi and He, Guande and Zhou, Mingyuan and Shechtman, Eli},
  journal={arXiv preprint arXiv:2506.08009},
  year={2025}
}

@inproceedings{jang2025pow3r,
  title={Pow3r: Empowering unconstrained 3d reconstruction with camera and scene priors},
  author={Jang, Wonbong and Weinzaepfel, Philippe and Leroy, Vincent and Agapito, Lourdes and Revaud, Jerome},
  booktitle={Proceedings of the Computer Vision and Pattern Recognition Conference},
  pages={1071--1081},
  year={2025}
}

@article{liu2025worldmirror,
  title={WorldMirror: Universal 3D World Reconstruction with Any-Prior Prompting},
  author={Liu, Yifan and Min, Zhiyuan and Wang, Zhenwei and Wu, Junta and Wang, Tengfei and Yuan, Yixuan and Luo, Yawei and Guo, Chunchao},
  journal={arXiv preprint arXiv:2510.10726},
  year={2025}
}

@article{yang2025flash,
  title={FlashWorld: High-quality 3D Scene Generation within Seconds},
  author={Xinyang Li and Tengfei Wang and Zixiao Gu and Shengchuan Zhang  and Chunchao Guo and Liujuan Cao},
  journal={arXiv preprint arXiv:2510.13678},
  year={2025}
}

@article{bahmani2025lyra,
  title={Lyra: Generative 3d scene reconstruction via video diffusion model self-distillation},
  author={Bahmani, Sherwin and Shen, Tianchang and Ren, Jiawei and Huang, Jiahui and Jiang, Yifeng and Turki, Haithem and Tagliasacchi, Andrea and Lindell, David B and Gojcic, Zan and Fidler, Sanja and others},
  journal={arXiv preprint arXiv:2509.19296},
  year={2025}
}

@article{schneider2025worldexplorer,
  title={WorldExplorer: Towards Generating Fully Navigable 3D Scenes},
  author={Schneider, Manuel-Andreas and H{\"o}llein, Lukas and Nie{\ss}ner, Matthias},
  journal={arXiv preprint arXiv:2506.01799},
  year={2025}
}

@article{li2025vmem,
  title={VMem: Consistent Interactive Video Scene Generation with Surfel-Indexed View Memory},
  author={Li, Runjia and Torr, Philip and Vedaldi, Andrea and Jakab, Tomas},
  journal={arXiv preprint arXiv:2506.18903},
  year={2025}
}

@article{wang2025moge,
  title={MoGe-2: Accurate Monocular Geometry with Metric Scale and Sharp Details},
  author={Wang, Ruicheng and Xu, Sicheng and Dong, Yue and Deng, Yu and Xiang, Jianfeng and Lv, Zelong and Sun, Guangzhong and Tong, Xin and Yang, Jiaolong},
  journal={arXiv preprint arXiv:2507.02546},
  year={2025}
}

@article{ravi2020pytorch3d,
    author = {Nikhila Ravi and Jeremy Reizenstein and David Novotny and Taylor Gordon
                  and Wan-Yen Lo and Justin Johnson and Georgia Gkioxari},
    title = {Accelerating 3D Deep Learning with PyTorch3D},
    journal = {arXiv:2007.08501},
    year = {2020},
}

@article{wang2023prolificdreamer,
  title={Prolificdreamer: High-fidelity and diverse text-to-3d generation with variational score distillation},
  author={Wang, Zhengyi and Lu, Cheng and Wang, Yikai and Bao, Fan and Li, Chongxuan and Su, Hang and Zhu, Jun},
  journal={Advances in neural information processing systems},
  volume={36},
  pages={8406--8441},
  year={2023}
}

@article{duan2025worldscore,
  title={Worldscore: A unified evaluation benchmark for world generation},
  author={Duan, Haoyi and Yu, Hong-Xing and Chen, Sirui and Fei-Fei, Li and Wu, Jiajun},
  journal={arXiv preprint arXiv:2504.00983},
  year={2025}
}

@inproceedings{cao2024leftrefill,
  title={Leftrefill: Filling right canvas based on left reference through generalized text-to-image diffusion model},
  author={Cao, Chenjie and Cai, Yunuo and Dong, Qiaole and Wang, Yikai and Fu, Yanwei},
  booktitle={Proceedings of the IEEE/CVF Conference on Computer Vision and Pattern Recognition},
  pages={7705--7715},
  year={2024}
}

@misc{marble_worldlabs_2026,
  author       = {{World Labs}}, 
  title        = {{Marble}}, 
  howpublished = {\url{https://marble.worldlabs.ai/}},
  year         = {2025}, 
  note         = {Accessed: 2026-03-30}
}

@article{sun2025worldplay,
  title={WorldPlay: Towards Long-Term Geometric Consistency for Real-Time Interactive World Modeling},
  author={Sun, Wenqiang and Zhang, Haiyu and Wang, Haoyuan and Wu, Junta and Wang, Zehan and Wang, Zhenwei and Wang, Yunhong and Zhang, Jun and Wang, Tengfei and Guo, Chunchao},
  journal={arXiv preprint arXiv:2512.14614},
  year={2025}
}

@misc{hunyuanvideo2025,
      title={HunyuanVideo 1.5 Technical Report}, 
      author={Tencent Hunyuan Foundation Model Team},
      year={2025},
      eprint={2511.18870},
      archivePrefix={arXiv},
      primaryClass={cs.CV},
      url={https://arxiv.org/abs/2511.18870}, 
}

@article{sitzmann2021light,
  title={Light field networks: Neural scene representations with single-evaluation rendering},
  author={Sitzmann, Vincent and Rezchikov, Semon and Freeman, Bill and Tenenbaum, Josh and Durand, Fredo},
  journal={Advances in Neural Information Processing Systems},
  volume={34},
  pages={19313--19325},
  year={2021}
}

@misc{worldstereo2026,
      title={WorldStereo: Bridging Camera-Guided Video Generation and Scene Reconstruction via 3D Geometric Memories}, 
      author={Tencent Hunyuan 3D Team},
      year={2026},
      archivePrefix={arXiv},
      primaryClass={cs.CV},
}

@article{su2024roformer,
  title={Roformer: Enhanced transformer with rotary position embedding},
  author={Su, Jianlin and Ahmed, Murtadha and Lu, Yu and Pan, Shengfeng and Bo, Wen and Liu, Yunfeng},
  journal={Neurocomputing},
  volume={568},
  pages={127063},
  year={2024},
  publisher={Elsevier}
}

@article{yang2026neoverse,
  title={NeoVerse: Enhancing 4D World Model with in-the-wild Monocular Videos},
  author={Yang, Yuxue and Fan, Lue and Shi, Ziqi and Peng, Junran and Wang, Feng and Zhang, Zhaoxiang},
  journal={arXiv preprint arXiv:2601.00393},
  year={2026}
}

@inproceedings{zhang2025sageattention2,
  title={Sageattention2: Efficient attention with thorough outlier smoothing and per-thread int4 quantization},
  author={Zhang, Jintao and Huang, Haofeng and Zhang, Pengle and Wei, Jia and Zhu, Jun and Chen, Jianfei},
  booktitle={International Conference on Learning Representations},
  year={2025}
}

@article{lin2025depth,
  title={Depth anything 3: Recovering the visual space from any views},
  author={Lin, Haotong and Chen, Sili and Liew, Junhao and Chen, Donny Y and Li, Zhenyu and Shi, Guang and Feng, Jiashi and Kang, Bingyi},
  journal={arXiv preprint arXiv:2511.10647},
  year={2025}
}

@article{carion2025sam,
  title={Sam 3: Segment anything with concepts},
  author={Carion, Nicolas and Gustafson, Laura and Hu, Yuan-Ting and Debnath, Shoubhik and Hu, Ronghang and Suris, Didac and Ryali, Chaitanya and Alwala, Kalyan Vasudev and Khedr, Haitham and Huang, Andrew and others},
  journal={arXiv preprint arXiv:2511.16719},
  year={2025}
}

@inproceedings{liu2024grounding,
  title={Grounding dino: Marrying dino with grounded pre-training for open-set object detection},
  author={Liu, Shilong and Zeng, Zhaoyang and Ren, Tianhe and Li, Feng and Zhang, Hao and Yang, Jie and Jiang, Qing and Li, Chunyuan and Yang, Jianwei and Su, Hang and others},
  booktitle={European conference on computer vision},
  pages={38--55},
  year={2024},
  organization={Springer}
}

@inproceedings{kim2025zim,
  title={Zim: Zero-shot image matting for anything},
  author={Kim, Beomyoung and Shin, Chanyong and Jeong, Joonhyun and Jung, Hyungsik and Lee, Se-Yun and Chun, Sewhan and Hwang, Dong-Hyun and Yu, Joonsang},
  booktitle={Proceedings of the IEEE/CVF International Conference on Computer Vision},
  pages={23828--23838},
  year={2025}
}

@article{yang2025qwen3,
  title={Qwen3 technical report},
  author={Yang, An and Li, Anfeng and Yang, Baosong and Zhang, Beichen and Hui, Binyuan and Zheng, Bo and Yu, Bowen and Gao, Chang and Huang, Chengen and Lv, Chenxu and others},
  journal={arXiv preprint arXiv:2505.09388},
  year={2025}
}

@misc{carion2025sam3segmentconcepts,
      title={SAM 3: Segment Anything with Concepts},
      author={Nicolas Carion and Laura Gustafson and Yuan-Ting Hu and Shoubhik Debnath and Ronghang Hu and Didac Suris and Chaitanya Ryali and Kalyan Vasudev Alwala and Haitham Khedr and Andrew Huang and Jie Lei and Tengyu Ma and Baishan Guo and Arpit Kalla and Markus Marks and Joseph Greer and Meng Wang and Peize Sun and Roman Rädle and Triantafyllos Afouras and Effrosyni Mavroudi and Katherine Xu and Tsung-Han Wu and Yu Zhou and Liliane Momeni and Rishi Hazra and Shuangrui Ding and Sagar Vaze and Francois Porcher and Feng Li and Siyuan Li and Aishwarya Kamath and Ho Kei Cheng and Piotr Dollár and Nikhila Ravi and Kate Saenko and Pengchuan Zhang and Christoph Feichtenhofer},
      year={2025},
      eprint={2511.16719},
      archivePrefix={arXiv},
      primaryClass={cs.CV},
      url={https://arxiv.org/abs/2511.16719},
}

@article{simeoni2025dinov3,
  title={Dinov3},
  author={Sim{\'e}oni, Oriane and Vo, Huy V and Seitzer, Maximilian and Baldassarre, Federico and Oquab, Maxime and Jose, Cijo and Khalidov, Vasil and Szafraniec, Marc and Yi, Seungeun and Ramamonjisoa, Micha{\"e}l and others},
  journal={arXiv preprint arXiv:2508.10104},
  year={2025}
}

@misc{recastnavigation,
  author       = {Mikko Mononen and the Recast Navigation Contributors},
  title        = {Recast Navigation: State-of-the-art navmesh generation and navigation for games},
  year         = {2009--2026},
  url          = {https://github.com/recastnavigation/recastnavigation},
  note         = {Accessed: [2026-02]},
  publisher    = {GitHub},
  howpublished = {Open-source software repository}
}

@incollection{dijkstra2022note,
  title={A note on two problems in connexion with graphs},
  author={Dijkstra, Edsger W},
  booktitle={Edsger Wybe Dijkstra: his life, work, and legacy},
  pages={287--290},
  year={2022}
}

@article{kerbl20233d,
  title={3D Gaussian Splatting for Real-Time Radiance Field Rendering},
  author={Kerbl, Bernhard and Kopanas, Georgios and Leimk{\"u}hler, Thomas and Drettakis, George},
  journal={ACM Transactions on Graphics},
  volume={42},
  number={4},
  year={2023}
}

@inproceedings{xie2024physgaussian,
  title={Physgaussian: Physics-integrated 3d gaussians for generative dynamics},
  author={Xie, Tianyi and Zong, Zeshun and Qiu, Yuxing and Li, Xuan and Feng, Yutao and Yang, Yin and Jiang, Chenfanfu},
  booktitle={Proceedings of the IEEE/CVF Conference on Computer Vision and Pattern Recognition},
  pages={4389--4398},
  year={2024}
}

@incollection{lorensen1998marching,
  title={Marching cubes: A high resolution 3D surface construction algorithm},
  author={Lorensen, William E and Cline, Harvey E},
  booktitle={Seminal graphics: pioneering efforts that shaped the field},
  pages={347--353},
  year={1998}
}

@article{hollein2026world,
  title={World Reconstruction From Inconsistent Views},
  author={H{\"o}llein, Lukas and Nie{\ss}ner, Matthias},
  journal={arXiv preprint arXiv:2603.16736},
  year={2026}
}

@article{jang2016categorical,
  title={Categorical reparameterization with gumbel-softmax},
  author={Jang, Eric and Gu, Shixiang and Poole, Ben},
  journal={arXiv preprint arXiv:1611.01144},
  year={2016}
}

@inproceedings{liu2025maskgaussian,
  title={Maskgaussian: Adaptive 3d gaussian representation from probabilistic masks},
  author={Liu, Yifei and Zhong, Zhihang and Zhan, Yifan and Xu, Sheng and Sun, Xiao},
  booktitle={Proceedings of the Computer Vision and Pattern Recognition Conference},
  pages={681--690},
  year={2025}
}

@inproceedings{johnson2016perceptual,
  title={Perceptual losses for real-time style transfer and super-resolution},
  author={Johnson, Justin and Alahi, Alexandre and Fei-Fei, Li},
  booktitle={European conference on computer vision},
  pages={694--711},
  year={2016},
  organization={Springer}
}

@article{wang2025pi,
  title={pi3: Scalable Permutation-Equivariant Visual Geometry Learning},
  author={Wang, Yifan and Zhou, Jianjun and Zhu, Haoyi and Chang, Wenzheng and Zhou, Yang and Li, Zizun and Chen, Junyi and Pang, Jiangmiao and Shen, Chunhua and He, Tong},
  journal={arXiv preprint arXiv:2507.13347},
  year={2025}
}

@inproceedings{dai2017scannet,
  title={Scannet: Richly-annotated 3d reconstructions of indoor scenes},
  author={Dai, Angela and Chang, Angel X and Savva, Manolis and Halber, Maciej and Funkhouser, Thomas and Nie{\ss}ner, Matthias},
  booktitle={Proceedings of the IEEE conference on computer vision and pattern recognition},
  pages={5828--5839},
  year={2017}
}

@inproceedings{silberman2012indoor,
  title={Indoor segmentation and support inference from rgbd images},
  author={Silberman, Nathan and Hoiem, Derek and Kohli, Pushmeet and Fergus, Rob},
  booktitle={European conference on computer vision},
  pages={746--760},
  year={2012},
  organization={Springer}
}

@inproceedings{koch2018evaluation,
  title={Evaluation of cnn-based single-image depth estimation methods},
  author={Koch, Tobias and Liebel, Lukas and Fraundorfer, Friedrich and Korner, Marco},
  booktitle={Proceedings of the European Conference on Computer Vision (ECCV) Workshops},
  pages={0--0},
  year={2018}
}

@inproceedings{chen2020oasis,
  title={Oasis: A large-scale dataset for single image 3d in the wild},
  author={Chen, Weifeng and Qian, Shengyi and Fan, David and Kojima, Noriyuki and Hamilton, Max and Deng, Jia},
  booktitle={Proceedings of the IEEE/CVF Conference on Computer Vision and Pattern Recognition},
  pages={679--688},
  year={2020}
}

@inproceedings{bae2021estimating,
  title={Estimating and exploiting the aleatoric uncertainty in surface normal estimation},
  author={Bae, Gwangbin and Budvytis, Ignas and Cipolla, Roberto},
  booktitle={Proceedings of the IEEE/CVF International Conference on Computer Vision},
  pages={13137--13146},
  year={2021}
}

@inproceedings{eftekhar2021omnidata,
  title={Omnidata: A scalable pipeline for making multi-task mid-level vision datasets from 3d scans},
  author={Eftekhar, Ainaz and Sax, Alexander and Malik, Jitendra and Zamir, Amir},
  booktitle={Proceedings of the IEEE/CVF International Conference on Computer Vision},
  pages={10786--10796},
  year={2021}
}

@article{liu2025survey,
  title={A survey on cache methods in diffusion models: Toward efficient multi-modal generation},
  author={Liu, Jiacheng and Wang, Xinyu and Lin, Yuqi and Wang, Zhikai and Wang, Peiru and Cai, Peiliang and Zhou, Qinming and Yan, Zhengan and Yan, Zexuan and Shi, Zhengyi and others},
  journal={arXiv preprint arXiv:2510.19755},
  year={2025}
}

@inproceedings{zhang2025flare,
  title={Flare: Feed-forward geometry, appearance and camera estimation from uncalibrated sparse views},
  author={Zhang, Shangzhan and Wang, Jianyuan and Xu, Yinghao and Xue, Nan and Rupprecht, Christian and Zhou, Xiaowei and Shen, Yujun and Wetzstein, Gordon},
  booktitle={Proceedings of the Computer Vision and Pattern Recognition Conference},
  pages={21936--21947},
  year={2025}
}

@inproceedings{bae2024rethinking,
  title={Rethinking inductive biases for surface normal estimation},
  author={Bae, Gwangbin and Davison, Andrew J},
  booktitle={Proceedings of the IEEE/CVF Conference on Computer Vision and Pattern Recognition},
  pages={9535--9545},
  year={2024}
}

@article{jiang2025anysplat,
  title={AnySplat: Feed-forward 3D Gaussian Splatting from Unconstrained Views},
  author={Jiang, Lihan and Mao, Yucheng and Xu, Linning and Lu, Tao and Ren, Kerui and Jin, Yichen and Xu, Xudong and Yu, Mulin and Pang, Jiangmiao and Zhao, Feng and others},
  journal={arXiv preprint arXiv:2505.23716},
  year={2025}
}

@inproceedings{kar20223d,
  title={3d common corruptions and data augmentation},
  author={Kar, O{\u{g}}uzhan Fatih and Yeo, Teresa and Atanov, Andrei and Zamir, Amir},
  booktitle={Proceedings of the IEEE/CVF Conference on Computer Vision and Pattern Recognition},
  pages={18963--18974},
  year={2022}
}

@inproceedings{fu2024geowizard,
  title={Geowizard: Unleashing the diffusion priors for 3d geometry estimation from a single image},
  author={Fu, Xiao and Yin, Wei and Hu, Mu and Wang, Kaixuan and Ma, Yuexin and Tan, Ping and Shen, Shaojie and Lin, Dahua and Long, Xiaoxiao},
  booktitle={European Conference on Computer Vision},
  pages={241--258},
  year={2024},
  organization={Springer}
}

@article{ye2024stablenormal,
  title={Stablenormal: Reducing diffusion variance for stable and sharp normal},
  author={Ye, Chongjie and Qiu, Lingteng and Gu, Xiaodong and Zuo, Qi and Wu, Yushuang and Dong, Zilong and Bo, Liefeng and Xiu, Yuliang and Han, Xiaoguang},
  journal={ACM Transactions on Graphics (TOG)},
  volume={43},
  number={6},
  pages={1--18},
  year={2024},
  publisher={ACM New York, NY, USA}
}

@article{ha2018recurrent,
  title={Recurrent world models facilitate policy evolution},
  author={Ha, David and Schmidhuber, J{\"u}rgen},
  journal={Advances in neural information processing systems},
  volume={31},
  year={2018}
}

@article{janner2022planning,
  title={Planning with diffusion for flexible behavior synthesis},
  author={Janner, Michael and Du, Yilun and Tenenbaum, Joshua B and Levine, Sergey},
  journal={arXiv preprint arXiv:2205.09991},
  year={2022}
}

@article{dong2026learning,
  title={Learning to model the world: A survey of world models in artificial intelligence},
  author={Dong, Jiahua and Lyu, Qi and Liu, Baichen and Wang, Xudong and Liang, Wenqi and Zhang, Duzhen and Tu, Jiahang and Li, Hongliu and Zhao, Hanbin and Ding, Henghui and others},
  year={2026},
  publisher={Preprints}
}

@misc{deepmind2025genie3,
  title        = {Genie 3: A New Frontier for World Models},
  author       = {{Google DeepMind}},
  year         = {2025},
  howpublished = {\url{https://deepmind.google/blog/genie-3-a-new-frontier-for-world-models/}},
  note         = {Blog post, August 5, 2025}
}

@article{wang2026worldcompass,
  title={Worldcompass: Reinforcement learning for long-horizon world models},
  author={Wang, Zehan and Wang, Tengfei and Zhang, Haiyu and Zuo, Xuhui and Wu, Junta and Wang, Haoyuan and Sun, Wenqiang and Wang, Zhenwei and Cao, Chenjie and Zhao, Hengshuang and others},
  journal={arXiv preprint arXiv:2602.09022},
  year={2026}
}

@inproceedings{yang2025layerpano3d,
  title={Layerpano3d: Layered 3d panorama for hyper-immersive scene generation},
  author={Yang, Shuai and Tan, Jing and Zhang, Mengchen and Wu, Tong and Wetzstein, Gordon and Liu, Ziwei and Lin, Dahua},
  booktitle={Proceedings of the special interest group on computer graphics and interactive techniques conference conference papers},
  pages={1--10},
  year={2025}
}

@article{team2026advancing,
  title={Advancing Open-source World Models},
  author={Team, Robbyant and Gao, Zelin and Wang, Qiuyu and Zeng, Yanhong and Zhu, Jiapeng and Cheng, Ka Leong and Li, Yixuan and Wang, Hanlin and Xu, Yinghao and Ma, Shuailei and others},
  journal={arXiv preprint arXiv:2601.20540},
  year={2026}
}

@article{mao2025yume,
  title={Yume-1.5: A Text-Controlled Interactive World Generation Model},
  author={Mao, Xiaofeng and Li, Zhen and Li, Chuanhao and Xu, Xiaojie and Ying, Kaining and He, Tong and Pang, Jiangmiao and Qiao, Yu and Zhang, Kaipeng},
  journal={arXiv preprint arXiv:2512.22096},
  year={2025}
}

\end{document}